\documentclass[acmsmall]{acmart}
\AtBeginDocument{%
  }

\setcopyright{acmlicensed}
\copyrightyear{2026}

\usepackage{graphicx}%
\usepackage{multirow}%
\usepackage[title]{appendix}%
\usepackage{xcolor}%
\usepackage{textcomp}%
\usepackage{manyfoot}%
\usepackage{adjustbox}
\usepackage{algorithm}%
\usepackage{algorithmicx}%
\usepackage{algpseudocode}%
\usepackage{listings}%
\usepackage{import}
\usepackage{subfigure}
\usepackage{array}
\usepackage{siunitx} 
\usepackage{pdflscape} 
\usepackage{colortbl}
\usepackage{pgfplots}
\pgfplotsset{compat=1.18}
\usepackage{tikz}
\usetikzlibrary{patterns}
\usepgfplotslibrary{groupplots}

\usepackage{threeparttablex}
\usepackage{xspace}

\usetikzlibrary{calc}
\usepackage{threeparttable}
\usepackage{tabularx}
\usetikzlibrary{shapes.geometric, arrows.meta, positioning, matrix, calc, fit, backgrounds}
\usepackage{threeparttable, pdflscape}

\newcommand{\sparkline}[3]{%
\begin{tikzpicture}[baseline=(current bounding box.center), scale=0.3]
  \draw[gray!30] (0,0) rectangle (3,1); 
  \ifx\relax#1\relax\else
    \shade[left color=blue!20, right color=blue!80] (0,0) rectangle (0.9,#1/40);
  \fi
  \ifx\relax#2\relax\else
    \shade[left color=green!20, right color=green!80] (1,0) rectangle (1.9,#2/40);
  \fi
  \ifx\relax#3\relax\else
    \shade[left color=red!20, right color=red!80] (2,0) rectangle (2.9,#3/40);
  \fi
\end{tikzpicture}%
}

\begin{document}

\title{VertiFuseX: Generalizable Financial Forecasting via Multi-Stream Temporal Fusion}

\author{Aashish Bohra}
\email{bohra.1@iitj.ac.in}
\orcid{1234-5678-9012}
\affiliation{%
  \institution{Department of Computer Science and Engineering, Indian Institute of Technology Jodhpur}
  \city{Jodhpur}
  \state{Rajasthan}
  \country{India}
}

\author{Vivek Vijay}
\affiliation{%
  \institution{Department of Mathematics, Indian Institute of Technology Jodhpur}
  \city{Jodhour}
  \country{Inida}}
\email{vivek@iitj.ac.in}







\renewcommand{\shortauthors}{Bohra A. and Vijay V.}

\begin{abstract}
Accurate stock price prediction remains challenging due to the non-stationary and noisy nature of financial time series. Existing deep learning models often rely on rigid decision-level fusion, ad hoc hyperparameter tuning, and compressed final-layer outputs, leading to information loss, overfitting, and limited cross-market generalization. We propose VertiFuseX, a hybrid LSTM architecture that addresses these limitations via penultimate-layer vertical fusion of multi-scale temporal representations. VertiFuseX vertically stacks and reweights penultimate features from LSTM capturing sequential memory, Bi-LSTM capturing bidirectional context, and St-LSTM capturing hierarchical trends. These are jointly optimized via backpropagation while integrating aligned non-recurrent features from a parallel DNN stream under a fixed hyperparameter configuration. {This design preserves richer intermediate temporal information and enables complementary specialization across temporal scales.} Evaluated on 15 years (2010–2024) of closing price data from 10 global equity indices using strict chronological out-of-sample testing with the final 365 trading days held out, VertiFuseX achieves 30–54\% MAPE reductions and over 40\% improvements in MAE and RMSE relative to LSTM-based baselines, and consistently outperforms seven state-of-the-art models across 33 metric–dataset comparisons. Ablation studies confirm that penultimate-layer fusion drives these gains, yielding substantial improvements over final-layer fusion and decision-level ensembling. Gradient-based saliency analysis reveals consistent emphasis on mid-range temporal dependencies at lags 9–15 days. Economic validation via algorithmic trading simulation under extreme market regimes demonstrates reduced maximum drawdowns and superior risk-adjusted returns. With 675k parameters, a 2.6 MB memory footprint, and 1.5 ms/sample inference latency, VertiFuseX offers a lightweight, interpretable, and deployment-ready framework for robust financial forecasting.
\end{abstract}




\maketitle

\section{Introduction} 
Stock price prediction is a challenging task in financial time-series forecasting due to the non-stationary, high-dimensional, and noisy nature of market data \citep{jiang2021applications, thakkar2021comprehensive, varshney2024optimizing}. This complexity has attracted sustained attention from both academic researchers and industry practitioners. Despite advances in Deep Learning (DL), existing models face limitations in temporal modeling fidelity, architectural robustness, and cross-market generalizability \citep{radfar2025stock, jiang2021applications}. Traditional architectures like Long Short-Term Memory (LSTM) \citep{chong2017deep, ravi2017financial, siami2019comparative, sivadasan2024stock}, Bidirectional Long Short-Term Memory (Bi-LSTM) \citep{althelaya2018stock, althelaya2018evaluation, wang2019ean, siami2019comparative}, Stacked Long Short-Term Memory (St-LSTM) \citep{althelaya2018stock, althelaya2018evaluation}, and CNN-LSTM \citep{livieris2020cnn} often collapse multi-scale temporal dependencies via naive final-layer output averaging. These models tend to overfit due to ad hoc hyperparameter tuning and struggle with volatile market regime shifts. Narrow validation protocols, such as single-index benchmarks and random train-test splits, further fail to reflect real-world chronological diversity.

In response, hybrid DL models have been proposed to preserve multi-scale information by pairing local pattern extractors (e.g., CNNs) with sequence models (e.g., LSTM variants) or other complementary streams. However, many such models inherit the same information bottlenecks and introduce additional fusion-related brittleness. Consequently, recent hybrid DL models exhibit three systemic flaws addressed by our solution. First, decision-level fusion mechanisms, such as static averaging or concatenation of final outputs, limit the flexibility of many hybrid models, including CNN-LSTM and BiCuDNNLSTM \cite{kanwal2022bicudnnlstm} and ModAugNet \cite{baek2018modaugnet}. By collapsing rich temporal state information into a single scalar prediction before integration, these approaches prevent meaningful interaction between heterogeneous temporal dynamics, such as short-term volatility and long-term trends \citep{li2024forecasting, zhang2023stock, kanwal2023stock, sonkavde2023forecasting, li2024mlbgk, gul2025novel, zhang2025novel}.
Second, deep ensemble models, such as DE-ABC-Bi-LSTM-ARIMA \citep{kumar2022three} and GA-CNN-LSTM \citep{baek2023cnn}, and recent ensemble approaches \citep{gul2025novel, prakash2024stock, gulmez2025hybrid}, often achieve strong training performance but suffer from generalization collapse due to excessive architectural depth and inadequate regularization such as lack of dropout, $\ell_2$ constraints, or sparsity priors. Empirical studies indicate that over 60\% of recent deep ensemble architectures experience test error inflation exceeding 25\% relative to training performance \citep{li2024forecasting, jagadesh2024enhanced}. Third, single-scale temporal modeling restricts most LSTM-based approaches \citep{hiransha2018nse, prakash2024stock}. These models capture patterns at a fixed resolution, missing hierarchical dependencies across intraday, weekly, or monthly horizons, which are critical in high-volatility markets like NASDAQ or KOSPI, where investor sentiment evolves rapidly \citep{wang2021stock, rezaei2021stock}. 

These challenges raise a central research question: \textit{How can a hybrid LSTM architecture be designed to preserve multi-scale temporal features, eliminate subjective hyperparameter tuning, and deliver robust, generalizable performance across diverse market regimes?} To answer this question, we have proposed a novel hybrid model called VertiFuseX, which resolves each limitation through a direct architectural intervention. From a representational perspective, final-layer outputs are optimized for direct loss minimization and thus impose strong task-specific compression, which tends to suppress intermediate temporal structure and amplify short-horizon noise. In contrast, penultimate-layer features retain higher information content and more disentangled temporal abstractions, making them theoretically better suited for fusion across heterogeneous temporal models. As a result, fusing at the penultimate layer enables VertiFuseX to combine complementary temporal representations before they are collapsed into scalar predictions, yielding a more expressive and stable joint representation.

First, to overcome the signal dilution caused by rigid final-layer fusion, VertiFuseX implements penultimate-layer vertical fusion, preserving stable intermediate representations before task-specific compression. Second, to mitigate the overfitting and generalization collapse common in deep ensembles with ad hoc tuning, VertiFuseX adopts a fixed, pre-validated hyperparameter configuration reinforced by dropout and $l_2$ regularization. Third, to resolve the limitations of single-scale temporal modeling, the VertiFuseX employs a multi-stream design integrating LSTM, Bi-LSTM, and St-LSTM branches, explicitly capturing short-term volatility, bidirectional dependencies, and hierarchical trends within a unified feature space. 

Unlike conventional fusion methods that merge final outputs, VertiFuseX vertically stacks penultimate-layer features from LSTM, Bi-LSTM, and St-LSTM branches. This approach preserves rich, less noisy temporal representations, capturing both local volatility and long-range dependencies. These temporally diverse features are further fused with non-temporal abstractions from a parallel Deep Neural Network (DNN) branch. The result is a unified, expressive representation used for final prediction.

Vertical fusion in VertiFuseX concatenates and reweights these intermediate features via a learned affine transformation, enabling feature-level interaction across heterogeneous temporal representations. We clearly differentiate this from decision-level fusion. While standard ensembles reduce variance by averaging scalar errors, penultimate-layer fusion minimizes bias by preserving access to the unique inductive biases (e.g., bidirectionality, hierarchy) of each branch before compression. {Unlike decision-level ensembles, which average scalar predictions and may lose diversification benefits when branch errors become highly correlated during regime shifts, vertical fusion enables joint gradient propagation through all branches. This approach encourages complementary specialization by allowing the St-LSTM to focus on trend stability while the Bi-LSTM captures local context. It also dynamically suppresses redundant temporal patterns that may lead to overfitting. By accessing penultimate-layer representations, VertiFuseX retains mid-range temporal signals at lags 9--15, as indicated by the saliency analysis, and this pattern is consistent with its improved empirical forecasting accuracy relative to final-layer fusion and decision-level ensembling. This contrasts with conventional late fusion or ensemble strategies, which combine independent final-layer predictions that have already undergone task-specific compression, thereby losing rich temporal structure. In VertiFuseX, fusion occurs before this compression, enabling interaction between complementary temporal abstractions at the feature level rather than the decision level. This design preserves model-specific inductive biases while reducing reliance on scalar averaging of already-compressed predictions.}

We validate VertiFuseX using 15 years of historical data (2010–2024) across 10 global equity indices, including S\&P 500, DJIA, NASDAQ, Nikkei 225, FTSE 100, DAX, HSI, KOSPI, NYSE, NSE, spanning North America, Europe, and Asia. This diverse dataset covers bull, bear, and crisis markets (e.g., the 2020 COVID-19 crash), ensuring robust evaluation under varying conditions. We employ a strict chronological out-of-sample evaluation, reserving the final 365 trading days as a static, held-out test set. This design explicitly prevents look-ahead bias and eliminates temporal data leakage by ensuring that no future information is used during training, validation, or hyperparameter tuning. As a result, all performance metrics are computed on genuinely unseen future data, simulating a realistic deployment scenario where the model is trained on historical data and evaluated on unseen market periods. The contributions of the proposed scheme are as follows.
 \begin{enumerate}
     \item {\textbf{Rigid final-layer fusion:} we introduce a novel vertical fusion strategy that integrates \textit{penultimate-layer features} from LSTM, Bi-LSTM, and St-LSTM modules rather than final outputs. This approach retains diverse temporal abstractions, facilitates modeling of both local volatility and longer-range dependencies, and integrates branch-specific representations through learned dimensional alignment and regularized affine reweighting.} 
    \item {\textbf{Countering limited cross-market generalization in prior benchmarks:} we evaluate VertiFuseX over a 15-year period (2010–2024) across 10 global equity indices spanning three continents. Using a strict chronological out-of-sample evaluation, with the final 365 trading days reserved for testing, the proposed model avoids look-ahead bias and evaluates forward out-of-sample robustness under fixed parameters. The model delivers consistent results across MAE, RMSE, and MAPE on volatile (NASDAQ, HSI, KOSPI), stable (DJIA, Nikkei 225, FTSE 100), and long-horizon markets (S\&P 500, DAX, NYSE, NIFTY 50), demonstrating consistent cross-market robustness across metrics under the stated chronological evaluation protocol.}
    \item \textbf{Eliminating overfitting from ad hoc hyperparameter tuning:} we adopt a fixed hyperparameter configuration pre-validated across indices to eliminate tuning overhead and improve robustness to short-term volatility. The model uses a 20-day lookback window, Adam optimizer (lr=$10^{-4}$), $\ell_2$ regularization ($10^{-4}$), and dropout (0.3) to ensure generalization without sacrificing accuracy. With only 675k parameters, 1.5 ms/sample latency, and under 2-minute training per index, VertiFuseX is lightweight and deployment-ready.
    \item \textbf{Performance Analysis:} VertiFuseX consistently matches or outperforms both standard baselines and seven state-of-the-art models, achieving up to 54.3\% lower MAPE and over 40\% reductions in MAE and RMSE relative to LSTM, Bi-LSTM, and St-LSTM baselines and up to 34.6\% lower MAE, 36.4\% lower RMSE, and 27.2\% lower MAPE compared to state-of-the-art methods across evaluated datasets.
    \item Economic validation through algorithmic trading simulation and stress-testing under extreme regimes (COVID-19 crash and 2022 bear market) confirms reduced maximum drawdowns and superior risk-adjusted returns relative to passive and momentum benchmarks. 
 \end{enumerate}

The rest of the paper is structured in the following manner: Section 2 details related work, and Section 3 explains VertiFuseX’s architecture, emphasizing vertical fusion mechanics. Section 4 presents the experimental environment and result analysis with baseline and state-of-the-art models. Section 5 discusses the robustness analysis and the limitations of the proposed model. Finally, Section 6 presents the conclusion and directions for future work.

\section{Related Work}
Recent DL advancements have shaped stock price prediction, but existing models face several limitations that are addressed by VertiFuseX. To improve clarity, we organize prior work conceptually by temporal modeling strategy, namely single-scale recurrent models, bidirectional extensions, stacked or hierarchical temporal models, and fusion-based approaches while retaining the broad categorization into DL and hybrid DL models for readability.

\subsection{Deep Learning based Models}
This category includes comprehensive end-to-end neural network architectures. In DL models, previous research can generally be categorized into single-scale recurrent architectures, bidirectional extensions for enhanced contextual coverage, and stacked or hierarchical versions that improve representational depth.
Early DL approaches applied Multilayer Perceptrons (MLPs), Recurrent Neural Networks (RNNs), and LSTM networks to stock prediction \cite{hiransha2018nse} on the NSE and NYSE indices. While these methods demonstrated improvements over ARIMA, they lacked comparisons with more advanced DL baselines. \cite{jain2018stock} proposed a Conv1D-LSTM hybrid for short-term forecasting on TCS and MRF stocks, which, although effective, required extensive per-stock hyperparameter tuning, severely limiting scalability. Similarly, \cite{lu2020cnn} developed a CNN-LSTM model for the Shanghai Composite index, but its generalization across diverse markets remains untested. \cite{rather2021lstm} combined LSTM with a DNN and an autoregressive moving pointer model for NIFTY-50 prediction. The evaluation’s focus on a single index restricted broader applicability. \cite{kanwal2022bicudnnlstm} introduced BiCuDNNLSTM, a bidirectional LSTM enhanced with a GPU-optimized DNN layer and 1D-CNN, for DAX and HSI indices. However, its reliance on computationally intensive hyperparameter tuning resulted in inconsistent performance. 
Transformer-based models such as \cite{wang2022stock}, which utilized an encoder-decoder with multi-head attention, showed promise in capturing long-range dependencies. However, these models often have higher parameter counts and computational demands, and their performance usually improves with larger training datasets and thorough regularization. Moreover, the evaluation in \cite{wang2022stock} was benchmarked only against basic deep learning baselines, limiting the depth of the comparative analysis.
\cite{gupta2022stocknet} introduced StockNet, a GRU-based model with Injection and Investigation modules, which depended on highly correlated stocks and struggled with irregular time series due to data augmentation issues. \cite{lu2021cnn} proposed a CNN-BiLSTM model with attention mechanisms for CSI300, yet evaluation was restricted to a single market, limiting claims of cross-market generalization. More recently, \cite{prakash2024stock} introduced a deep attention Bi-LSTM for stock prediction, enhancing bidirectional modeling but still reliant on single-scale temporal resolution, which overlooks multi-horizon dependencies.

From a temporal modeling perspective, most existing DL approaches operate at a single implicit time scale determined by the input window and network depth. Standard LSTM and GRU models primarily capture short- to mid-term dependencies but struggle to disentangle overlapping temporal patterns such as daily noise and longer market cycles. Bidirectional variants improve contextual awareness by incorporating reverse-time information, yet still rely on a fixed temporal resolution. CNN-LSTM and attention-based hybrids enhance local pattern extraction or weighting of historical steps, but they do not explicitly separate temporal dynamics across multiple horizons. As a result, these models often emphasize either recent volatility or long-term trends, but not both in a structured and simultaneous manner.
\begin{table}[htbp]
\centering
\small
\caption{Comparative summary of DL and hybrid models for stock price prediction}
\label{tab:comparative_summary}
\resizebox{\textwidth}{!}{
\begin{tabular}{m{1.5cm}m{2cm}m{2cm}m{2cm}m{2cm}m{2cm}}
\toprule
\textbf{Model / Reference} & \textbf{Architecture Type} & \textbf{Temporal Modeling Strategy} & \textbf{Fusion Method} & \textbf{Evaluation Scope} & \textbf{Key Limitations} \\
\midrule
LSTM  & Recurrent & Single-scale, unidirectional & None & Single / few indices & Overfits recent noise, limited long-range abstraction \\
Bi-LSTM & Recurrent & Single-scale, bidirectional & None & Single / few indices & Fixed temporal resolution, no hierarchy \\
St-LSTM & Recurrent (stacked) & Implicit multi-depth & None & Limited indices & Depth $\neq$ explicit multi-scale separation \\
CNN--LSTM & Hybrid & Local + sequential & Early / static fusion & Single market & Static fusion dilutes temporal signals \\
GA-CNN-LSTM & Hybrid + evolutionary  & Single-scale & Late fusion & Single index & Overfitting, high tuning cost \\
DE-ABC-Bi-LSTM-ARIMA & Hybrid ensemble & Single-scale & Decision-level ensemble & Limited markets & High complexity, poor scalability \\
ModAugNet & Hybrid & Single-scale & Task-level fusion & Single index & Heavy tuning, limited generalization \\
Reservoir Computing  & Recurrent & Fixed temporal dynamics & None & Multiple indices & Limited adaptability across regimes \\
VertiFuseX (Ours) & Multi-stream LSTM & Explicit multi-scale (short, mid, long) & Penultimate-layer vertical fusion & 10 indices in total covering 3 continents & --- \\
\bottomrule
\end{tabular}}
\color{black} 
\end{table}

\subsection{Hybrid Deep Learning based Models}
Hybrid models typically combine recurrent architectures with convolutional, or ensemble-based components, differing primarily in how temporal features are fused or optimized rather than in their underlying temporal scale.
Hybrid models aim to combine complementary architectures for improved performance. \cite{baek2018modaugnet} presented ModAugNet, integrating data augmentation with LSTM modules. The model exhibited strong reliance on correlated stocks and required extensive trial-and-error tuning to optimize LSTM module parameters, leading to limited robustness.
\cite{zhang2019stock} developed a Generative Adversarial Network (GAN) with an LSTM generator and MLP discriminator for S\&P 500 prediction. However, limited evaluation across market conditions weakened its generalization claims. \cite{baek2023cnn} enhanced CNN-LSTM with genetic algorithm (GA) optimization for KOSPI, but manual adjustments to mutation rates hindered practical deployment. \cite{kumar2022three} proposed a three-stage fusion model combining Bi-LSTM-ARIMA and an Artificial Bee Colony algorithm with Differential Evolution (DE-ABC-Bi-LSTM-ARIMA). Its small sliding window amplified noise sensitivity, limiting its capacity to detect long-term market trends effectively. \cite{lee2020stock} proposed NuNet for handling high-dimensional data, but max-pooling and dataset randomization introduced biases, overlooking individual stock dynamics. \cite{wang2021stock} applied a reservoir computing (RC) based approach using random, scale-free, and small-world networks. Despite computational efficiency, RC models showed inconsistent performance across datasets due to fixed reservoir structures that lack adaptability to varying market regimes. Recent hybrids include \cite{zhang2025novel}, who fused GRU with a weighted fuzzy candlestick model, improving error-based forecasting but suffering from static integration that dilutes temporal signals. Gul \cite{gul2025novel} proposed an ensemble combining bagging, boosting, dagging, and stacking, offering robust predictions yet prone to overfitting without sparsity priors. \cite{li2024mlbgk} developed MLBGK, a feature fusion model that blends multiple streams, aligning with multi-scale needs but relying on ad hoc tuning, which limits cross-market scalability.

Although hybrid and ensemble models combine multiple architectures, their temporal modeling remains largely single-scale, as fused components typically share the same input resolution and forecasting horizon. Stacked or ensemble designs increase representational capacity but do not guarantee separation of short, medium, and long range temporal dependencies, often leading to redundancy or overfitting. Consequently, multi-scale temporal abstraction is treated implicitly rather than as a structured architectural objective.
Taken together, prior DL and hybrid forecasting models exhibit three recurring structural patterns that limit their robustness and generalizability, including reliance on decision-level fusion, extensive hyperparameter tuning, and single-scale temporal modeling. To the best of our knowledge, no prior work in computational economics or financial forecasting has explored penultimate-layer vertical fusion of heterogeneous temporal representations as an alternative to decision-level ensemble or econometric forecast combination methods.

Table \ref{tab:comparative_summary} provides a comparative summary of representative models, capturing key architectural characteristics, temporal modeling strategies, and evaluation scopes discussed above. By highlighting their limitations, the table provides context for the design motivation behind VertiFuseX, presenting illustrative examples rather than a comprehensive listing of all reviewed approaches. These recurring limitations motivate the design of VertiFuseX, a hybrid LSTM framework that leverages penultimate-layer vertical fusion and multi-scale abstraction to overcome the architectural, temporal, and generalization shortcomings identified in prior work. The following section details the proposed methodology and architectural innovations of VertiFuseX.

\section{VertiFuseX: Proposed Methodology}

In this section, we present our proposed scheme, VertiFuseX, a hybrid deep learning model for financial time-series forecasting, leveraging penultimate-layer vertical fusion and multi-scale abstraction. VertiFuseX integrates LSTM, Bi-LSTM, and St-LSTM by vertically stacking (i.e., concatenating along the feature dimension) their penultimate-layer latent representations prior to task-specific regression.

\begin{figure*}[htp]
  \centering
  {\includegraphics[width=1.0\textwidth]{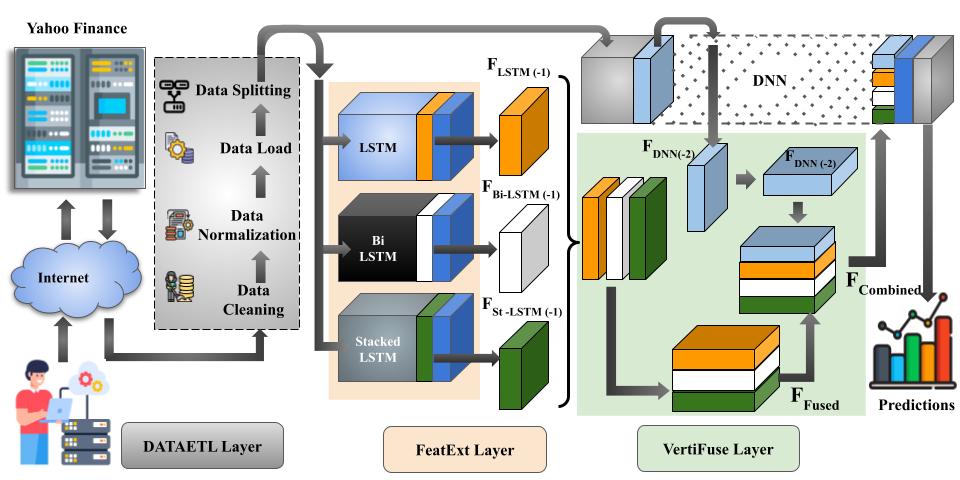}}
   \caption{VertiFuseX architecture illustrating (i) multi-stream temporal feature extraction via LSTM, Bi-LSTM, and St-LSTM, (ii) penultimate-layer vertical fusion, and (iii) integration with a parallel DNN stream prior to prediction.}
   \Description{}
 \label{fig:ProposedModel}
\end{figure*}

The architecture comprises of three layers as shown in Fig.~\ref{fig:ProposedModel}, which are DATAETL (Data Extraction, Transformation, Load), FeatExt (Feature Extraction), and the VertiFuse layer (Vertical Fusion), which illustrates data flow from input processing to feature extraction and fusion. Each layer contributes distinct technical functionalities, which collectively enhance prediction accuracy and model robustness. 

VertiFuseX operates in three algorithmic stages as formalized in Algorithms~\ref{alg:VertiFuseX-1} and~\ref{alg:VertiFuseX-2}. The DATAETL procedure preprocesses historical stock prices, the FeatExt procedure defines penultimate-layer mappings for temporal abstractions (LSTM, Bi-LSTM, St-LSTM) and the complementary non-temporal DNN stream, and the VertiFuse procedure performs joint vertical fusion with end-to-end optimization for final prediction.

\subsection{DATAETL Layer}
The DATAETL layer serves as the foundational preprocessing block for VertiFuseX, ensuring the temporal and structural fidelity of financial time-series data prior to learning. Stock market data, inherently irregular due to holidays and trading suspensions, demands rigorous preprocessing to prevent temporal leakage and maintain sequential consistency.

All time series are indexed in trading time rather than calendar time to preserve signal fidelity and prevent temporal leakage. Non-trading days arising from weekends, exchange holidays, or market suspensions are removed entirely, rather than being filled via forward-fill, backward-fill, or interpolation. This design ensures that each transition processed by the recurrent models corresponds to a valid market event, preventing the introduction of artificial plateaus or spurious low-volatility patterns.
Importantly, the identical training and testing sample sizes reported for the S\&P 500, DJIA, NYSE, and NASDAQ arise because these indices share the same U.S. trading calendar. After removing non-trading days and applying an identical chronological split, their valid trading-day counts align exactly (3388 training days and 365 testing days) as shown in Table \ref{tab:Indices Used}. This alignment is intentional and enables a controlled baseline comparison across architectures.
In contrast, global indices used for state-of-the-art comparisons are evaluated over their original study-specific periods and therefore exhibit differing data lengths, reflecting market-specific calendars rather than any forced alignment.
To mitigate lookahead bias, the moving window strategy ensures that only past data is used for prediction, simulating real-world temporal causality. The DATAETL layer (Algorithm~\ref{alg:VertiFuseX-1}, lines 1-7) handles data extraction, transformation, and loading.
Historical OHLC (Open-High-Low-Close) prices are obtained from Yahoo Finance for various global indices, including the Standard \& Poor’s 500 Index (S\&P 500), New York Stock Exchange (NYSE), Dow Jones Industrial Average (DJIA), and the National Association of Securities Dealers Automated Quotations (NASDAQ). The data covers the period from January 1, 2010, to December 31, 2024 (see Table~\ref{tab:Indices Used}). Yahoo Finance is chosen for its comprehensive historical records. 

\begin{algorithm}[ht]
\caption{VertiFuseX: Vertical Fusion for Generalizable Financial Forecasting (Part 1: Data Preparation \& Feature Extraction)}
\label{alg:VertiFuseX-1}
\begin{algorithmic}[1]
\Require Historical univariate closing price series $P = \{p_1, \dots, p_T\}$, window size $w = 20$, shared dimensionality $d_T = 64$, fixed hyperparameters (Table~\ref{tab:initial_params})
\Ensure Out-of-sample error metrics (MAE, RMSE, MAPE) on chronological held-out test set

\Procedure{DATAETL}{start\_date, end\_date, index}
    \State $P \gets \text{ExtractClosingPrices}(start\_date, end\_date, index)$ \Comment{Yahoo Finance, trading days only}
     \State {$\{\mathbf{x}^{\mathrm{raw}}_t\}_{t=w}^{T-1}, \{y^{\mathrm{raw}}_t\}_{t=w}^{T-1} \gets \text{SlidingWindow}(P, w)$} \Comment{{Raw chronological windows, $\mathbf{x}_t \in \mathbb{R}^{w \times 1}$, $y_t = p_{t+1}$}}
    \State {$D^{\mathrm{raw}}_{\text{train}}, D^{\mathrm{raw}}_{\text{val}}, D^{\mathrm{raw}}_{\text{test}} \gets \text{ChronologicalSplit}(\{\mathbf{x}^{\mathrm{raw}}_t, y^{\mathrm{raw}}_t\})$}
    \State {$N_{\min}, N_{\max} \gets \text{FitMinMax}(D^{\mathrm{raw}}_{\text{train}})$} \Comment{{Training partition only}}
    \State {$D_{\text{train}}, D_{\text{val}}, D_{\text{test}} \gets \text{ApplyMinMax}(D^{\mathrm{raw}}_{\text{train}}, D^{\mathrm{raw}}_{\text{val}}, D^{\mathrm{raw}}_{\text{test}}, N_{\min}, N_{\max})$} \Comment{{Eq.~\eqref{EquNorm}, train scaler}}
    \Statex \hspace{\algorithmicindent}// Train: first $\approx 3388$ trading days ($\sim$14 years), Val: final 10\% of train, Test: final 365 trading days (strict OOS, no leakage)
    \State \textbf{return} $D_{\text{train}} = (X_{\text{train}}, Y_{\text{train}})$, $D_{\text{val}} = (X_{\text{val}}, Y_{\text{val}})$, $D_{\text{test}} = (X_{\text{test}}, Y_{\text{test}})$
\EndProcedure

\Procedure{FeatExt}{}
    \State Define parameterized mappings with penultimate-layer outputs fixed to $\mathbb{R}^{d_T}$:
    \State \hspace{\algorithmicindent}$\phi_{\text{LSTM}}(\cdot; \theta_{\text{LSTM}}) : \mathbb{R}^{w \times 1} \to \mathbb{R}^{d_T}$ \Comment{2-layer LSTM, final hidden state before regression head}
    \State \hspace{\algorithmicindent}$\phi_{\text{Bi-LSTM}}(\cdot; \theta_{\text{Bi-LSTM}}) : \mathbb{R}^{w \times 1} \to \mathbb{R}^{d_T}$ \Comment{2-layer bidirectional, concatenated final states}
    \State \hspace{\algorithmicindent}$\phi_{\text{St-LSTM}}(\cdot; \theta_{\text{St-LSTM}}) : \mathbb{R}^{w \times 1} \to \mathbb{R}^{d_T}$ \Comment{3-layer stacked LSTM, final hidden state}
    \State \hspace{\algorithmicindent}$\phi_{\text{DNN}}(\cdot; \theta_{\text{DNN}}) : \mathbb{R}^{w \times 1} \to \mathbb{R}^{d_T}$ \Comment{MLP: flatten $\to 128 \to 64$ units, ReLU, dropout (0.3); penultimate output aligned to $d_T$}
    \Statex \hspace{\algorithmicindent}// All mappings include dropout (0.3) and batch normalization where applicable; architectures per Fig.~\ref{fig:fusemodel}
    \State \textbf{return} $\phi_{\text{LSTM}}, \phi_{\text{Bi-LSTM}}, \phi_{\text{St-LSTM}}, \phi_{\text{DNN}}$
\EndProcedure
\end{algorithmic}
\end{algorithm}

To ensure uniformity of data, Min-Max normalization is used.  Min-Max Normalization in Eq. \eqref{EquNorm} is used to scale input features to a [0,1] range. This step promotes efficient convergence and prevents dominance of larger numeric ranges.
\begin{equation} \label{EquNorm}
    \bar{N} = \frac{N - N_{min}}{N_{max} - N_{min}}
\end{equation}
where $N$ is the original feature value, and {$N_{\max}$ and $N_{\min}$ are the maximum and minimum values computed only on the training partition. The scaler is fit on $D_{\text{train}}$ and the resulting $N_{\min}$ and $N_{\max}$ are then applied to transform the held-out test data. No test-set statistic is used at any stage of fitting, so the normalization introduces no information leakage from future observations into training. Normalization is therefore performed after the chronological train and test split (Algorithm~\ref{alg:VertiFuseX-1}), not on the full series.}

\begin{figure}[ht]
  \centering
  \includegraphics[width=0.7\columnwidth]{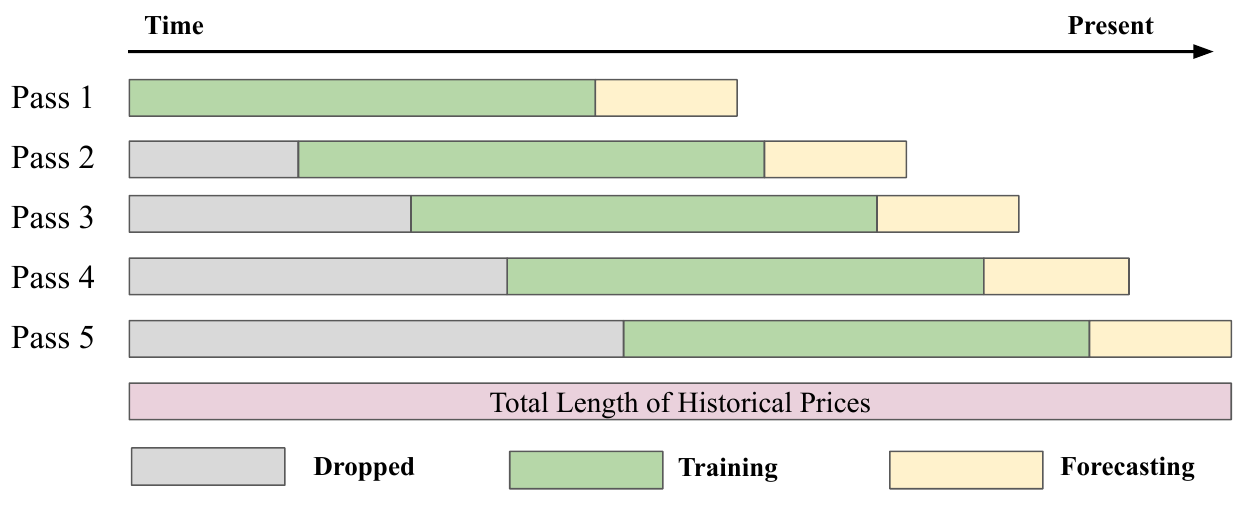}
   \caption{Illustration of the moving window strategy. As the window advances, earlier data is dropped (grey) while new segments are included for training (green) and forecasting (yellow), enabling dynamic evaluation over time.}
   \Description{}
 \label{fig:window}
\end{figure}

VertiFuseX is implemented in a strictly univariate configuration ($f = 1$), using only the raw closing price as model input. Although the source data include OHLC fields, only closing prices are used. No auxiliary price components or derived technical indicators enter the model.
This design choice serves two purposes. First, it eliminates the risk of lookahead bias that can arise when indicators are computed using information outside the active forecasting window. Second, it isolates the contribution of the proposed vertical fusion architecture, demonstrating its ability to learn complex temporal dynamics directly from raw price history without reliance on manual feature engineering.
{The raw closing-price series is first converted into 20-day chronological moving windows (Fig.~\ref{fig:window}), where each input window contains only past observations relative to the prediction target. The resulting raw windows are then split chronologically, after which the Min-Max scaler fitted on the training partition is applied to the train, validation, and test windows.}

This approach balances short-term responsiveness with long-term trend capture. As the window advances, older data is discarded while new data is incorporated for dynamic forecasting.
To simulate a realistic deployment scenario in which future market data is unavailable, we employ a static chronological out-of-sample evaluation,  in line with established practices in ML-based financial forecasting \citep{gu2020empirical}. The dataset is split once at a fixed temporal cutoff into two non-overlapping partitions. The training set consists of the first 3,388 trading days (approximately 14 years), and the held-out test set consists of the final 365 trading days (1 year). The model is trained exclusively on the training set, after which all parameters are frozen. During the test phase, no retraining, fine-tuning, or rolling re-estimation is performed. For each test day $t$, predictions are generated using a fixed sliding input window of length $w = 20$ days, containing only observations prior to $t$. Table \ref{tab:Indices Used} summarizes the indices, symbols, and data splits.

{Algorithm~\ref{alg:VertiFuseX-1} receives the start date, end date, and index identifier, extracts the chronological closing-price series, constructs raw sliding windows, and then performs the chronological train, validation, and test split before normalization. The Min-Max scaler is fitted only on the raw training partition and is then applied to the validation and test partitions. This ordering ensures that no statistic from the held-out test period influences preprocessing. Table~\ref{tab:Indices Used} summarizes the indices, symbols, and data splits. The identical sample counts for S\&P~500, DJIA, NYSE, and NASDAQ arise because these four U.S. baseline indices share the same trading calendar. Global indices used for state-of-the-art comparisons are evaluated over their corresponding benchmark periods and are not forced into identical calendar lengths.}

\begin{table*}[ht]
\centering
\caption{Indices under study}
\label{tab:Indices Used}
\resizebox{\textwidth}{!}{
\begin{tabular}{ l l c c c c c }
\hline
\textbf{Name of Indices} & \textbf{Symbol}   & \textbf{Duration}       &  \textbf{Total Data }     & \textbf{Data after Windowing} & \textbf{Train data} & \textbf{Test Data} \\ 
\hline
 
S\&P 500  & \textasciicircum{}GSPC  & 01-01-2010 - 31-12-2024 & 3773 & 3753 & 3388 & 365 \\ 
DJI & \textasciicircum{}DJI  & 01-01-2010 - 31-12-2024 & 3773 & 3753 & 3388 & 365 \\ 
NYSE & \textasciicircum{}NYA  & 01-01-2010 - 31-12-2024 & 3773 & 3753 & 3388 & 365 \\
NASDAQ & \textasciicircum{}IXIC  & 01-01-2010 - 31-12-2024 & 3773 & 3753       & 3388 & 365       
\\
\hline
\end{tabular}
}
\end{table*}

\begin{figure*}[ht]
    \centering
    \subfigure[S \& P500 Index]{\includegraphics[width=0.48\textwidth]{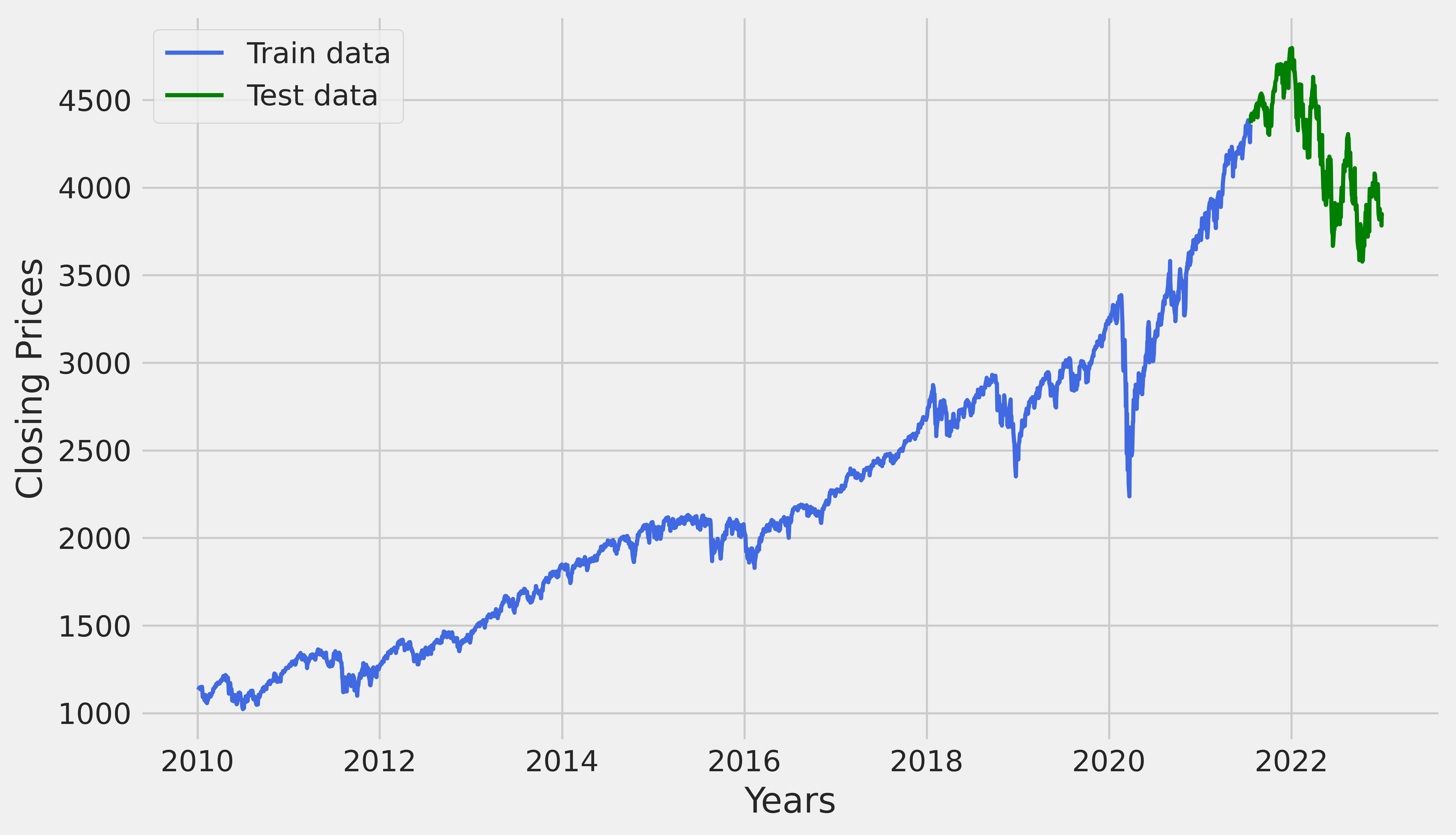}} 
     \subfigure[DJI]{\includegraphics[width=0.48\textwidth]{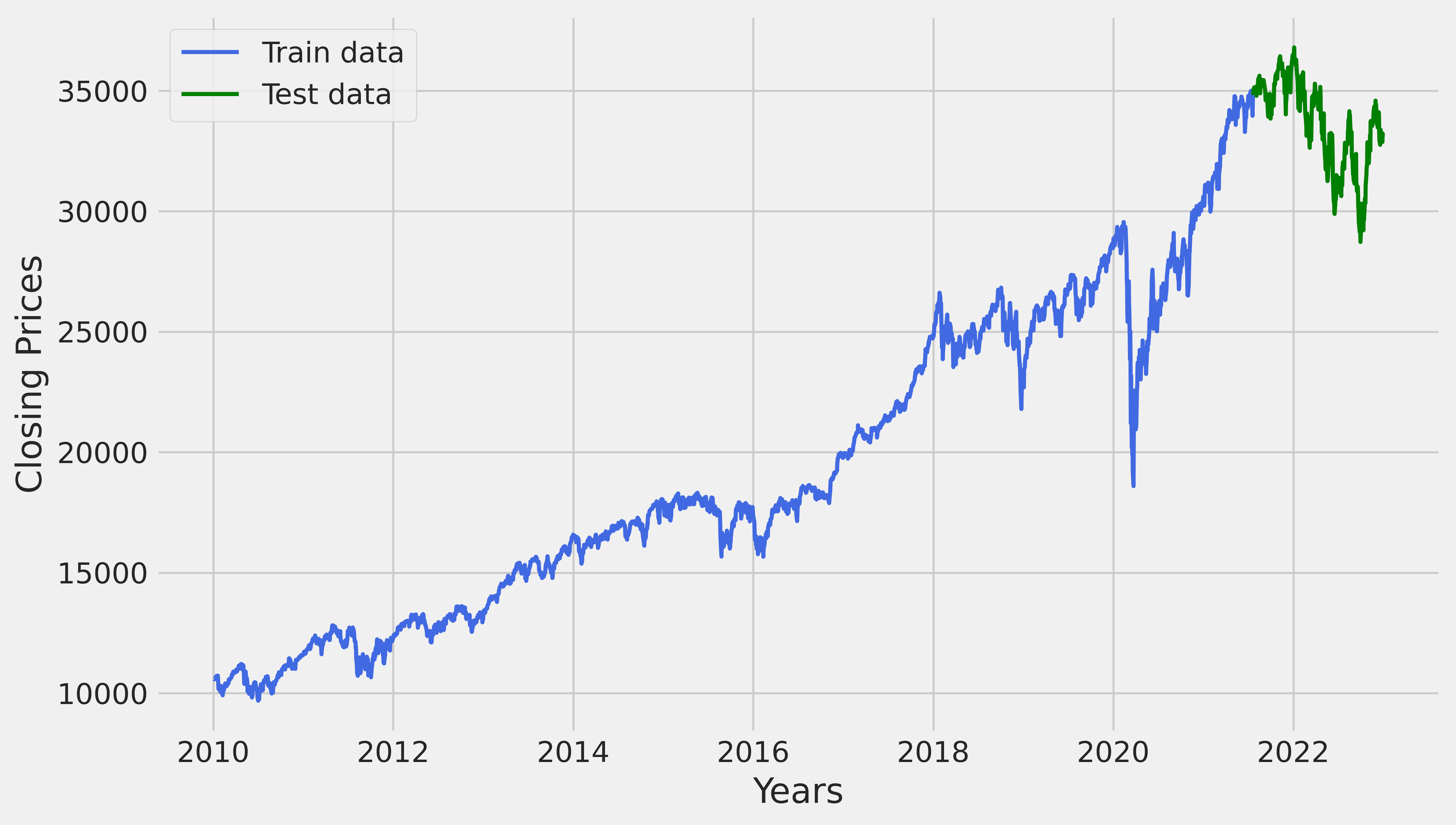}} 
    \subfigure[NYSE]{\includegraphics[width=0.48\textwidth]{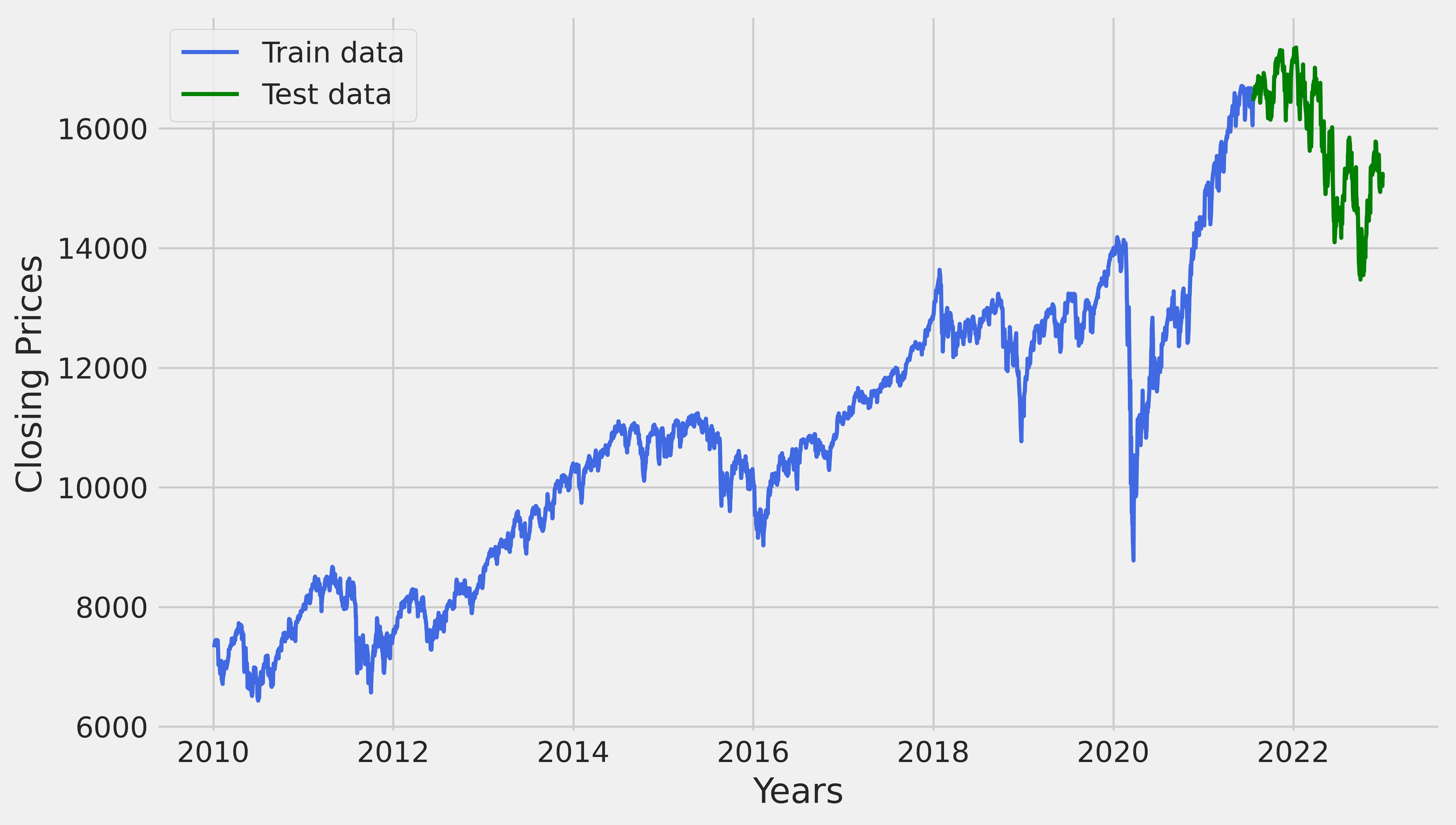}} 
    \subfigure[NASDAQ]{\includegraphics[width=0.48\textwidth]{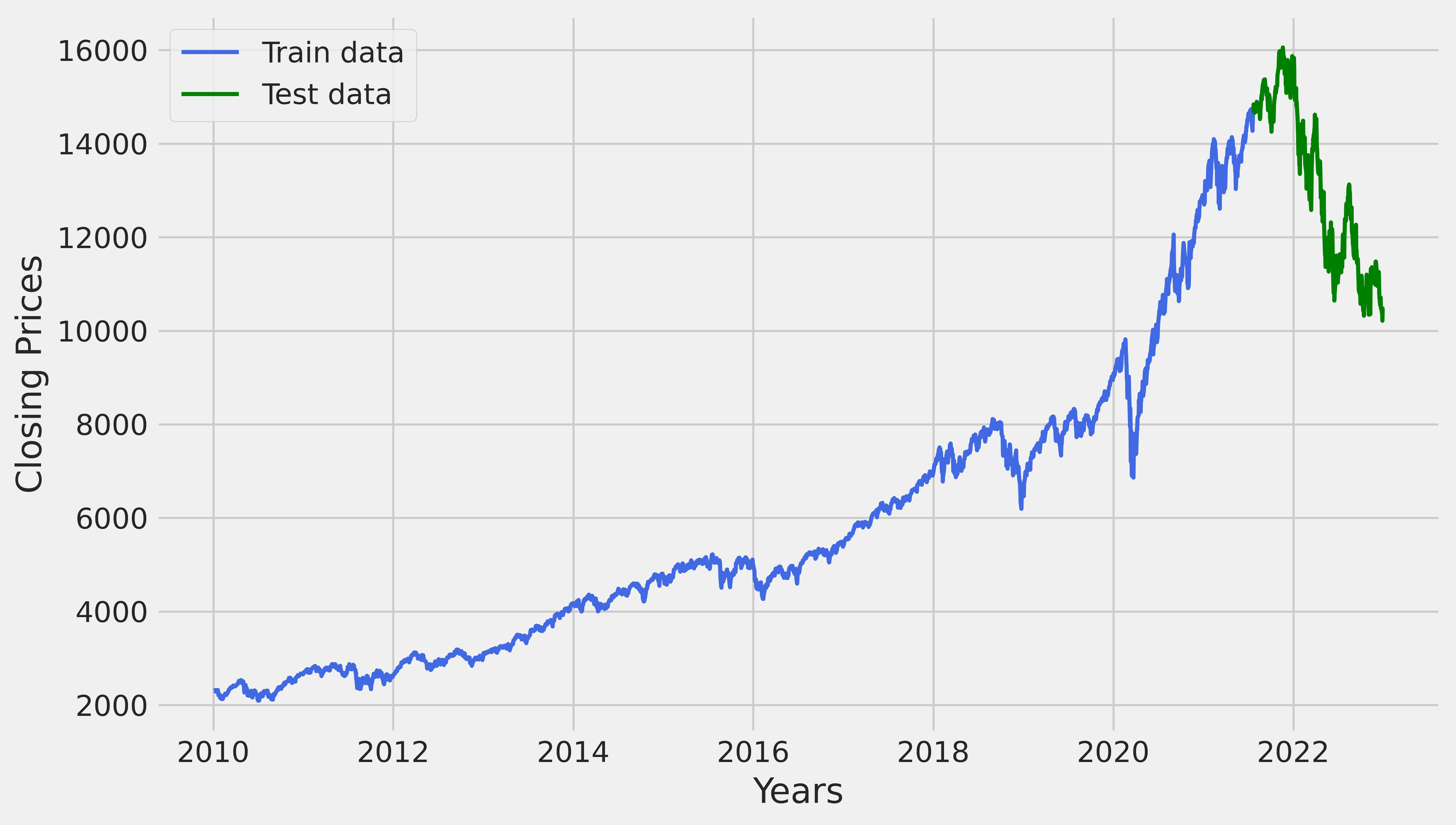}} 
    \caption{The training set and testing set are denoted by the blue and green lines for utilized indices. The temporal dimension is represented by the x-axis, denoting the duration in years, whereas the y-axis portrays the stock index's closing prices.}
    \Description{}
    \label{fig: Train_test Split}
\end{figure*}

\subsection{FeatExt Layer}
\begin{figure*}[ht]
  \centering
  \includegraphics[width=\textwidth]{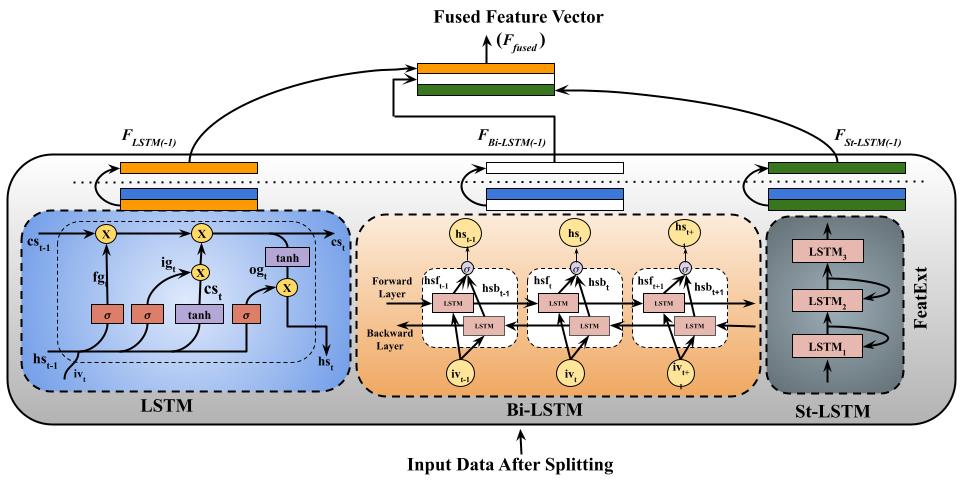}
   \caption{Feature extraction from penultimate layers (highlighted) are used for fusion to preserve intermediate temporal representations before task-specific compression.}
   \Description{}
 \label{fig:fusemodel}
\end{figure*}

The FeatExt layer (Algorithm \ref{alg:VertiFuseX-1}, lines 8--15) extracts temporal features using LSTM, Bi-LSTM, and St-LSTM to extract temporal features, each capturing distinct dynamics. Fig. \ref{fig:fusemodel} shows feature extraction from the penultimate layers of each model. Each model is selected for its complementary capacity to capture different temporal patterns, namely LSTM for sequential memory, Bi-LSTM for bidirectional dependencies, and St-LSTM for multi-scale abstraction. The branch set was chosen to place three recurrent inductive biases under a common fusion interface. The LSTM provides ordered sequential memory, the Bi-LSTM encodes context in both directions within an already observed window, and the St-LSTM builds progressively deeper temporal abstractions. This keeps the model compact and makes the fusion hypothesis testable without adding an attention mechanism. Because the fusion layer accepts any encoder projected into the shared latent space, the same interface extends to Transformer-family branches for longer contexts or multivariate settings. In Algorithm \ref{alg:VertiFuseX-1}, this process is depicted as procedure $FEATEXT(X_{train}, Y_{train})$.  There are three sub-processes which are $FitLSTM(X_{train}, Y_{train})$, $FitBiLSTM(X_{train}, Y_{train})$, and $FitStLSTM(X_{train}, Y_{train})$ in line 10,11, and 12 respectively. To ensure clarity and maintain notational consistency, Table \ref{tab:lstmgates} summarizes the variables used in the gating mechanisms across all LSTM-based architectures. Each notation is introduced along with its specific function across the LSTM, Bi-LSTM, and St-LSTM architectures.
 
\begin{table}[ht]
\centering
\caption{List of variables used in LSTM, Bi-LSTM, and St-LSTM Models}
\label{tab:lstmgates}
\begin{tabular}{lp{3.7cm}|lp{3.7cm}}
\toprule
\textbf{Symbol} & \textbf{Description} & \textbf{Symbol} & \textbf{Description} \\
\midrule
$iv_t$ & Input vector at time step~$t$ (closing price in the univariate setting) 
& $hs_t$ & LSTM hidden output at time~$t$ \\[2pt]

$h^{sf}_t,\;h^{sb}_t$ & Forward and backward directional hidden states 
& $h^s_t$ & Bi-LSTM output $h^{sf}_t \Vert h^{sb}_t$ \\[2pt]

$h^{s,l}_t$ & Hidden output of St-LSTM layer~$l$ 
& $cs_t$ & LSTM memory cell at time~$t$ \\[2pt]

$c^{sf}_t,\;c^{sb}_t$ & Forward and backward directional cell states 
& $c^{s,l}_t$ & Memory cell of St-LSTM layer~$l$ \\[2pt]

$\tilde{c}_t$ & Candidate update in LSTM 
& $\tilde{c}^{sf}_t,\;\tilde{c}^{sb}_t$ & Direction-specific candidate updates \\[2pt]

$\tilde{c}^{s,l}_t$ & Layer-specific candidate update 
& $ig_t,\;fg_t,\;og_t$ & Layer-specific input, forget, and output gates (layer index $l$ is implied by the weights) \\[2pt]

$W_{*}^{(h)},\;W_{*}^{(x)}$ & Hidden-to-gate and input-to-gate matrices 
& $W_{*}^{(l,h)},\;W_{*}^{(l,r)}$ & Inter-layer and temporal-recurrent weights for layer~$l$ \\[2pt]

$b_{*},\;b_{*}^l$ & Bias parameters (global or layer-specific) 
& & \\ 
\bottomrule
\end{tabular}
\end{table}

The LSTM branch leverages gating mechanisms to selectively retain or discard information over extended temporal horizons, thereby mitigating vanishing gradient effects and capturing long-range dependencies. To produce a compact yet expressive feature representation suitable for vertical fusion, the branch is configured with two stacked LSTM layers, each assigned a distinct functional role. The first layer, comprising 128 units with \texttt{return\_sequences=True}, preserves the full temporal resolution of the input sequence, yielding a sequence of hidden states in $\mathbb{R}^{w \times 128}$ that encode fine-grained, step-wise dynamics within the observation window. This output is then fed into a second layer containing 64 units with \texttt{return\_sequences=False}, which functions not as a predictor but as a temporal encoder: it collapses the entire sequence into a single fixed-length latent vector, specifically the final hidden state $\mathbf{F}_{\text{LSTM}}^{(-1)} \in \mathbb{R}^{64}$. This 64-dimensional vector constitutes the penultimate representation used in subsequent fusion stages. Both layers are regularized via batch normalization and dropout (rate = 0.3) applied to their outputs, enhancing generalization and reducing co-adaptation among units (see Fig.~\ref{fig:fusemodel}).

It employs gating mechanisms to retain long-term dependencies: Each cell contains an input gate \((ig_t)\), forget gate \((fg_t)\), output gate \((og_t)\), cell state \((cs_t)\), and hidden state \((hs_t)\). The \((ig_t)\) activation in  Eq. \eqref{eq:lstm1} determines the proportion of new data ($iv_t$) integrated into the cell state, combining the current input and prior hidden state $hs_{t-1}$:

\begin{equation}  \label{eq:lstm1}
    ig_t = \sigma\!\bigl(W_{ig}^{(h)}\,hs_{t-1} + W_{ig}^{(x)}\,iv_t + b_{ig}\bigr),
\end{equation}
where \(W_{ig}^{(h)}\) and \(W_{ig}^{(x)}\) are the input-gate weights for the hidden and input signals, and \(b_{ig}\) is the bias term.

The \((fg_t)\) operation in Eq. \eqref{eq:lstm2} computes a discard factor for outdated \((cs_{t-1})\):
\begin{equation}  \label{eq:lstm2}
   fg_t = \sigma\!\bigl(W_{fg}^{(h)}\,hs_{t-1} + W_{fg}^{(x)}\,iv_t + b_{fg}\bigr).
\end{equation}

The cell state update in Eq.~\eqref{eq:lstm3} and \eqref{eq:lstm4} begins with the generation of a candidate memory vector \((\tilde{c}_t)\) using a hyperbolic tangent activation:
\begin{equation}  \label{eq:lstm3}
    \tilde{c}_t = \tanh\!\bigl(W_{c}^{(h)}\,hs_{t-1}+W_{c}^{(x)}\,iv_t+b_c\bigr)
\end{equation}
The updated cell state \(cs_t\) merges retained history
\(\bigl(fg_t\odot cs_{t-1}\bigr)\) and new inputs \(\bigl(ig_t\odot\tilde{c}_t\bigr)\):
\begin{equation}
\label{eq:lstm4}
    cs_t = fg_t\odot cs_{t-1} + ig_t\odot\tilde{c}_t.
\end{equation}
The \((og_t)\) modulation in (Eq. \eqref{eq:lstm5}, \eqref{eq:lstm6}) regulates the \((hs_t)\) using a sigmoid-activated \((og_t)\):
\begin{equation} \label{eq:lstm5}
     og_t = \sigma\!\bigl(W_{og}^{(h)}\,hs_{t-1} + W_{og}^{(x)}\,iv_t + b_{og}\bigr)
\end{equation}
The final \((hs_t)\) is scaled by the normalized \((cs_t)\): 
\begin{equation}  \label{eq:lstm6}
    hs_t = og_t \odot \tanh(cs_t)
\end{equation}

The Bi-LSTM enhances temporal modeling by processing input sequences in both forward (past-to-future) and backward (future-to-past) directions. This dual analysis captures contextual dependencies from historical and future states, outperforming unidirectional LSTMs in complex time-series forecasting tasks. The Bi-LSTM branch utilises bidirectional gating mechanisms to selectively retain or discard information over extended temporal horizons, thereby incorporating both past and future context while mitigating vanishing gradient effects and capturing long-range bidirectional dependencies. To produce a compact yet expressive feature representation suitable for vertical fusion, the branch is configured with two stacked bidirectional LSTM layers, each assigned a distinct functional role. The first layer, comprising 128 units per direction with \texttt{return\_sequences=True}, preserves the full temporal resolution of the input sequence, yielding concatenated sequences of hidden states in $\mathbb{R}^{w \times 256}$ that encode fine-grained, bidirectional step-wise dynamics within the observation window. This output is then fed into a second layer containing 32 units per direction with \texttt{return\_sequences=False}, which functions not as a predictor but as a temporal encoder which collapses the entire sequence into a single fixed-length latent vector, specifically the concatenated final hidden states $\mathbf{F}_{\text{Bi-LSTM}}^{(-1)} \in \mathbb{R}^{64}$. This 64-dimensional vector constitutes the penultimate representation used in subsequent fusion stages. Both layers are regularized via batch normalization and dropout (rate = 0.3) applied to their outputs, enhancing generalization and reducing co-adaptation among units (see Fig.~\ref{fig:fusemodel}).

{We note that bidirectional processing introduces no look-ahead bias. The reverse pass operates only over observations within the lookback window $[t-w,\dots,t-1]$, all of which precede the forecast origin $t$, and it never accesses any observation at or beyond $t$. Since the target is $t+1$, every index $\tau$ used by any branch satisfies $\tau \le t-1 < t+1$, so no information from the prediction horizon enters the model.}

The forward LSTM layer computes hidden states $h^{sf}_t$ through sequential gating mechanisms. In which the $ig_t$ activation in Eq. \eqref{eq:bilstm1}, determines new data retention using sigmoid-activated transformations of the prior hidden state $h^{sf}_{t-1}$ and current input $iv_t$:
\begin{equation}
ig_t = \sigma\!\bigl(W_{ig}^{(h)}\,h^{sf}_{t-1} + W_{ig}^{(x)}\,iv_t + b_{ig}\bigr)
\label{eq:bilstm1}
\end{equation}
The \((fg_t)\) operation in Eq. \eqref{eq:bilstm2}, identifies obsolete information for removal from the cell state \( c^{sf}_{t-1} \):
\begin{equation}
 fg_t = \sigma\!\bigl(W_{fg}^{(h)}\,h^{sf}_{t-1} + W_{fg}^{(x)}\,iv_t + b_{fg}\bigr)
\label{eq:bilstm2}
\end{equation}
The candidate cell state \( \tilde{c}^{sf}_t \) in Eq.~\eqref{eq:bilstm3}, produces a temporally modulated input vector via hyperbolic tangent:
\begin{equation}
\tilde{c}^{sf}_t = \tanh\!\bigl(W_{c}^{(h)}\,h^{sf}_{t-1}+W_{c}^{(x)}\,iv_t+b_c\bigr)
\label{eq:bilstm3}
\end{equation}
The updated forward cell state \( c^{sf}_t \) in Eq. \eqref{eq:bilstm4}, merges retained memory and new inputs through element-wise operations:
\begin{equation}
c^{sf}_t = fg_t\odot c^{sf}_{t-1} + ig_t\odot\tilde{c}^{sf}_t
\label{eq:bilstm4}
\end{equation}
The \((og_t)\) modulation in Eq. \eqref{eq:bilstm5} and Eq. \eqref{eq:bilstm6}, regulates hidden state exposure using sigmoid-activated gating:
\begin{equation}
og_t = \sigma\!\bigl(W_{og}^{(h)}\,h^{sf}_{t-1} + W_{og}^{(x)}\,iv_t + b_{og}\bigr)
\label{eq:bilstm5}
\end{equation}
\begin{equation}
h^{sf}_t = og_t\odot\tanh(c^{sf}_t)
\label{eq:bilstm6}
\end{equation}
We reuse the symbols $ig_t, fg_t, og_t$ for both forward and backward passes. The direction is indicated in superscripts on the hidden and cell states.
The backward LSTM layer operates inversely, mirrors Eq. \eqref{eq:bilstm1}–\eqref{eq:bilstm6} with superscript \(sf\) replaced by \(sb\) and time flowing from \(t_n\rightarrow t_1\) to compute hidden states \( h^{sb}_t \). The final Bi-LSTM hidden state \( h^s_t \) concatenates forward \( h^{sf}_t \) and backward \( h^{sb}_t \) outputs:
\begin{equation}
 h^s_t = \text{concatenate}\!\bigl(h^{sf}_t,\,h^{sb}_t\bigr)
\label{eq:bilstm7}
\end{equation}
Bi-LSTM complements LSTM by adding reverse-time context, which is especially beneficial in noisy and regime-switching financial data. The Bi-LSTM design mitigates contextual blindness, lag bias, and volatility oversimplification, which are key limitations of unidirectional models. To enhance hierarchical feature abstraction, an St-LSTM is used that cascades multiple LSTM layers to enable progressive multi-scale refinement of temporal patterns. The St-LSTM branch leverages layered gating mechanisms to selectively retain or discard information across increasingly abstract temporal horizons, thereby mitigating vanishing gradient effects and capturing long-range dependencies at multiple scales. To produce a compact yet expressive feature representation suitable for vertical fusion, the branch is configured with three stacked LSTM layers with progressive functional roles. The first two layers, each comprising 128 units with \texttt{return\_sequences=True}, preserve and hierarchically refine the temporal resolution of the input sequence, yielding sequences of hidden states in $\mathbb{R}^{w \times 128}$ that encode granular to intermediate dynamics within the observation window. These outputs are then fed into a third layer containing 64 units with \texttt{return\_sequences=False}, which functions not as a predictor but as a temporal encoder which collapses the entire sequence into a single fixed-length latent vector, specifically the final hidden state $\mathbf{F}_{\text{St-LSTM}}^{(-1)} \in \mathbb{R}^{64}$. This 64-dimensional vector constitutes the penultimate representation used in subsequent fusion stages. All layers are regularized via batch normalization and dropout (rate = 0.3) applied to their outputs, enhancing generalization and reducing co-adaptation among units (see Fig.~\ref{fig:fusemodel}). This architecture enables progressive refinement of temporal patterns. Lower layers model daily fluctuations, while higher layers capture weekly or quarterly trends.

The St-LSTM propagates sequential data through its layered structure, where each layer $l$ receives the hidden state $h^{s,l-1}_t$ from the preceding layer as input. This layered approach first filters noise and extracts primary trends and deeper layers, then enriches context by integrating multi-scale dependencies. Although we write the gates as $ig_t, fg_t,$ and $og_t$ but they are specific to each layer, as indicated by the layer $l$ in the weight matrices $W_{*}^{(l,h)}$ and $W_{*}^{(l,r)}$ and are implicitly tied to the layer index $l$, as clarified by their associated weight matrices and biases.
The $(ig_t)$ in Eq. \eqref{eq:stlstm1} modulates the integration of inter-layer inputs $h^{s,l-1}_t$ and prior hidden states $h^{s,l}_{t-1}$:

\begin{equation}
ig_t = \sigma\!\bigl(
    W_{ig}^{(l,h)}\,h^{s,l-1}_t
  + W_{ig}^{(l,r)}\,h^{s,l}_{t-1}
  + b_{ig}^l
\bigr)
\label{eq:stlstm1}
\end{equation}
The $(fg_t)$ in Eq. \eqref{eq:stlstm2} governs retention of historical cell state $ c^{s,l}_{t-1} $:
\begin{equation}
fg_t = \sigma\!\bigl(
    W_{fg}^{(l,h)}\,h^{s,l-1}_t
  + W_{fg}^{(l,r)}\,h^{s,l}_{t-1}
  + b_{fg}^l
\bigr)
\label{eq:stlstm2}
\end{equation}
The candidate cell state $ \tilde{c}^{s,l}_t $ in Eq. \eqref{eq:stlstm3} generates modulated inputs via hyperbolic tangent:
\begin{equation}
\tilde{c}^{s,l}_t = \tanh\!\bigl(
    W_{c}^{(l,h)}\,h^{s,l-1}_t
  + W_{c}^{(l,r)}\,h^{s,l}_{t-1}
  + b_{c}^l
\bigr)
\label{eq:stlstm3}
\end{equation}
The updated cell state $ c^{s,l}_t $ in Eq. \eqref{eq:stlstm4} merges retained memory and new features:
\begin{equation}
c^{s,l}_t =
    fg_t \odot c^{s,l}_{t-1}
  + ig_t \odot \tilde{c}^{s,l}_t
\label{eq:stlstm4}
\end{equation}
The $(og_t)$ in Eq. \eqref{eq:stlstm5} and \eqref{eq:stlstm6} regulates hidden state exposure:
\begin{equation}
og_t = \sigma\!\bigl(
    W_{og}^{(l,h)}\,h^{s,l-1}_t
  + W_{og}^{(l,r)}\,h^{s,l}_{t-1}
  + b_{og}^l
\bigr)
\label{eq:stlstm5}
\end{equation}
\begin{equation}
h^{s,l}_t = og_t \odot \tanh\!\bigl(c^{s,l}_t\bigr)
\label{eq:stlstm6}
\end{equation}
The St-LSTM addresses shallow feature limitations and temporal sparsity by cascading layers to model granular trends (layer 1), weekly cycles (layer 2), and macroeconomic shifts (layer 3). Single-layer LSTMs struggle to disentangle overlapping volatility cycles (e.g., daily noise vs.\ quarterly trends), while rare market events (e.g., flash crashes) require multi-scale contextualization. By cascading three LSTM layers, the St-LSTM enables progressive abstraction, error mitigation, and robustness. Lower layers (\( l = 1 \)) capture local trends, while higher layers (\( l = 3 \)) model global regimes. Layer-wise gating mitigates gradient instability, and hierarchical processing reduces overfitting to transient noise.

\subsection{VertiFuse Layer}
\label{sec:verti}
The VertiFuse layer, implemented as the third procedure VertiFuse in Algorithm \ref{alg:VertiFuseX-2}, is responsible for integrating heterogeneous temporal features from the LSTM, Bi-LSTM, and St-LSTM branches, and integrating the fused temporal representation with the aligned penultimate features from the parallel non-temporal DNN mapping. Instead of using naive output averaging or decision-level ensembling, which are typical in econometric forecast combination and hybrid forecasting frameworks, VertiFuse takes a different approach. It directly operates on the representations from the penultimate layer. By fusing latent feature vectors $F_i$ rather than scalar predictions $\hat{y}_i$, the VertiFuseX enables learned, feature-level interaction between heterogeneous temporal priors to task-specific compression.

\begin{algorithm}[ht]
\caption{VertiFuseX: Vertical Fusion for Generalizable Financial Forecasting (Part 2: Vertical Fusion \& Optimization)}
\label{alg:VertiFuseX-2}
\begin{algorithmic}[1]
\Procedure{VertiFuse}{$D_{\text{train}}, D_{\text{val}}, D_{\text{test}}$}
    \State $\phi_{\text{LSTM}}, \phi_{\text{Bi-LSTM}}, \phi_{\text{St-LSTM}}, \phi_{\text{DNN}} \gets \text{FeatExt}()$
    
    \State Define vertical fusion operators (learnable parameters $\Theta_{\text{fuse}}$, $\sigma = \text{ReLU}$):
    \State \hspace{\algorithmicindent}$\mathbf{F}_{\text{stack}}^{\text{temp}}(\mathbf{x}) \gets \text{concat}\!\bigl(\phi_{\text{LSTM}}(\mathbf{x}), \phi_{\text{Bi-LSTM}}(\mathbf{x}), \phi_{\text{St-LSTM}}(\mathbf{x})\bigr) \in \mathbb{R}^{3d_T}$ \Comment{Eq.~\eqref{eq:vertifuse_stack}}
    \State \hspace{\algorithmicindent}$\mathbf{F}_{\text{fused}}(\mathbf{x}; \Theta_{\text{temp}}) \gets \sigma\!\bigl(\mathbf{W}_{\text{fuse}} \mathbf{F}_{\text{stack}}^{\text{temp}}(\mathbf{x}) + \mathbf{b}_{\text{fuse}}\bigr) \in \mathbb{R}^{d_T}$ \Comment{Eq.~\eqref{eq:vertifuse_temp}}
    
    \State \hspace{\algorithmicindent}$\mathbf{F}_{\text{DNN}(-1)}(\mathbf{x}) \gets \phi_{\text{DNN}}(\mathbf{x})$ \Comment{Penultimate DNN representation, already $\mathbb{R}^{d_T}$}
    \State \hspace{\algorithmicindent}$\hat{\mathbf{F}}_{\text{DNN}}(\mathbf{x}; \Theta_{\text{align}}) \gets \sigma\!\bigl(\mathbf{W}_{\text{align}} \mathbf{F}_{\text{DNN}(-1)}(\mathbf{x}) + \mathbf{b}_{\text{align}}\bigr) \in \mathbb{R}^{d_T}$ \Comment{Eq.~\eqref{eq:vertifuse_align}; retained for generality}
    
    \State \hspace{\algorithmicindent}$\mathbf{F}_{\text{stack}}^{\text{joint}}(\mathbf{x}) \gets \text{concat}\!\bigl(\hat{\mathbf{F}}_{\text{DNN}}(\mathbf{x}), \mathbf{F}_{\text{fused}}(\mathbf{x})\bigr) \in \mathbb{R}^{2d_T}$
    \State \hspace{\algorithmicindent}$\mathbf{F}_{\text{combined}}(\mathbf{x}; \Theta_{\text{joint}}) \gets \sigma\!\bigl(\mathbf{W}_{\text{comb}} \mathbf{F}_{\text{stack}}^{\text{joint}}(\mathbf{x}) + \mathbf{b}_{\text{comb}}\bigr) \in \mathbb{R}^{d_T}$ \Comment{Eq.~\eqref{eq:vertifuse_joint}}
    
    \State \hspace{\algorithmicindent}$\hat{y}(\mathbf{x}; \Theta_{\text{out}}) \gets \mathbf{w}_{\text{out}}^\top \mathbf{F}_{\text{combined}}(\mathbf{x}) + b_{\text{out}}$ \Comment{Eq.~\eqref{eq:vertifuse_output}}
    
    \State Let $\Theta = \{\theta_{\text{LSTM}}, \theta_{\text{Bi-LSTM}}, \theta_{\text{St-LSTM}}, \theta_{\text{DNN}}, \Theta_{\text{fuse}}, \Theta_{\text{align}}, \Theta_{\text{joint}}, \Theta_{\text{out}}\}$ be all trainable parameters
    \State Apply $\ell_2$ regularization ($\lambda = 10^{-4}$) across $\Theta$
    \State Optimize $\Theta$ jointly via backpropagation to minimize MSE loss:
    \Statex \hspace{\algorithmicindent}$\hat{\Theta} = \arg\min_{\Theta} \frac{1}{|D_{\text{train}}|} \sum_{(\mathbf{x}, y) \in D_{\text{train}}} \bigl(\hat{y}(\mathbf{x}; \Theta) - y\bigr)^2 + \lambda \|\Theta\|_2^2$
    \Statex \hspace{\algorithmicindent}// Adam optimizer (initial $\text{lr}=10^{-4}$), batch size 64, max 50 epochs
    \Statex \hspace{\algorithmicindent}// Early stopping (patience=10) and ReduceLROnPlateau (patience=5, factor=0.5) monitored on $D_{\text{val}}$
    \Statex \hspace{\algorithmicindent}// Gradients propagate through all temporal and non-temporal mappings and fusion layers simultaneously
    
    \State Compute predictions on held-out test set: $\hat{Y}_{\text{test}} = \{\hat{y}(\mathbf{x}; \hat{\Theta}) \mid \mathbf{x} \in X_{\text{test}}\}$
    \State \text{metrics} $\gets$ \text{Evaluate}($\hat{Y}_{\text{test}}, Y_{\text{test}}$) \Comment{MAE, RMSE, MAPE per Table~\ref{tab:evaluation_criteria}}
    \State \textbf{return} \text{metrics}
\EndProcedure
\end{algorithmic}
\end{algorithm}

Let $\mathbf{X} \in \mathbb{R}^{w \times f}$ denote the input window of length $w$ (here $w = 20$) with $f$ features. While the proposed architecture natively supports multivariate inputs (\(f > 1\)), this study deliberately adopts a strictly univariate configuration (\(f = 1\), closing price).

This choice is not a limitation of VertiFuseX, but a controlled experimental design intended to isolate the performance contribution of the vertical fusion mechanism itself. By restricting all models to identical raw price input, we ensure that performance gains observed in section 4 arise from architectural innovation rather than from exogenous variables or manual feature engineering. In operational settings, additional variables such as volume, technical indicators, and macroeconomic signals can be easily incorporated by increasing the feature dimension \(f\). The three temporal branches \textit{extract features} via:

\begin{equation}
    \phi_{\text{LSTM}}, \phi_{\text{Bi-LSTM}}, \phi_{\text{St-LSTM}} :
\mathbb{R}^{w \times f} \rightarrow \mathbb{R}^{d_T^{(i)}},
\end{equation}

where $d_T^{(i)}$ denotes the dimensionality of the penultimate-layer representation of the $i$-th temporal branch. In practice, we enforce $d_T^{(1)} = d_T^{(2)} = d_T^{(3)} = N$ through architectural design, enabling direct vertical stacking without projection. In our implementation, this shared dimensionality is fixed to $N = d_T = 64$ for all temporal branches. {The shared width was fixed a priori to permit direct stacking across recurrent branches without separate branch-specific alignment projections, while keeping the fusion layer compact.}
\begin{align}
    \mathbf{F}_{\text{LSTM}} &= \phi_{\text{LSTM}}(\mathbf{X}) = [X_{(1)}, X_{(2)}, \dots, X_{(N)}], \label{eq:lstm_vertifuse} \\
    \mathbf{F}_{\text{Bi-LSTM}} &= \phi_{\text{Bi-LSTM}}(\mathbf{X}) = [Y_{(1)}, Y_{(2)}, \dots, Y_{(N)}], \label{eq:bilstm_vertifuse} \\
    \mathbf{F}_{\text{St-LSTM}} &= \phi_{\text{St-LSTM}}(\mathbf{X}) = [Z_{(1)}, Z_{(2)}, \dots, Z_{(N)}], \label{eq:stlstm_vertifuse}
\end{align}
These representations correspond to the outputs of the last hidden (penultimate) layer of each temporal model, immediately before the task-specific prediction layer. Working at this level ensures that the model-specific inductive biases, such as sequential memory in LSTMs, bidirectional context in Bi-LSTMs, and hierarchical abstraction in St-LSTMs, are retained before being combined.

\subsubsection{Theoretical Motivation for Penultimate-Layer Vertical Fusion}\label{subsec:theory_motivation}
Let $\mathbf{X} \in \mathbb{R}^{w \times f}$ denote the input time window and $y \in \mathbb{R}$ the one-step-ahead target. Standard late-fusion and ensemble approaches combine predictions $\hat{y}_i = g_i(\phi_i(\mathbf{X}))$, where $\phi_i$ is a temporal encoder and $g_i$ is a task-specific regression head. These final-layer outputs are optimized directly for loss minimization and therefore constitute compressed task-specific functions of the input, which may discard higher-order temporal structure.

In contrast, VertiFuseX operates on penultimate-layer representations $\mathbf{F}_i = \phi_{i(-1)}(\mathbf{X})$, which retain richer temporal abstractions prior to task-specific compression. From an information-theoretic perspective, intermediate representations satisfy
\begin{equation}
    I(\mathbf{X}; \mathbf{F}_i) \geq I(\mathbf{X}; \hat{y}_i),
\end{equation}
implying that fusing at this level preserves more mutual information with the input signal. This is particularly important in non-stationary financial time series, where predictive structure is distributed across multiple temporal scales and cannot be reliably summarized by scalar outputs alone.

Vertical fusion in VertiFuseX concatenates and reweights these intermediate features via a learned affine transformation, enabling feature-level interaction across heterogeneous temporal representations. {Unlike decision-level ensembles, which average independent scalar predictions and may lose diversification benefits when branch errors become highly correlated during regime shifts, vertical fusion allows gradients to propagate jointly through all branches.} This encourages complementary specialization while suppressing redundant temporal patterns. Crucially, because this reweighting is entirely data-driven, it enables inference-time flexibility within a fixed-parameter model. Even without retraining, the fusion layer can conditionally emphasize Bi-LSTM features during periods of elevated noise or St-LSTM features during more stable trending regimes, allowing the model to respond to changing market conditions through feature-level reallocation rather than parameter updates.

{By fusing after each branch has established a stable temporal abstraction, but before final regression, VertiFuseX effectively avoids both early fusion, which combines raw, noisy features, and late fusion, which merges overly compressed outputs. This intermediate fusion point yields a more expressive representation and provides a representational rationale for the empirical reductions in MAE, RMSE, and MAPE observed across markets.}

{
\subsubsection{Information-Preservation Rationale for Penultimate-Layer Fusion}
\label{sec:rationale_R2}
 
The inequality $I(\mathbf{X};\mathbf{F}_i)\ge I(\mathbf{X};\hat y_i)$ stated above follows from the data-processing relation between a representation and its deterministic output.  Consider the penultimate features of the branches, written jointly as $Z_{\mathrm{vert}}=(\mathbf{F}_1,\dots,\mathbf{F}_m)$, and their scalar final outputs $Z_{\mathrm{late}}=(\hat y_1,\dots,\hat y_m)$. Since each final output is a deterministic function of its own penultimate representation, $Z_{\mathrm{late}}$ is a function of $Z_{\mathrm{vert}}$. Conditioning on the penultimate features therefore cannot reduce the information available about the target relative to conditioning on the final outputs, which we summarize as: 
\begin{equation}
I(y;Z_{\mathrm{vert}}) \;\ge\; I(y;Z_{\mathrm{late}}).
\label{eq:rationale_R2}
\end{equation}
In words, the penultimate features carry at least as much information about the target as the deterministic final outputs do. Two qualifications are important: First, Eq.~\eqref{eq:rationale_R2} describes a representational property. Under the sufficiency assumption that the fusion head can access the relevant structure in $Z_{\mathrm{vert}}$, fusing at the penultimate layer offers a representational advantage over fusing already-compressed final outputs. Second, this property does not guarantee finite-sample superiority, because a richer representation can also be harder to estimate and may, under limited data or weak regularisation, lead to overfitting.
 
For this reason we treat Eq.~\eqref{eq:rationale_R2} as a rationale that is then tested empirically. The ablation in Section~\ref{sec:ablation} (Table~\ref{tab:ablation}) supports the mechanism. Under the fixed regularisation regime, penultimate fusion improves over final-layer fusion at a comparable fusion-stage parameter count, which is consistent with the representational view above rather than with an increase in raw capacity.
 
A related observation concerns decision-level averaging, which is common in econometric forecast combination. When branch errors become highly correlated, as they may during regime shifts, the variance of an averaged forecast approaches that of the individual branches, so the usual diversification benefit may be lost. VertiFuseX can mitigate this because it combines representations through learned feature-level reweighting rather than averaging scalar outputs, which allows the fusion layer to emphasise different branches as conditions change. We do not claim that this guarantees lower variance. We note only that this view is consistent with the empirical evidence in the paper, namely the larger relative gains over
decision-level averaging on the volatile NASDAQ in Section~\ref{sec:ablation} and the shallower drawdowns during the stress periods in Section~\ref{subsec:extrememarket}. }

\subsubsection{Stage 1: Temporal VertiFuse}
The first stage of VertiFuse performs feature-level fusion across the three temporal branches. We begin by vertically stacking (concatenating) the three feature vectors. This ‘vertical stacking’ denotes concatenation along the feature axis, preserving each branch’s latent temporal abstraction before regression. Unlike horizontal fusion that combines final predictions, our approach fuses deep representations prior to task-specific compression:
\begin{equation}
    \mathbf{F}_{\text{stack}} = \begin{bmatrix} \mathbf{F}_{\text{LSTM}} \\ \mathbf{F}_{\text{Bi-LSTM}} \\ \mathbf{F}_{\text{St-LSTM}} \end{bmatrix} \in \mathbb{R}^{3N}. \label{eq:vertifuse_stack}
\end{equation}
In the original notation, this corresponds to the vertical stacking operation in Eq.~\eqref{eq:fused_stack_original}
\begin{equation}
    \mathbf{F}_{\text{fused}} = \begin{bmatrix} \mathbf{F}_{\text{LSTM}(-1)} \\ \mathbf{F}_{\text{Bi-LSTM}(-1)} \\ \mathbf{F}_{\text{St-LSTM}(-1)} \end{bmatrix}, \label{eq:fused_stack_original}
\end{equation}
where ''$(-1)$'' indicates the penultimate layer. However, VertiFuseX does not stop at simple concatenation. We define a temporal VertiFuse operator
\begin{equation}
    \mathcal{F}_{\text{temp}} : \mathbb{R}^{N} \times \mathbb{R}^{N} \times \mathbb{R}^{N} \rightarrow \mathbb{R}^{d_T},
\end{equation}
which applies a learned affine transformation followed by a non-linearity:
\begin{equation}
    \mathbf{F}_{\text{fused}} = \mathcal{F}_{\text{temp}}\big(\mathbf{F}_{\text{LSTM}}, \mathbf{F}_{\text{Bi-LSTM}}, \mathbf{F}_{\text{St-LSTM}}\big) = \sigma\left(\mathbf{W}_{\text{fuse}} \mathbf{F}_{\text{stack}} + \mathbf{b}_{\text{fuse}}\right), \label{eq:vertifuse_temp}
\end{equation}
where $\mathbf{W}_{\text{fuse}} \in \mathbb{R}^{d_T \times 3N}$ and $\mathbf{b}_{\text{fuse}} \in \mathbb{R}^{d_T}$ are trainable parameters and $\sigma(\cdot)$ denotes a pointwise non-linear activation (e.g., ReLU). Each coordinate of $\mathbf{F}_{\text{fused}}$ is therefore a learned linear combination of all entries from the three temporal branches, rather than an unweighted sum. This construction allows VertiFuseX to reweight LSTM, Bi-LSTM, and St-LSTM specific temporal patterns in a data driven manner, preserving model-specific inductive biases while mitigating redundancy and multicollinearity across streams. All fusion parameters $\{\mathbf{W}_{\text{fuse}}, \mathbf{b}_{\text{fuse}}, \mathbf{W}_{\text{comb}}, \mathbf{b}_{\text{comb}}\}$ are optimized jointly with the temporal branches via backpropagation under the forecasting loss.

The temporal VertiFuse stage is designed to perform intermediate feature-level fusion prior to task-specific prediction. This means that model-specific abstractions are not mixed at the raw input or low-level feature stages. Instead, they are combined after each branch has established a stable penultimate representation. This design is consistent with the theoretical motivation in Section~\ref{subsec:theory_motivation} that intermediate representations retain richer mutual information than compressed final-layer outputs.

\subsubsection{Stage 2: Joint Temporal--DNN Fusion}
In parallel with the temporal streams, VertiFuseX trains a DNN on the same training data $(\mathbf{X}_{\text{train}}, \mathbf{Y}_{\text{train}})$ to capture complementary non-recurrent abstractions from the same lagged window.  The DNN processes the input window as a static feature vector through three fully connected layers with ReLU activations. Specifically, it comprises two hidden layers with 128 and 64 units, respectively, each followed by dropout ($p=0.3$) for regularization. This three-layer design provides sufficient non-linear representational capacity while remaining computationally efficient.
Let
\begin{equation}
    \phi_{\text{DNN},-2} : \mathbb{R}^{w \times f} \rightarrow \mathbb{R}^{d_D}
\end{equation}
denote the mapping to the penultimate (''$-2$'') layer of the DNN, and
\begin{equation}
    \mathbf{F}_{\text{DNN}(-2)} = \phi_{\text{DNN},-2}(\mathbf{X}) \in \mathbb{R}^{d_D}, \label{eq:dnn_vertifuse}
\end{equation}
denote the corresponding feature vector, whose elements we previously denoted as $\{f_1, f_2, \dots, f_n\}$ for notational convenience.

This parallel DNN stream serves a dual purpose. It captures global non-linear relationships within the price data and provides a dedicated architectural pathway for integrating static or lower-frequency auxiliary features, such as quarterly fundamental data. By processing these inputs separately from the recurrent branches, the DNN preserves non-temporal signal components before they are fused with the temporal dynamics in the VertiFuse layer.
Since $d_D$ and $d_T$ may differ, we introduce a dimension alignment layer to map $\mathbf{F}_{\text{DNN}(-2)}$ into the same feature space as $\mathbf{F}_{\text{fused}}$:
\begin{equation}
    \hat{\mathbf{F}}_{\text{DNN}} = \sigma\left(\mathbf{W}_{\text{align}} \mathbf{F}_{\text{DNN}(-2)} + \mathbf{b}_{\text{align}}\right), \quad \mathbf{W}_{\text{align}} \in \mathbb{R}^{d_T \times d_D}, \ \mathbf{b}_{\text{align}} \in \mathbb{R}^{d_T}, \label{eq:vertifuse_align}
\end{equation}
so that $\hat{\mathbf{F}}_{\text{DNN}} \in \mathbb{R}^{d_T}$ has the same dimensionality as $\mathbf{F}_{\text{fused}}$. In implementation, this corresponds to the dense layer applied to $\mathbf{F}_{\text{DNN}(-2)}$ in Algorithm~\ref{alg:VertiFuseX-2}, ensuring $\text{dim}(\mathbf{F}_{\text{DNN}(-2)}) = \text{dim}(\mathbf{F}_{\text{fused}})$ and avoiding tensor broadcasting issues during fusion.

We then perform a second vertical fusion between the aligned DNN features and the fused temporal representation by stacking and reweighting:
\begin{equation}
    \mathbf{F}_{\text{comb,stack}} = \begin{bmatrix} \hat{\mathbf{F}}_{\text{DNN}} \\ \mathbf{F}_{\text{fused}} \end{bmatrix} \in \mathbb{R}^{2d_T}, \label{eq:vertifuse_stack_joint}
\end{equation}
\begin{equation}
    \mathbf{F}_{\text{combined}} = \mathcal{F}_{\text{joint}}\big(\hat{\mathbf{F}}_{\text{DNN}}, \mathbf{F}_{\text{fused}}\big) = \sigma\left(\mathbf{W}_{\text{comb}} \mathbf{F}_{\text{comb,stack}} + \mathbf{b}_{\text{comb}}\right), \label{eq:vertifuse_joint}
\end{equation}
where $\mathbf{W}_{\text{comb}} \in \mathbb{R}^{d_C \times 2d_T}$ and $\mathbf{b}_{\text{comb}} \in \mathbb{R}^{d_C}$ are trainable parameters. This expression refines the earlier notational shorthand
\begin{equation}
    \mathbf{F}_{\text{combined}} = \begin{bmatrix} \mathbf{F}_{\text{DNN}(-2)} \\ \mathbf{F}_{\text{fused}} \end{bmatrix}, \label{eq:Fcomb_original}
\end{equation}
by making the learned reweighting explicit. Finally, the scalar next-day prediction $\hat{y}$ is obtained via a linear regression head:
\begin{equation}
    \hat{y} = \mathbf{w}_{\text{out}}^{\top} \mathbf{F}_{\text{combined}} + b_{\text{out}}, \label{eq:vertifuse_output}
\end{equation}
with $\mathbf{w}_{\text{out}} \in \mathbb{R}^{d_C}$ and $b_{\text{out}} \in \mathbb{R}$.

While Eq. \eqref{eq:vertifuse_output} formulates a scalar one-step-ahead prediction (\(k = 1\)), the architecture natively supports direct multi-step forecasting by expanding the output projection. Replacing the vector \(\mathbf{w}_{\text{out}} \in \mathbb{R}^{d_C}\) with a matrix \(\mathbf{W}_{\text{out}} \in \mathbb{R}^{d_C \times k}\) allows the fused representation \(\mathbf{F}_{\text{combined}}\) to be mapped to a forecast vector \(\hat{\mathbf{Y}} \in \mathbb{R}^k\). This extension utilizes the multi-scale temporal features already encoded in \(\mathbf{F}_{\text{combined}}\), particularly the longer-horizon abstractions captured by the St-LSTM branch, to generate consistent multi-horizon predictions without altering the upstream fusion architecture.

The formulation of VertiFuseX distinctly sets it apart from existing fusion approaches in several key ways. Firstly, it does not utilize weighted averaging, meaning that it does not impose static scalar weights or convex combinations across branches. Additionally, this formulation distinguishes VertiFuseX from conventional late-fusion pipelines that merge final predictions or logits. Unlike ensemble voting methods where branches are trained independently, this architecture allows gradients to propagate back from the final loss through $\mathbf{W}_{\text{fuse}}$ to all temporal branches simultaneously, enabling co-adaptation. Finally, rather than employing raw concatenation without learned interaction, VertiFuseX utilizes the matrices $\mathbf{W}_{\text{fuse}}$ and $\mathbf{W}_{\text{comb}}$ to learn task-specific reweighting of all feature dimensions, ensuring that the concatenated features are not treated uniformly. 
In summary, VertiFuseX performs sequence-to-vector encoding in each temporal branch, aligns all penultimate representations to a shared latent space ($d_T = 64$), and applies learned vertical fusion prior to task-specific regression, ensuring expressive yet causally valid forecasting.
{Intermediate feature fusion, while established in multimodal learning for combining different input modalities, is here applied to representations from multiple temporal encoders of the same univariate time series. The key contribution lies in performing this fusion at the penultimate layer within a purely temporal setting. A simple learned affine transformation was chosen as the fusion operator due to its low computational overhead while maintaining effectiveness.}


\section{Experimental results}
We evaluate VertiFuseX’s quantitative performance using MAE, RMSE, and MAPE, capturing average error, error magnitude, and percentage error, respectively. These evaluation metrics are defined in Table~\ref{tab:evaluation_criteria}. The evaluation in this study is organized on a dataset-centric basis rather than a model-centric one. Baseline comparisons are restricted to four U.S. indices that share an identical trading calendar, enabling fair and controlled comparison across baseline architectures. In contrast, state-of-the-art comparisons are performed on a per-index basis, where VertiFuseX is evaluated using the same datasets and periods as those used in prior studies. This design ensures both internal consistency in baseline evaluation and external validity across diverse global markets.
\begin{table}[ht]
\centering
\caption{Performance metrics used in evaluating VertiFuseX against baselines and state-of-the-art models}
\label{tab:evaluation_criteria}
\begin{tabular}{@{}llp{7.5cm}@{}}
\toprule
\textbf{Metric} & \textbf{Mathematical Expression} & \textbf{Purpose and Meaning} \\
\midrule
MAE & $\displaystyle  \frac{1}{N} \sum_{k=1}^{N} |\hat{y}_k - y_k|$ & Average absolute error, treating all deviations equally. \\[6pt]
RMSE & $\displaystyle  \sqrt{\frac{1}{N} \sum_{k=1}^{N} (\hat{y}_k - y_k)^2}$ & Penalizes larger errors more due to squaring. \\[6pt]
MAPE & $\displaystyle  \frac{100}{N} \sum_{k=1}^{N} \left|\frac{\hat{y}_k - y_k}{y_k}\right|$ & Percentage error, enabling scale-independent comparison. \\
\bottomrule
\end{tabular}

\vspace{4pt}
\footnotesize{$N$: Total number of test samples; $\hat{y}_k$: Forecasted value; $y_k$: Actual observed value.}
\end{table}

Hyperparameters were selected prior to the main experiments based on standard practices in financial time-series forecasting and preliminary stability checks across multiple indices. All hyperparameters were determined prior to final evaluation using only the training portion of the data. To preserve strict out-of-sample validity, the last 365 trading days were held out as a locked test set and were never accessed during hyperparameter selection, early stopping, or learning-rate scheduling. Within the training period, an internal validation split was constructed by reserving the final 10\% of the training data (chronologically preceding the test horizon). This validation set was used solely to monitor convergence and to control Early Stopping (patience = 10) and the ReduceLROnPlateau scheduler (patience = 5, factor = 0.5).

Once established, the hyperparameter configuration (e.g., window size $w=20$, dropout = 0.3, $\ell_2 = 10^{-4}$) was held fixed across all datasets. We intentionally avoided dataset-specific hyperparameter optimization in order to assess the robustness and generalization ability of the proposed VertiFuseX architecture under a unified configuration. Table~\ref{tab:initial_params} summarizes this fixed configuration. This unified setup relies on the representational capacity of the VertiFuse layer to provide market-specific flexibility. While core training constraints (e.g., learning rate, window size) remain rigid to ensure stability, the fusion layer's learnable weight matrices ($\mathbf{W}_{\text{fuse}}$) dynamically adjust the contribution of each temporal branch. This enables the model to adapt to high-volatility markets (e.g., NASDAQ) or stable indices (e.g., FTSE 100) through internal feature reweighting, rather than relying on external hyperparameter tuning. All preprocessing, model training, and hyperparameter selection were conducted strictly on training data, with the final 365-day test window held out and never used during model development or selection.

{The input shape is \((\texttt{window\_size}=20, \texttt{features}=1)\), reflecting the controlled univariate setting used to isolate the architectural contribution of penultimate-layer fusion from feature engineering effects. The test set comprises 365 samples, representing one full trading year.} A batch size of 64 ensures gradient stability, with training run over 50 epochs. To rigorously mitigate overfitting given the extensive 15-year training horizon, we enforce a multi-tier regularization strategy. An adaptive learning rate is employed via the Adam optimizer (initial $\text{lr} = 10^{-4}$) coupled with a \texttt{ReduceLROnPlateau} scheduler (patience = 5, factor = 0.5) and Early Stopping (patience = 10). The latter ensures training ceases immediately upon validation loss stagnation, preventing VertiFuseX from overfitting to regime-specific patterns in the deep history. Furthermore, structural complexity is constrained using a high dropout rate (0.3) and \(\ell_2\)  weight decay across all fusion branches. This regularization profile encourages the model to learn adaptable temporal representations, as shown by the reduction of noisy lags in our saliency analysis, rather than overfitting to specific historical patterns. MSE loss prioritizes large errors, aligning with financial risk sensitivity. {Seed 150 is used for the main reported results, and an additional three-seed robustness analysis using seeds 42, 101, and 150 is reported in Table~\ref{tab:ci-R2} to assess sensitivity to random initialization.} Experiments were conducted on Google Colab using a Tesla P100-PCIE-16GB GPU, an Intel Xeon 2.20GHz CPU, 12GB RAM, Python 3.10, TensorFlow 2.14, and CUDA 11.8.  The performance evaluation is structured in two parts: (1) Comparison with baselines, and (2) Comparison with state-of-the-art models. While core hyperparameters remain fixed to ensure rigorous cross-market comparison, VertiFuseX retains substantial adaptive capacity through the data-driven fusion matrices $\mathbf{W}_{\mathrm{fuse}}$ and $\mathbf{W}_{\mathrm{comb}}$ (Eqs.~\eqref{eq:vertifuse_temp}--\eqref{eq:vertifuse_joint}). These matrices learn to reweight temporal branch contributions in a market-specific manner, emphasizing Bi-LSTM features under high-volatility regimes (e.g., NASDAQ) and St-LSTM abstractions under more stable trends (e.g., DJIA), thereby enabling consistent performance gains without index-specific hyperparameter tuning.

\begin{table}[ht]
\centering
\caption{VertiFuseX initialization and training settings}
\label{tab:initial_params}
\begin{tabular}{p{1.8cm}p{2cm}p{3cm}p{4cm}}
\toprule
\textbf{Category} & \textbf{Parameter} & \textbf{Symbol / Value} & \textbf{Rationale} \\
\midrule
\multirow{3}{*}{Data} & Input window size & $w=20$ (\texttt{nb\_days}) & Captures short-term patterns \\
 & Features per step & $f=1$ (close price) & Controlled evaluation to isolate architectural contribution \\
 & Test samples & $N_{\text{test}}=365$ & One full trading year \\[4pt] \hline
\multirow{6}{*}{Training} & Batch size & 64 & GPU-aligned mini-batch \\
 & Epochs & 50 & Faster convergence with LR scheduler and early stop \\
 & Dropout & 0.30 & Stronger regularization \\
 & $\ell_{2}$ regularizer & $10^{-4}$ & Weight decay \\
 & Optimizer & Adam ($lr=10^{-4}$) & Adaptive updates for financial data \\
 & Early stopping & Patience = 10 & Prevents overfitting \\[4pt] \hline
\multirow{2}{*}{Objective} & Loss & MSE & Penalizes large errors \\
 & Evaluation & MAE, RMSE, MAPE & Comprehensive error profile \\ \hline
\multirow{2}{*}{Reproducibility} & Random seed & 150 & Ensures repeatable runs \\
 & Validation split & Final 10\% of the development period & Chronological validation before the locked test set \\
\bottomrule
\end{tabular}
\end{table}

\subsection{Comparison with Baseline Models}\label{subsec:baseline_com}
In this section, we compare VertiFuseX with LSTM, Bi-LSTM, and St-LSTM on the S\&P 500, DJI, NYSE, and NASDAQ indices using MAE, RMSE, and MAPE. To ensure a strictly fair comparison that isolates the architectural benefits of vertical fusion, all baseline models are trained under the identical training and regularization protocol used for VertiFuseX. As summarized in Table \ref{tab:initial_params}, this includes a shared fixed hyperparameter configuration and consistent trading calendars across the four U.S. indices listed in Table \ref{tab:Indices Used}, following the evaluation methodology established in Section~\ref{sec:verti}. This controlled setup ensures that the performance differentials reported in Table \ref{tab:eva_compact} arise from the VertiFuse layer’s feature-level integration, rather than from differences in tuning effort, regularization strength, or data preprocessing.

\begin{figure}[ht]
    \centering
    \includegraphics[width=0.8\textwidth]{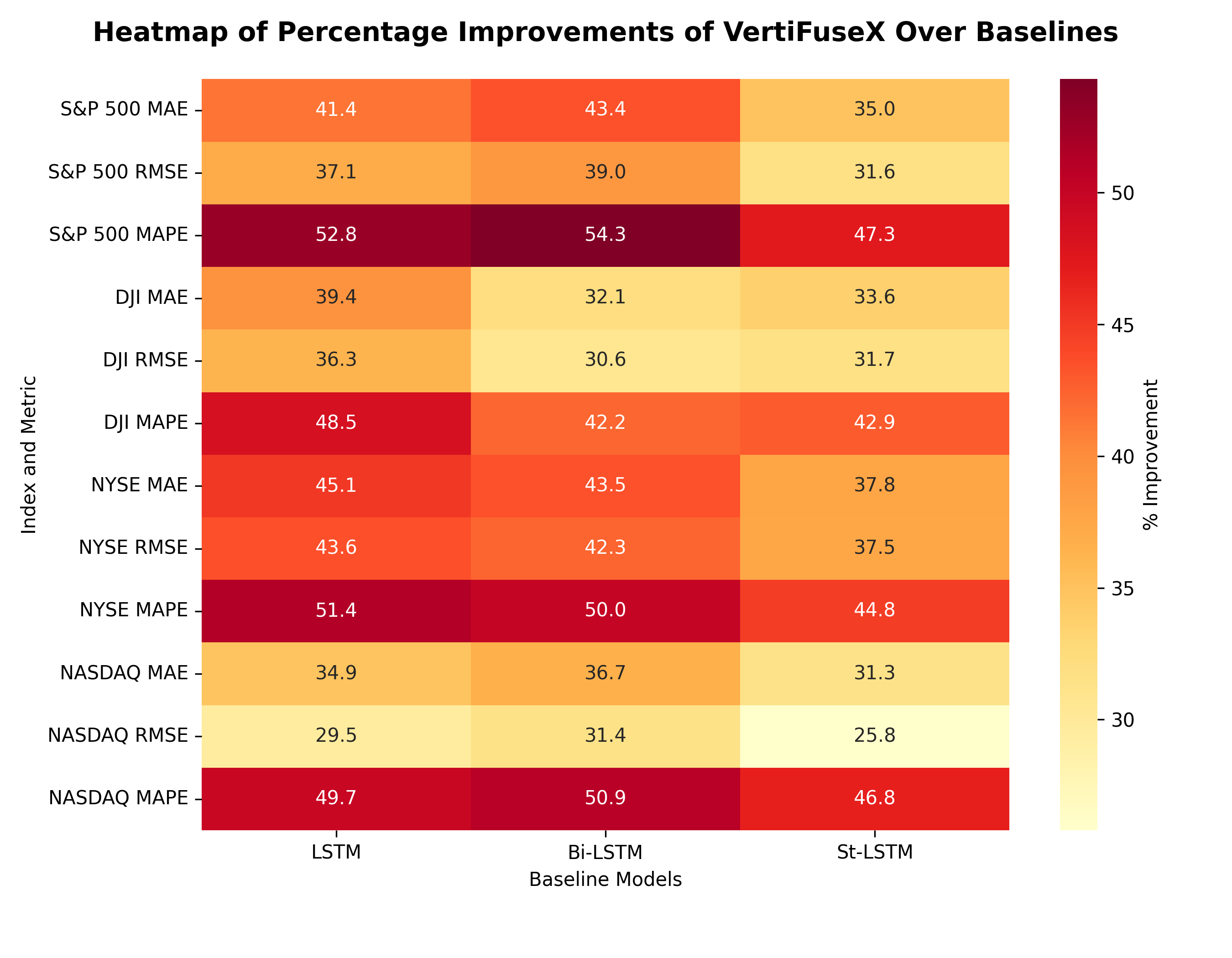}
    \caption{Heatmap of VertiFuseX percentage improvements over baselines}
    \Description{}
    \label{fig:heatmap_baseline}
\end{figure}

Table~\ref{tab:eva_compact} presents the test results, with percentage reductions over baselines italicized for each index–metric pair. VertiFuseX attains the lowest error in all twelve index–metric pairs, with the largest relative gain up to \textit{54.3\%} on the S\&P 500-MAPE.  A visual representation of these improvements is shown in Fig. \ref{fig:heatmap_baseline}, which highlights the percentage reductions of VertiFuseX across all indices and metrics using a heatmap. While our analysis emphasizes non-linear DL baselines that represent contemporary forecasting practice for non-stationary financial time series, we also acknowledge the foundational importance of classical linear models such as ARIMA. These models are well studied in prior literature but typically struggle with regime shifts and high-frequency volatility in modern equity indices. To provide contextual grounding, our state-of-the-art evaluation (Section \ref{subsec:sota_comp}) includes the DE-ABC-Bi-LSTM-ARIMA model \citep{kumar2022three}, which integrates an optimized ARIMA component within a hybrid DL architecture. VertiFuseX’s consistent improvements over this ARIMA-enhanced model further demonstrate its robustness even relative to frameworks that incorporate classical linear dynamics.

{To provide a direct linear baseline, we also evaluate a standalone ARIMA$(p,d,q)$ model with orders selected by AIC on the training partition only. Forecasts are generated in a causal one-step-ahead out-of-sample setting, in which each prediction uses only observations available up to the corresponding forecast origin. As shown in Table~\ref{tab:eva_compact}, ARIMA underperforms all recurrent models, suggesting that nonlinear temporal models better capture the dynamics of the evaluated financial series. Relative to ARIMA, VertiFuseX achieves an average error reduction of approximately 50\% across the twelve index--metric pairs.}

\begin{table}[htbp]
\centering
\caption{VertiFuseX vs.\ baseline comparison on four indices, including standalone ARIMA baseline (best values \textbf{bold}; \% reduction in \textit{italics})}
\label{tab:eva_compact}
\footnotesize
\setlength{\tabcolsep}{3pt}

\begin{tabular}{lllllllllllll}
\toprule
\textbf{Model}
& \multicolumn{3}{c}{\textbf{S\&P 500}}
& \multicolumn{3}{c}{\textbf{DJI}}
& \multicolumn{3}{c}{\textbf{NYSE}}
& \multicolumn{3}{c}{\textbf{NASDAQ}} \\
\cmidrule(lr){2-4} \cmidrule(lr){5-7} \cmidrule(lr){8-10} \cmidrule(lr){11-13}
& M & R & MP & M & R & MP & M & R & MP & M & R & MP \\
\midrule
ARIMA$(p,d,q)$
& 59.12 & 74.85 & 1.45
& 378.45 & 482.67 & 1.15
& 198.76 & 249.34 & 1.27
& 245.89 & 304.56 & 1.89 \\
LSTM
& 51.57 & 64.52 & 1.25
& 333.28 & 424.64 & 1.01
& 170.07 & 213.70 & 1.09
& 206.81 & 256.85 & 1.65 \\
Bi-LSTM
& 53.42 & 66.48 & 1.29
& 297.50 & 389.82 & 0.90
& 165.28 & 208.55 & 1.06
& 212.46 & 263.70 & 1.69 \\
St-LSTM
& 46.49 & 59.31 & 1.12
& 303.97 & 396.18 & 0.91
& 150.18 & 192.65 & 0.96
& 195.82 & 243.98 & 1.56 \\
\textbf{VertiFuseX}
& \textbf{30.22} & \textbf{40.57} & \textbf{0.59}
& \textbf{201.93} & \textbf{270.60} & \textbf{0.52}
& \textbf{93.34} & \textbf{120.43} & \textbf{0.53}
& \textbf{134.54} & \textbf{180.87} & \textbf{0.83} \\
\midrule
\multicolumn{1}{l}{vs.\ ARIMA:}
&\textit{48.9} & {\textit{45.8}} & {\textit{59.3}}
& {\textit{46.6}} & {\textit{43.9}} & {\textit{54.8}}
& {\textit{53.0}} & {\textit{51.7}} & {\textit{58.3}}
& {\textit{45.3}} & {\textit{40.6}} & {\textit{56.1}} \\
\multicolumn{1}{l}{vs.\ LSTM:}
& \textit{41.4} & \textit{37.1} & \textit{52.8}
& \textit{39.4} & \textit{36.3} & \textit{48.5}
& \textit{45.1} & \textit{43.6} & \textit{51.4}
& \textit{34.9} & \textit{29.5} & \textit{49.7} \\
\multicolumn{1}{l}{vs.\ Bi-LSTM:}
& \textit{43.4} & \textit{39.0} & \textit{54.3}
& \textit{32.1} & \textit{30.6} & \textit{42.2}
& \textit{43.5} & \textit{42.3} & \textit{50.0}
& \textit{36.7} & \textit{31.4} & \textit{50.9} \\
\multicolumn{1}{l}{vs.\ St-LSTM:}
& \textit{35.0} & \textit{31.6} & \textit{47.3}
& \textit{33.6} & \textit{31.7} & \textit{42.9}
& \textit{37.8} & \textit{37.5} & \textit{44.8}
& \textit{31.3} & \textit{25.8} & \textit{46.8} \\
\bottomrule
\end{tabular}

\vspace{5pt}
{\footnotesize
Key: M = MAE, R = RMSE, MP = MAPE.\\
ARIMA$(p,d,q)$ orders are selected by AIC on training data only. Italic values show percentage reductions achieved by VertiFuseX relative to each baseline.
}
\end{table}

Starting with the S\&P 500, VertiFuseX achieves the lowest error across all metrics. The model records an MAE of 30.22, an RMSE of 40.57, and an MAPE of 0.59. These represent improvements of \textit{41.4\%}, \textit{37.1\%}, and \textit{52.8\%} respectively over the standard LSTM, and \textit{43.4\%}, \textit{39.0\%}, and \textit{54.3\%} compared to the Bi-LSTM. Even against the strongest baseline, St-LSTM, the model achieves \textit{35.0\%}, \textit{31.6\%}, and \textit{47.3\%} reductions across the three metrics. The consistent gains across both absolute and percentage-based errors demonstrate that the model’s advantage is metric-agnostic. On the DJI index, VertiFuseX achieves an MAE of 201.93, an RMSE of 270.60, and an MAPE of 0.52. It outperforms LSTM by \textit{39.4\%}, \textit{36.3\%}, and \textit{48.5\%}, Bi-LSTM by \textit{32.1\%}, \textit{30.6\%}, and \textit{42.2\%}, and St-LSTM by \textit{33.6\%}, \textit{31.7\%}, and \textit{42.9\%}, respectively. These consistent margins reinforce VertiFuseX's stability across indices with differing volatility and scale, indicating maintained relative accuracy despite larger absolute price scales.

Evaluated on the NYSE index, VertiFuseX achieves an MAE of 93.34, RMSE of 120.43, and MAPE of 0.53. The performance gains over LSTM are \textit{45.1\%}, \textit{43.6\%}, and \textit{51.4\%}, over Bi-LSTM \textit{43.5\%}, \textit{42.3\%}, and \textit{50.0\%}, and over St-LSTM \textit{37.8\%}, \textit{37.5\%}, and \textit{44.8\%}. The uniformity of these results across all three metrics indicates that the model’s strategy delivers stable and generalized enhancements in a broad-based composite index. When applied to the NASDAQ index, known for its high volatility due to its concentration in technology stocks, it also reflects significant improvements. VertiFuseX achieves an MAE of 134.54, an RMSE of 180.87, and an MAPE of 0.83. Compared to LSTM, these correspond to \textit{34.9\%}, \textit{29.5\%}, and \textit{49.7\%} reductions, and against Bi-LSTM, \textit{36.7\%}, \textit{31.4\%}, and \textit{50.9\%}. Compared to St-LSTM, these correspond to \textit{31.3\%}, \textit{25.8\%}, and \textit{46.8\%} reductions. Contrasting this with the more stable, bluechip DJIA (MAPE 0.52\%) reveals consistent relative performance gains despite fundamental differences in volatility structure, where NASDAQ exhibits rapid regime shifts and DJIA displays slower mean-reversion. VertiFuseX maintains approximately 49\% MAPE reduction over the LSTM baseline in both cases. This consistency indicates that the vertical fusion mechanism stabilizes predictive performance across heterogeneous volatility regimes rather than overfitting to a specific market profile. It also highlights the model's ability to adapt to quick market shifts, maintaining robustness even in high-noise, fast-moving situations.

\begin{figure*}[htbp]
    \centering
    \subfigure[S\&P 500]{\includegraphics[width=0.45\textwidth]{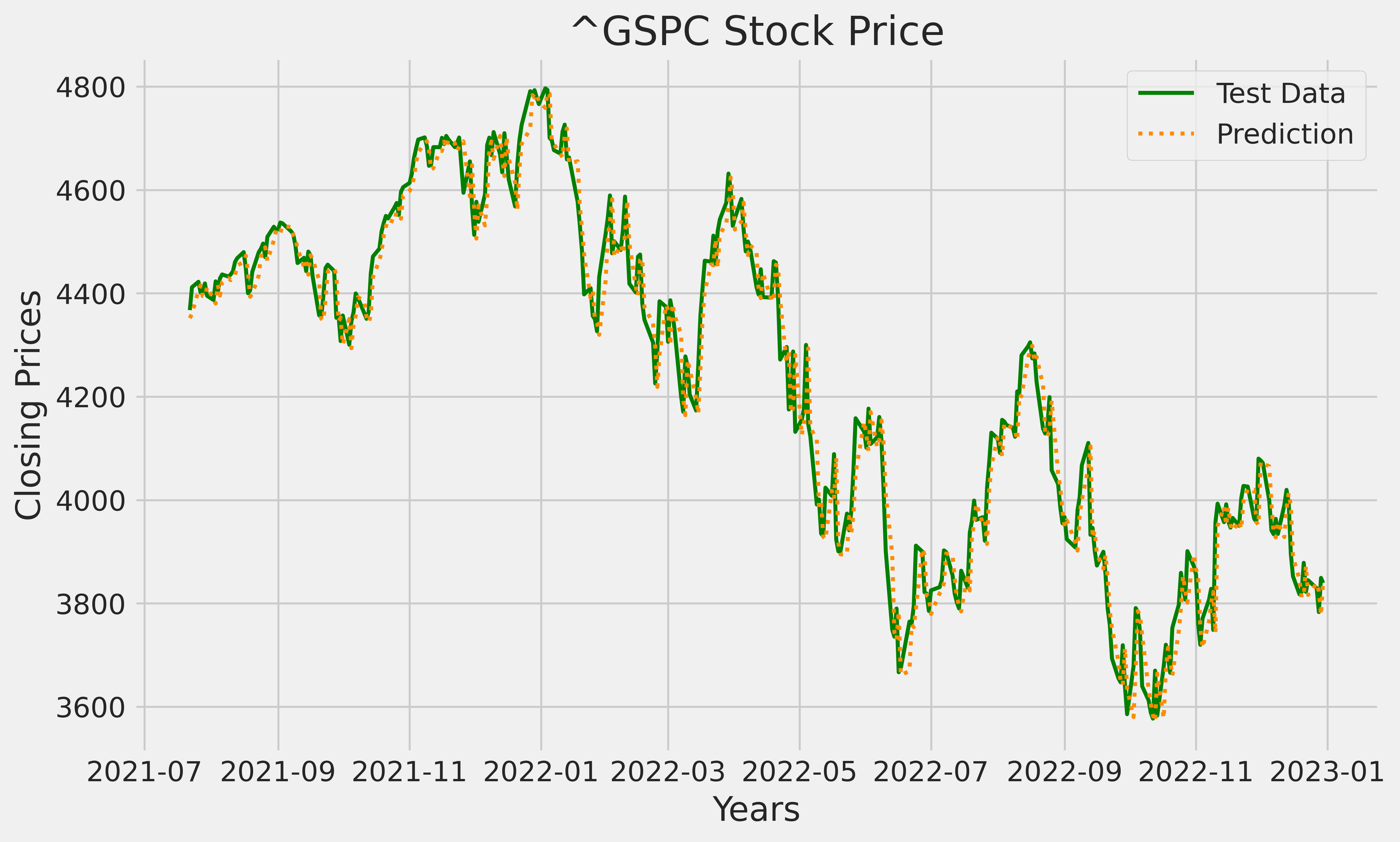}} 
    \subfigure[DJI]{\includegraphics[width=0.45\textwidth]{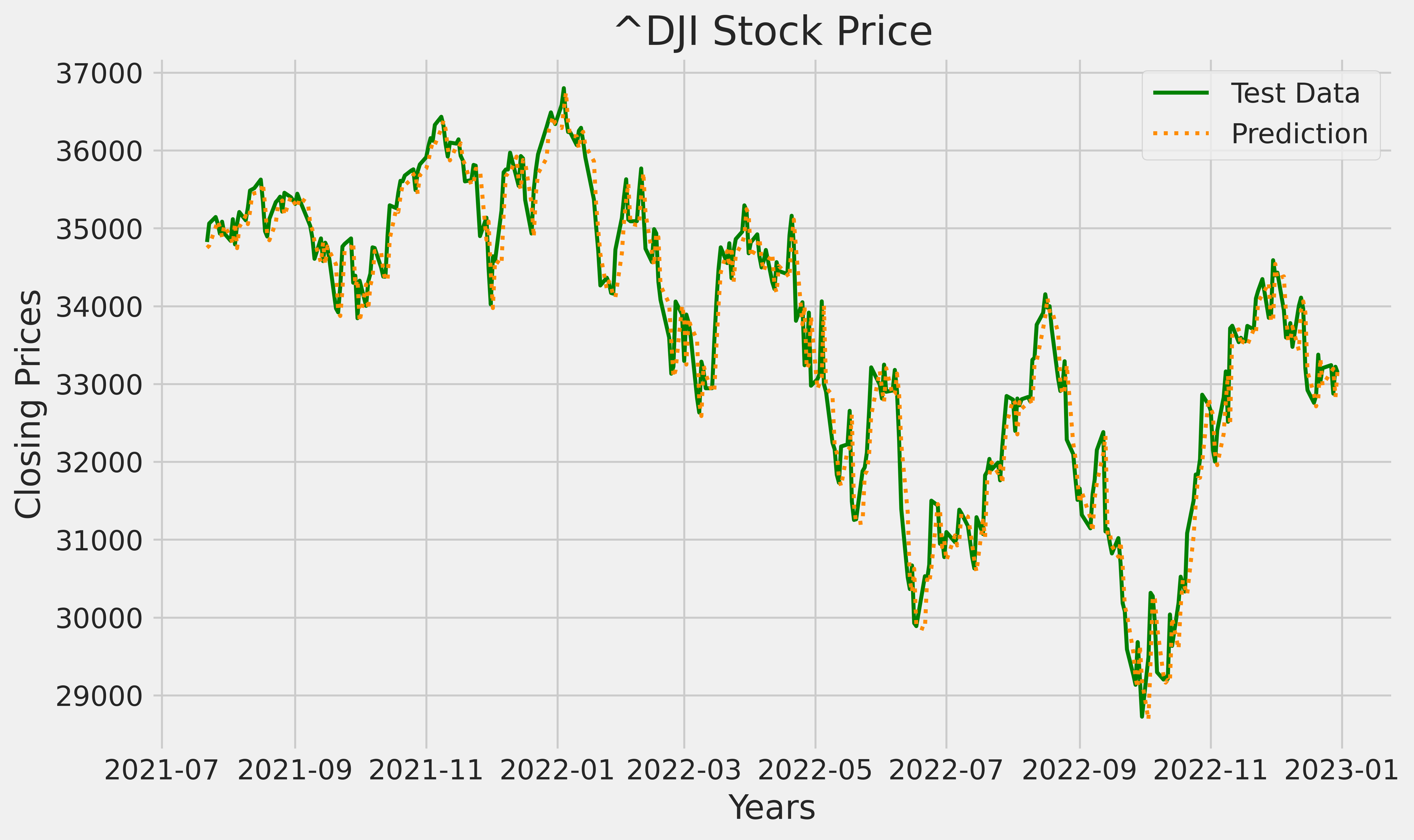}} 
    \subfigure[NYSE]{\includegraphics[width=0.45\textwidth]{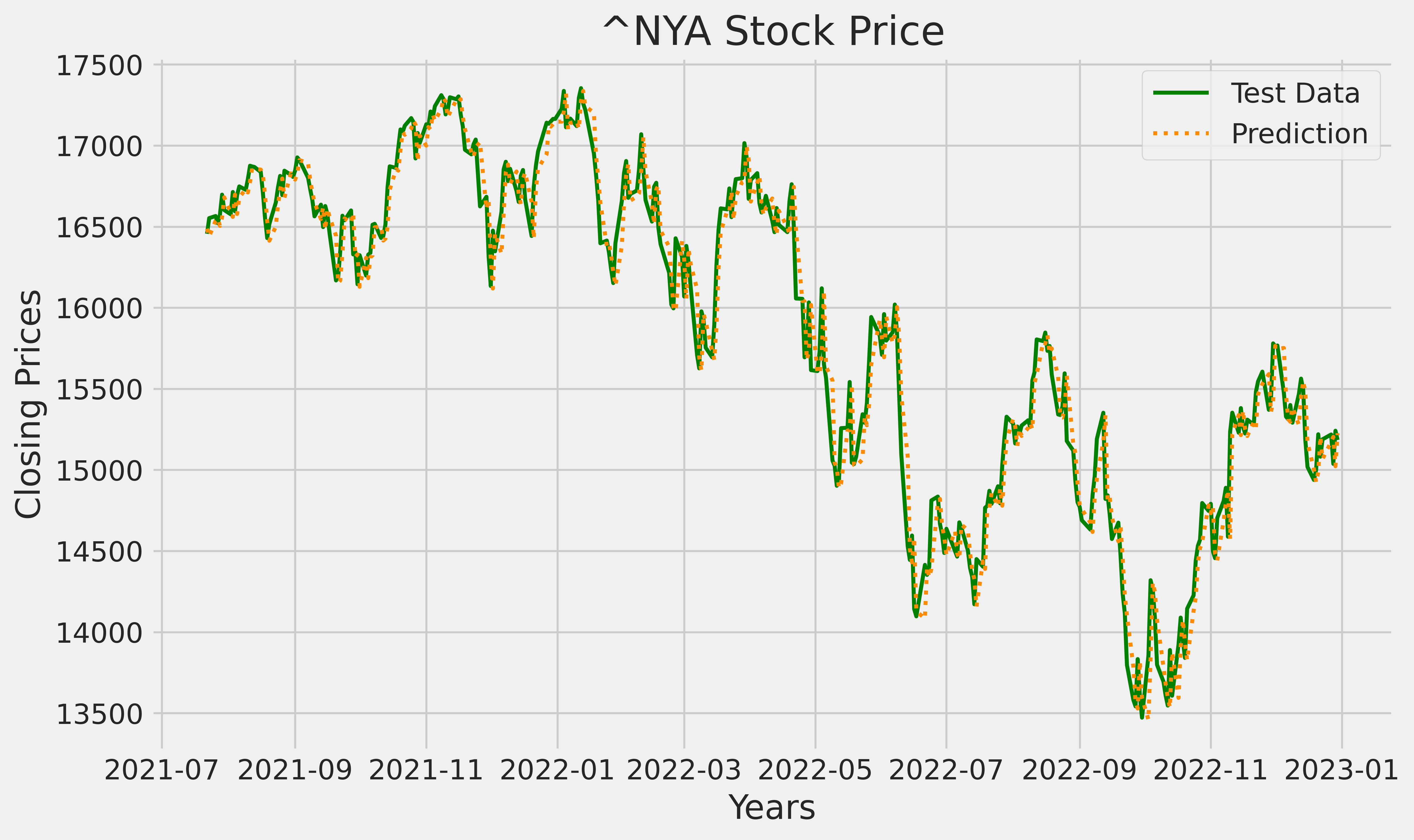}} 
    \subfigure[NASDAQ]{\includegraphics[width=0.45\textwidth]{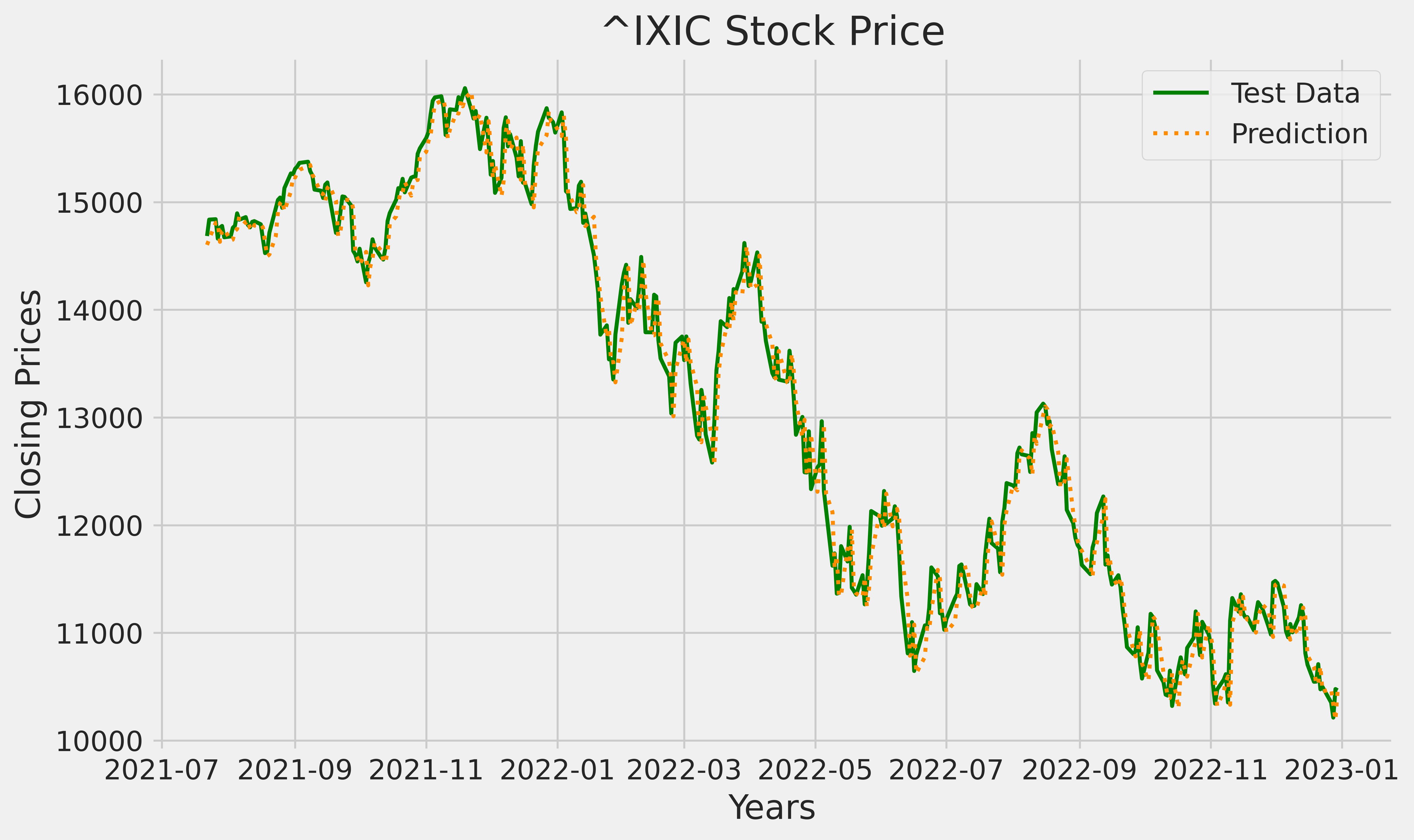}} 
    \caption{Predictions using training and test data. The X-axis represents years, while the Y-axis shows stock index closing prices. The blue and green lines represent actual values, while the orange line indicates predicted values.}
    \Description{}
    \label{fig: Train_test pred}
   
    \subfigure[S\&P 500]{\includegraphics[width=0.43\textwidth]{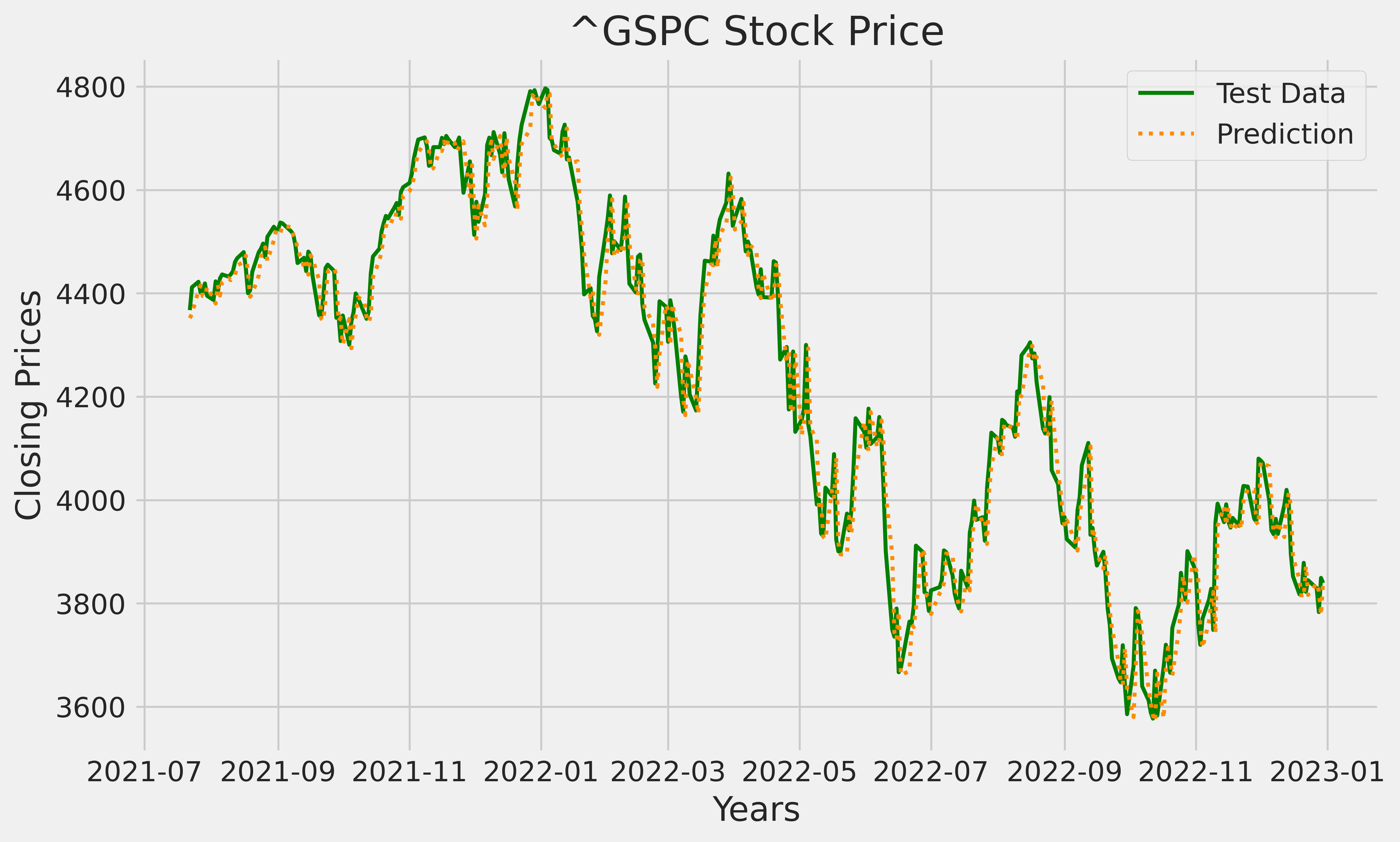}} 
    \subfigure[DJI]{\includegraphics[width=0.43\textwidth]{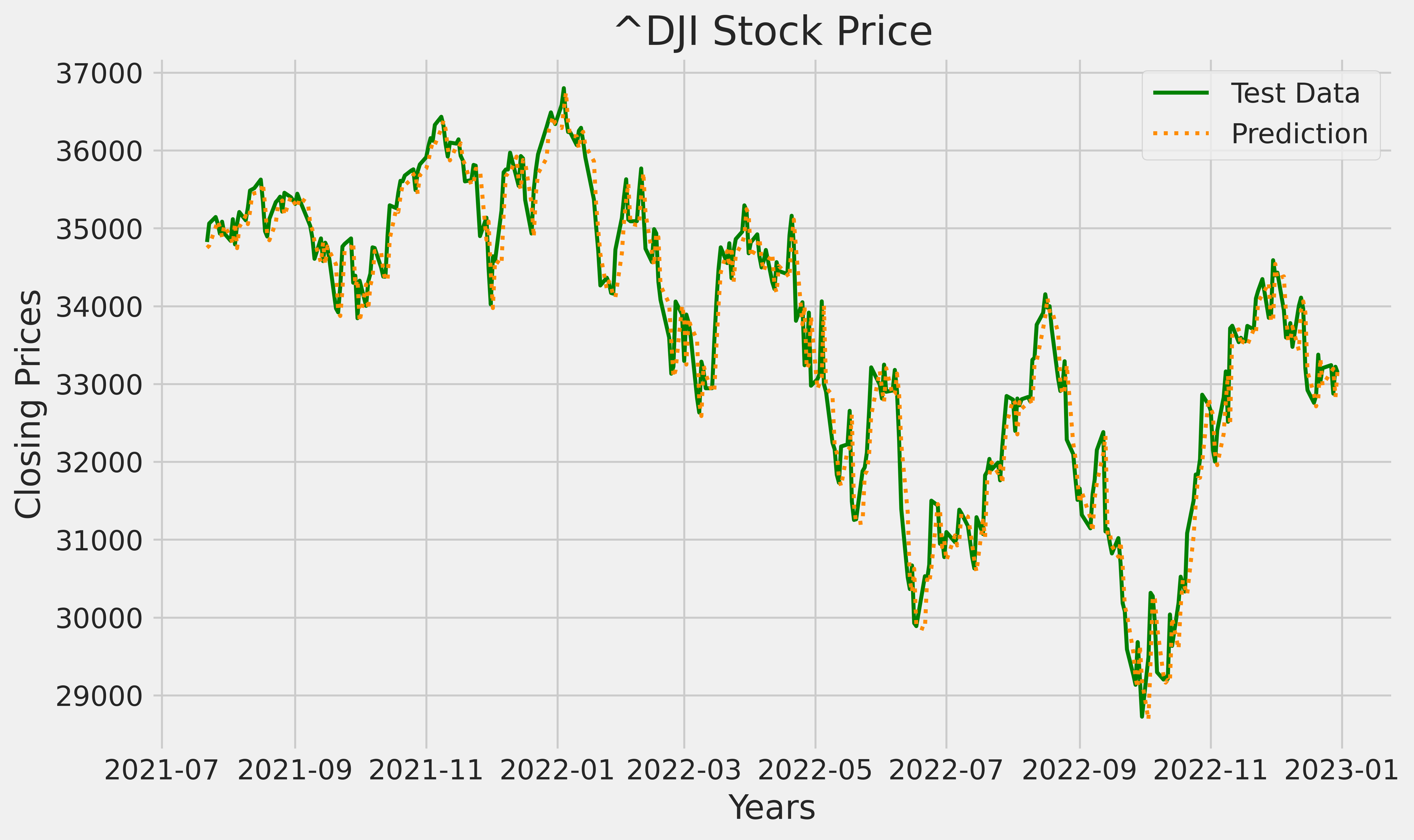}} 
    \subfigure[NYSE]{\includegraphics[width=0.43\textwidth]{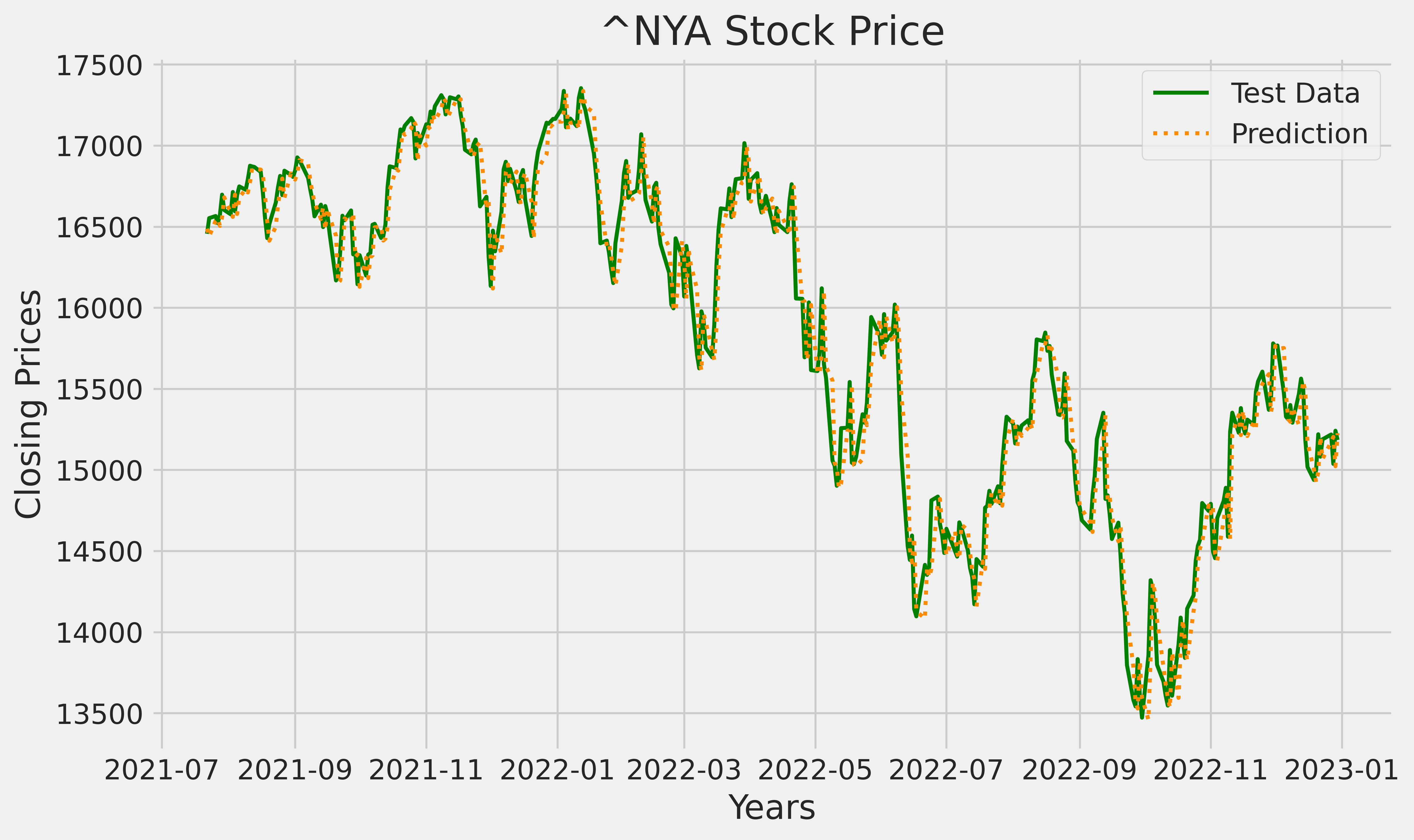}} 
    \subfigure[NASDAQ]{\includegraphics[width=0.43\textwidth]{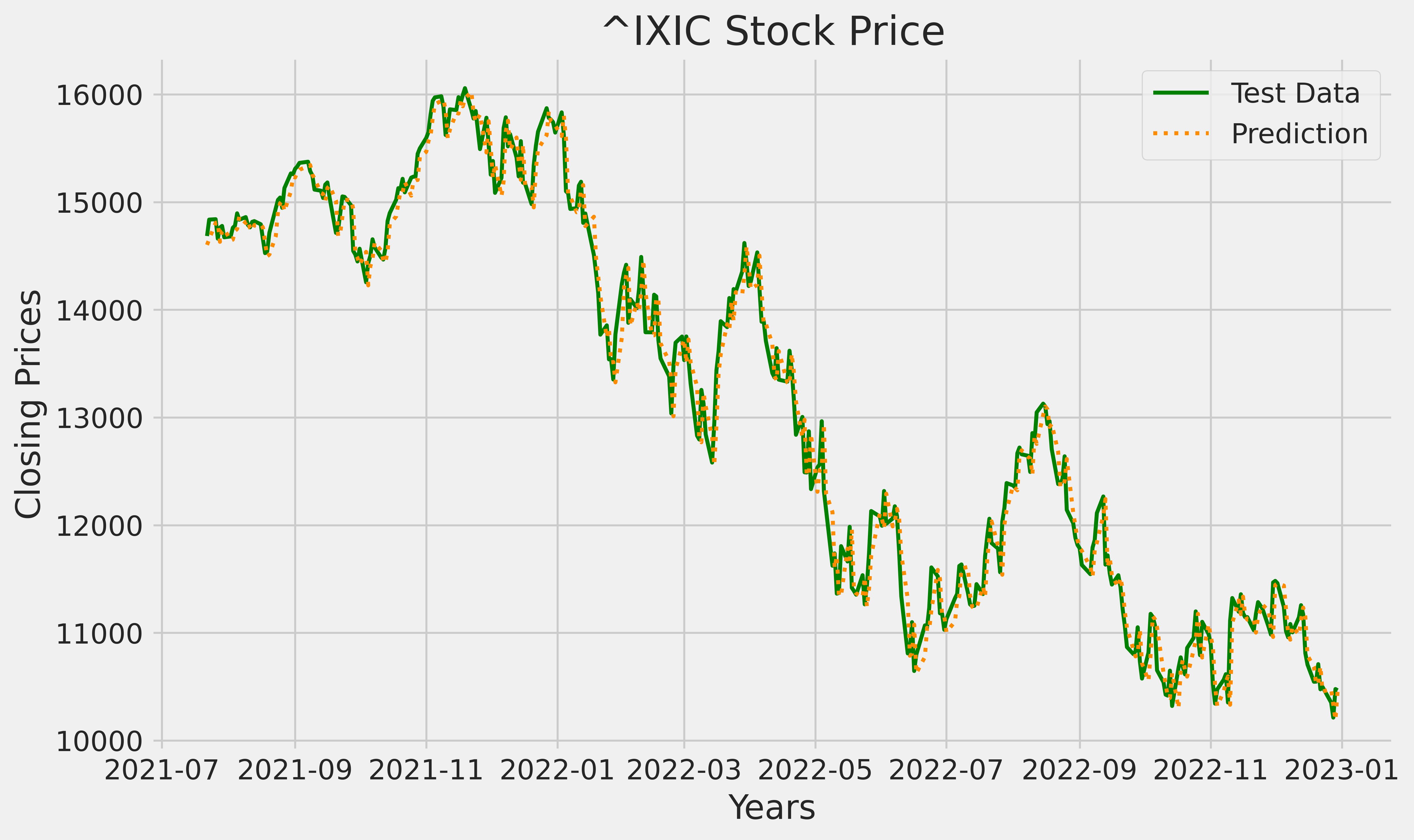}} 
    \caption{Test and predicted data (zoomed in for clarity). Subplots highlight alignment between actual (blue/green lines) and predicted (orange line) stock prices for S\&P 500, DJI, NYSE, and NASDAQ indices.}
    \Description{}
    \label{fig: Test pred}
\end{figure*}

The graphical representation of VertiFuseX’s performance is depicted in  Fig. \ref{fig: Train_test pred} and \ref{fig: Test pred}, which show predicted stock prices versus actual values over time. The consistent superiority across indices and metrics indicates that the bidirectional and stacked temporal blocks capture complementary long and short-range dependencies. The vertical fusion of penultimate-layer outputs enables the model to integrate multi-scale temporal patterns such as local price fluctuations and broader market trends without introducing redundant representations. This design keeps important features that help the model make accurate and reliable forecasts. As a result, VertiFuseX performs well in both stable and volatile market conditions.

\begin{figure}[htbp]
  \centering
  \subfigure[S\&P 500]{\includegraphics[width=0.43\linewidth]{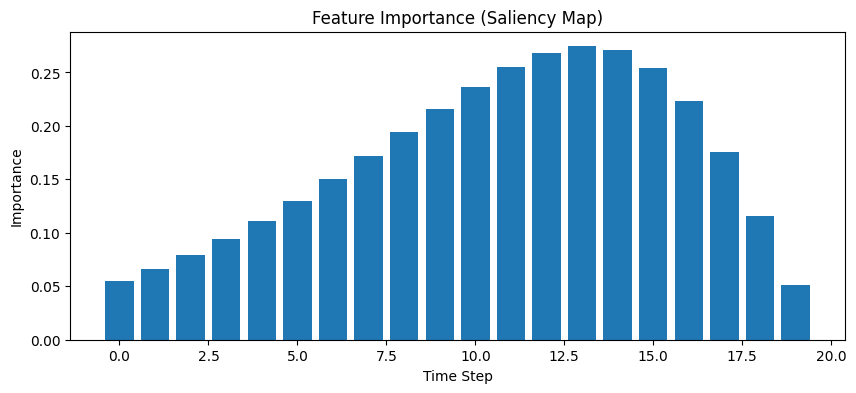} \label{fig:baseline_SP500}}
  \subfigure[NYSE]{\includegraphics[width=0.43\linewidth]{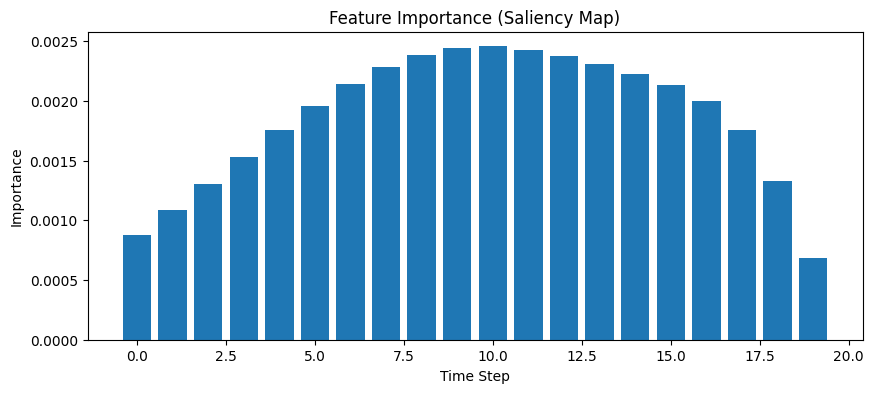}\label{fig:baseline_NYSE}}
  \subfigure[NASDAQ]{\includegraphics[width=0.43\linewidth]{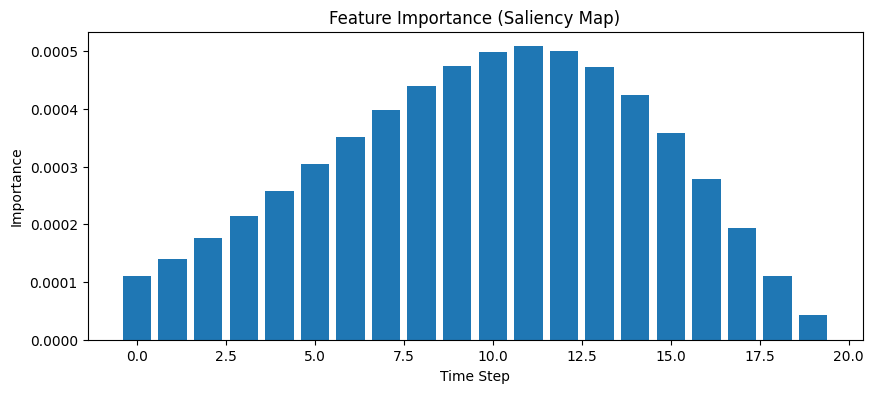}\label{fig:baseline_NASDAQ}}
  \subfigure[DJI]{\includegraphics[width=0.43\linewidth]{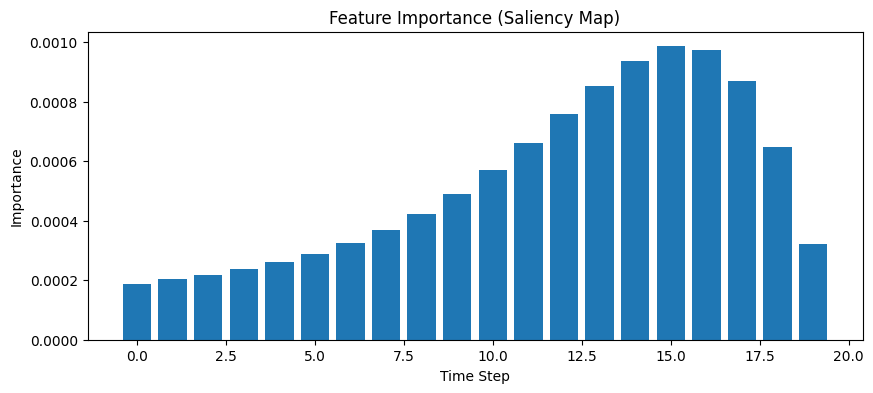}\label{fig:baseline_DJI}}
  \caption{Gradient-based saliency maps for VertiFuseX on the four benchmark indices.  Higher bars denote greater influence of the corresponding lag on the one-step-ahead forecast.}
  \Description{}
  \label{fig:saliency_baseline}
\end{figure}

To understand how VertiFuseX attends to historical inputs, we compute gradient-based saliency maps \cite{rebuffi2020there} over the 20-day look-back window. These maps visualize the absolute gradient magnitudes, $\lvert\partial\hat{y}/\partial x_t\rvert$, at each time step $t$, where $\hat{y}$ is the predicted output and $x_t$ denotes the input at lag $t$, highlighting which days exert the greatest influence on the model’s prediction. Fig.~\ref{fig:saliency_baseline} presents the feature importance (saliency) across multiple indices. Gradient-based saliency is used because it directly measures the sensitivity of the forecast to each input lag without requiring a surrogate model or sampling procedure. 

For the S\&P 500 (Fig.~\ref{fig:baseline_SP500}), the importance curve peaks around lags 12–14, indicating that the model relies on approximately ~3 weeks of accumulated context before recent information becomes dominant. This pattern aligns with the index’s diversified composition, where broader market trends tend to unfold gradually.  The NYSE Composite (Fig.~\ref{fig:baseline_NYSE}) peaks slightly earlier (lags 9–11), then plateaus, suggesting a stronger role for older data, possibly due to the faster mean-reversion behavior of large-cap constituents. The NASDAQ Composite (Fig.~\ref{fig:baseline_NASDAQ}) exhibits a compact, symmetric peak centered around lags 10–12, rapidly declining afterward. This reflects high sensitivity to recent trends, fitting for volatile, growth-heavy tech stocks where new information rapidly overshadows the past. In contrast, the DJI (Fig.~\ref{fig:baseline_DJI}) displays a delayed peak (lags 14–16), implying greater weight on the most recent trading days when predicting this index. This is consistent with the slower-moving dynamics of blue-chip firms.

These differences reveal that VertiFuseX exhibits inference-time responsiveness to differing market regimes. Without the need for retraining or hyperparameter adjustments, the vertical fusion mechanism conditionally shifts its temporal sensitivity. It places greater emphasis on recent volatility for high-beta inputs (like NASDAQ) while utilizing deeper historical context for more stable, mean-reverting assets (such as DJI). This behavior emerges from data-dependent reweighting of the LSTM (short-term), Bi-LSTM (contextual), and St-LSTM (hierarchical) representations via the learned fusion matrix $W_{\text{fuse}}$, rather than from explicit regime detection or online learning. Importantly, the term “noise” is used here in an operational rather than a strict statistical sense. Lower saliency assigned to very recent lags (t>17) and very old lags (t<3) does not imply these observations are inherently noisy. Rather, it indicates they have decreased predictive relevance for one-step-ahead forecasting. In contrast, mid-range lags (approximately 9–15 days) consistently show higher gradient attribution, suggesting that aggregated temporal context over this horizon contributes more reliably to predictions. This pattern reflects VertiFuseX’s learned reliance on temporally stable signals rather than an a priori assumption about recent observations. The 9–15 day attribution peak aligns with common intermediate-horizon technical analysis windows (e.g., 14-day lookbacks), yet emerges purely from data-driven learning in VertiFuseX, balancing noise reduction and trend characterization without predefined indicators. This shift in attention toward mid-range lags is a direct consequence of the vertical fusion mechanism.

{The concentration around lags 9 to 15 also overlaps the common two-to-three-week monitoring horizon used in short-term technical analysis, including the frequently used 14-day window. We offer this correspondence as a practical interpretation of model sensitivity rather than as evidence of a causal market cycle or a standalone trading rule.}
By accessing uncompressed feature vectors rather than final predictions, the fusion layer effectively learns a denoising transformation, suppressing high-variance features associated with immediate volatility (lags 17–20) in favor of stable predictive signals present in the penultimate representations of the St-LSTM and Bi-LSTM branches. This contrasts with final-layer fusion methods, which typically overfit to the most recent lags due to the loss of intermediate temporal resolution. Collectively, these findings affirm VertiFuseX’s robustness and adaptability across heterogeneous equity markets.

\subsection{Comparison of VertiFuseX with state-of-the-art models}\label{subsec:sota_comp}
To assess the robustness and generalizability of VertiFuseX, we compare it against seven state-of-the-art models: BiCuDNNLSTM-1dCNN \citep{kanwal2022bicudnnlstm}, ModAugNet \citep{baek2018modaugnet}, Reservoir Computing \citep{wang2021stock}, NuNet \citep{lee2020stock}, StockNet \citep{gupta2022stocknet}, DE-ABC-Bi-LSTM-ARIMA \citep{kumar2022three}, and GA-CNN-LSTM \citep{baek2023cnn}. Since prior studies use different datasets and timeframes, we retrain VertiFuseX on the exact index and duration reported in each work to ensure fair comparison. This 'dataset-centric' protocol eliminates the bias often introduced by selecting favorable time windows. By aligning our training and testing horizons exactly with those of the reference studies, ranging from the multi-decade history used for DE-ABC-Bi-LSTM-ARIMA model (1991–2010) to the recent volatile periods analyzed for KOSPI (2014–2022). By doing this, we are able to compare VertiFuseX against the state-of-the-art models under the same market volatility conditions.

Furthermore, the selected state-of-the-art models represent a diverse methodological spectrum, including evolutionary hybrids, reservoir computing, and deep ensembles, supporting the robustness of the proposed approach across diverse modeling paradigms. This evaluation covers various market regimes, including crisis periods, post-crisis recovery phases, and stable bull markets. This approach ensures that performance gains reflect inherent architectural generalization rather than benefiting from specific favorable timeframes. Our evaluation spans ten major global indices (S\&P 500, DAX, HSI, NYSE, DJI, NASDAQ, FTSE 100, KOSPI, NIKKEI 225, and NSE) covering markets across three continents. This selection constitutes a stratified evaluation set capturing diverse global market behaviors rather than a single economic regime. It encompasses geographic diversity by including regions such as North America, Europe, and Asia to mitigate region-specific overfitting. Volatility Profiles ranging from high-beta technology sectors (NASDAQ, KOSPI) to stable blue-chip indices (DJI, FTSE 100). Furthermore, it contrasts established developed markets with emerging high-growth indices, including NSE and HSI, to provide a comprehensive overview. By validating on this heterogeneous set, we demonstrate that VertiFuseX captures universal temporal dynamics, such as volatility clustering and regime shifts, rather than specific patterns unique to a single economy or sector.

\begin{figure}[htbp]
  \centering
  \includegraphics[width=0.8\linewidth]{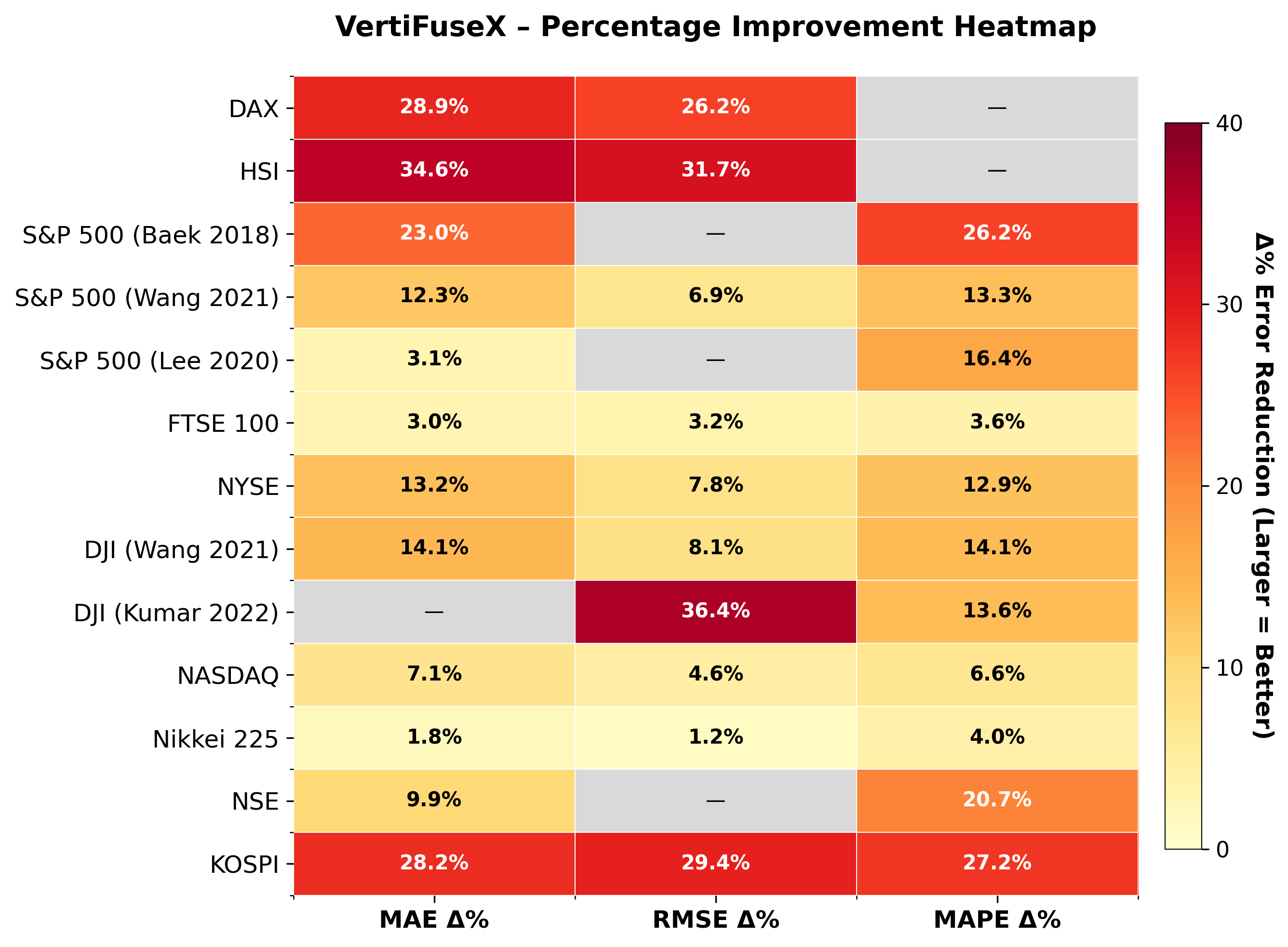}
  \caption{Heatmap of percentage error reductions (\(\Delta\%\)) achieved by VertiFuseX versus state-of-the-art models.}
  \Description{}
  \label{fig:vertifusex heatmap}
\end{figure}

MAE, RMSE, and MAPE are reported wherever data are available. Table~\ref{tab:merged-comparison} presents a unified comparison of absolute errors and percentage improvements. Fig.~\ref{fig:vertifusex heatmap} provides a color-coded heatmap summarizing these results, where lighter shades (yellow) indicate smaller improvements, darker shades (red) denote larger gains, and grey cells (\(\text{—}\)) represent metrics not reported in the original studies.  For the directly implemented baselines, we additionally report the Diebold--Mariano (DM) test of equal predictive accuracy using squared-error loss and the small-sample correction of \cite{harvey1997testing}. The loss differential is defined as $d_t = e^2_{\mathrm{baseline},t} - e^2_{\mathrm{VertiFuseX},t}$. Thus, a positive DM statistic indicates lower squared forecast loss for VertiFuseX. Table~\ref{tab:dm-R2} reports the DM statistics and two-sided $p$-values against each baseline over the 365-day out-of-sample period.

\begin{table}[htbp]
\centering
\caption{Diebold--Mariano test results: VertiFuseX vs.\ baselines using squared-error loss with Harvey--Leybourne--Newbold small-sample correction}
\label{tab:dm-R2}
\footnotesize
\setlength{\tabcolsep}{5pt}
\begin{tabular}{lcccc}
\toprule
\textbf{Baseline} & \textbf{S\&P 500} & \textbf{DJIA} & \textbf{NYSE} & \textbf{NASDAQ} \\
\midrule
ARIMA$(p,d,q)$     
& 3.96 (0.0001) 
& 3.71 (0.0002) 
& 4.18 ($<0.0001$) 
& 3.58 (0.0004) \\
LSTM               
& 3.12 (0.0020) 
& 2.85 (0.0046) 
& 3.37 (0.0008) 
& 2.79 (0.0055) \\
Bi-LSTM            
& 3.28 (0.0011) 
& 2.97 (0.0032) 
& 3.19 (0.0015) 
& 2.91 (0.0038) \\
St-LSTM            
& 2.89 (0.0041) 
& 2.74 (0.0064) 
& 2.96 (0.0033) 
& 2.68 (0.0077) \\
\bottomrule
\end{tabular}

\vspace{5pt}
{\footnotesize
Numbers are reported as DM statistic followed by two-sided $p$-value in parentheses. Positive values favour VertiFuseX because the loss differential is defined as $d_t = e^2_{\mathrm{baseline},t} - e^2_{\mathrm{VertiFuseX},t}$. All comparisons are statistically significant at the 1\% level.
}
\end{table}

{The DM test results provide statistical evidence that the performance gains of VertiFuseX reported in Table~\ref{tab:eva_compact} are robust under a formal predictive-accuracy comparison. The largest test statistics are observed against ARIMA, consistent with the substantial reductions over the classical linear benchmark. Importantly, VertiFuseX remains statistically superior to St-LSTM, the strongest single-branch recurrent baseline, across all four indices. These findings support the value added by the proposed multi-stream penultimate-layer vertical fusion.}

{To verify that the reported gains are not seed-specific, VertiFuseX was trained across three independent random seeds (42, 101, and 150) using identical chronological splits, hyperparameters, and preprocessing. Table~\ref{tab:ci-R2} reports the seed-wise results together with the mean and 95\% confidence interval across seeds. Since only three seeds are used, the interval is computed using the Student-$t$ multiplier $t_{0.975,2}=4.303$ rather than a normal approximation.}

\begin{table}[htbp]
\centering
\caption{Multi-seed robustness of VertiFuseX on the four baseline indices (seeds 42, 101, 150)}
\label{tab:ci-R2}
\footnotesize
\setlength{\tabcolsep}{6pt}
\begin{tabular}{llccc}
\toprule
\textbf{Index} & \textbf{Seed} & \textbf{MAE} & \textbf{RMSE} & \textbf{MAPE} \\
\midrule
\multirow{4}{*}{S\&P 500}
& 42   & 29.74 & 39.88 & 0.58 \\
& 101  & 31.08 & 41.36 & 0.61 \\
& 150  & 30.22 & 40.57 & 0.59 \\
& Mean $\pm$ 95\% CI & 30.35 $\pm$ 1.69 & 40.60 $\pm$ 1.84 & 0.59 $\pm$ 0.04 \\
\midrule
\multirow{4}{*}{DJIA}
& 42   & 198.76 & 266.84 & 0.51 \\
& 101  & 205.40 & 275.18 & 0.53 \\
& 150  & 201.93 & 270.60 & 0.52 \\
& Mean $\pm$ 95\% CI & 202.03 $\pm$ 8.25 & 270.87 $\pm$ 10.38 & 0.52 $\pm$ 0.02 \\
\midrule
\multirow{4}{*}{NYSE}
& 42   & 91.86 & 118.70 & 0.52 \\
& 101  & 95.12 & 122.31 & 0.54 \\
& 150  & 93.34 & 120.43 & 0.53 \\
& Mean $\pm$ 95\% CI & 93.44 $\pm$ 4.06 & 120.48 $\pm$ 4.49 & 0.53 $\pm$ 0.02 \\
\midrule
\multirow{4}{*}{NASDAQ}
& 42   & 131.96 & 177.95 & 0.81 \\
& 101  & 137.82 & 184.12 & 0.85 \\
& 150  & 134.54 & 180.87 & 0.83 \\
& Mean $\pm$ 95\% CI & 134.77 $\pm$ 7.30 & 180.98 $\pm$ 7.67 & 0.83 $\pm$ 0.05 \\
\bottomrule
\end{tabular}

\vspace{5pt}
{\footnotesize
The 95\% confidence interval is computed as $\bar{x} \pm t_{0.975,2} \cdot s / \sqrt{3}$, where $s$ is the sample standard deviation across the three seeds. Seed 150 corresponds to the main results reported in Table~\ref{tab:eva_compact}.
}
\end{table}

{The multi-seed analysis shows that VertiFuseX is stable across random initializations. The seed-wise variation is small relative to the performance gap versus LSTM, Bi-LSTM, St-LSTM, and ARIMA reported in Table~\ref{tab:eva_compact}. Even the upper bounds of the 95\% confidence intervals remain below the strongest single-branch baseline errors across all indices and metrics, supporting the robustness of the proposed multi-stream penultimate-layer vertical fusion.}

The results are discussed \textit{market by market}, highlighting key improvements in error metrics.
On the DAX index \mbox{(01/01/2000–16/06/2021)}, VertiFuseX outperformed BiCuDNNLSTM-1dCNN \citep{kanwal2022bicudnnlstm} with a 28.86\% lower MAE (133.94 vs. 188.27) and a 26.21\% lower RMSE (190.87 vs. 258.65), confirming its effectiveness in moderately volatile European markets. For the HSI \mbox{(01/01/2000–25/06/2021)}, a major Asian index, VertiFuseX achieved even greater reductions of 34.63\% in MAE (268.84 vs.\ 411.27) and 31.69\% in RMSE (359.48 vs.\ 526.21), highlighting robustness under high volatility.

The S\&P 500 is utilized by three prominent DL models, namely ModAugNet \citep{baek2018modaugnet}, Reservoir Computing \citep{wang2021stock}, and NuNet \citep{lee2020stock}. 
When evaluated over the period \mbox{(01/04/2000–27/06/2017)} our approach shows a 23\% MAE reduction (9.28 vs.\ 12.05) and a substantial 26.1\% MAPE reduction (0.79 vs.\ 1.07) compared to ModAugNet \citep{baek2018modaugnet}. This holds despite the baseline’s heavy LSTM-only architecture and data-augmentation layer.
In a subsequent evaluation period spanning \mbox{(04/01/2010–31/12/2018)}, our approach yields 12.34\%, 6.92\%, and 13.33\% improvements in MAE (13.85 vs.\ 15.80), RMSE (21.64 vs.\ 23.25), and MAPE (0.52 vs.\ 0.60), respectively, over Reservoir Computing \citep{wang2021stock}. Demonstrating superior long-horizon generalisation. Furthermore, when assessed against NuNet \citep{lee2020stock} over period spanning \mbox{(24/10/2002–13/09/2018)}. The VertiFuseX reduces error with 3.1\% lower MAE (12.21 vs.\ 12.60) and 16.4\% lower MAPE (0.46 vs.\ 0.55).  This is achieved without factorial column shuffling, highlighting both efficiency and accuracy.

On the FTSE 100 \mbox{(04/01/2010–31/12/2018)}, \textit{VertiFuseX} surpassed Reservoir Computing \citep{wang2021stock} by 3.00\% in MAE (39.80 vs.\ 41.03), 3.20\% in RMSE (52.29 vs.\ 54.02), and 3.57\% in MAPE (0.54 vs.\ 0.56). While the absolute improvements are small, the gains are steady in a market where price changes are less volatile, and trading spreads are narrow. On the NYSE for a period \mbox{(31/12/2009–28/12/2018)}, MAE decreased by 13.21\% (58.98 vs.\ 67.96), RMSE by 7.80\% (88.63 vs.\ 96.13), and MAPE by 12.92\% (0.48 vs.\ 0.55) compared with Reservoir Computing \citep{wang2021stock}. The performance of VertiFuseX is more significant as the NYSE includes more than 2,000 stocks; these results show that the vertical-fusion architecture can handle very large and diverse markets.

The DJI is utilized by two DL models, namely Reservoir Computing \citep{wang2021stock} and DE-ABC-Bi-LSTM-ARIMA \citep{kumar2022three}.
When compared with \citep{wang2021stock} over the period \mbox{(31/12/2009–28/12/2018)}. The VertiFuseX achieved 14.14\% lower MAE (133.18 vs.\ 155.11) , 8.08\% lower RMSE (206.83 vs.\ 225.01), and 14.06\% lower MAPE (0.55 vs.\ 0.64).
Also, when assessed against DE-ABC-Bi-LSTM-ARIMA \citep{kumar2022three} over \mbox{(01/01/1991–31/12/2010)}, the VertiFuseX delivers a 36.4\% RMSE drop (100.84 vs.\ 158.66) and a 13.6\% MAPE drop (0.70 vs.\ 0.81). These results illustrate robustness on multi-decade horizons and highlight the limitations of over-tuned evolutionary hybrids. 
  
When considering a highly technology-driven index like the NASDAQ. The \textit{VertiFuseX} in comparison with Reservoir Computing \citep{wang2021stock} for a duration \mbox{(31/12/2009–28/12/2018)}. It achieves a 7.12\,\% lower MAE (49.45 vs.\ 53.24), 4.61\,\% lower RMSE (74.49 vs.\ 78.09), and 6.58\,\% lower MAPE (0.71 vs.\ 0.76). These results signify the effectiveness of VertiFuseX in technology-centric markets.

\begin{table*}[htbp]
\centering
\begin{adjustbox}{width=\textwidth}
\begin{threeparttable}
\caption{Unified comparison of \textit{VertiFuseX} against multiple state-of-the-art models across diverse stock index datasets. Boldface highlights the lowest error in each row. Parenthesized values indicate percentage improvements of \textit{VertiFuseX} over the corresponding baseline.}
\label{tab:merged-comparison}
\begin{tabular}{@{}l l c ccc@{}}
\toprule
\textbf{Dataset} & \textbf{Duration} & \textbf{Reference} & \textbf{MAE} & \textbf{RMSE} & \textbf{MAPE} \\
\midrule

\multirow{2}{*}{DAX} & 
\multirow{2}{*}{\begin{tabular}[c]{@{}l@{}}01/01/2000 \\ to 16/06/2021\end{tabular}} & 
\cite{kanwal2022bicudnnlstm} &
188.27 & 258.65 & --- \\
& & \textit{VertiFuseX} & 
\textbf{133.94} (\textit{28.86\%}$\downarrow$) & 
\textbf{190.87} (\textit{26.21\%}$\downarrow$) & 
--- \\
\cmidrule{1-6}

\multirow{2}{*}{HSI} & 
\multirow{2}{*}{\begin{tabular}[c]{@{}l@{}}01/01/2000 \\ to 25/06/2021\end{tabular}} & 
\cite{kanwal2022bicudnnlstm} &
411.27 & 526.21 & --- \\
& & \textit{VertiFuseX} & 
\textbf{268.84} (\textit{34.63\%}$\downarrow$) & 
\textbf{359.48} (\textit{31.69\%}$\downarrow$) & 
--- \\
\cmidrule{1-6}

\multirow{2}{*}{S\&P 500} & 
\multirow{2}{*}{\begin{tabular}[c]{@{}l@{}}01/04/2000 \\ to 27/07/2017\end{tabular}} & 
\cite{baek2018modaugnet} &
12.05 & --- & 1.07 \\
& & \textit{VertiFuseX} & 
\textbf{9.28} (\textit{23\%}$\downarrow$) & 
--- & 
\textbf{0.79} (\textit{26.1\%}$\downarrow$) \\
\cmidrule{1-6}

\multirow{2}{*}{S\&P 500} & 
\multirow{2}{*}{\begin{tabular}[c]{@{}l@{}}04/01/2010 \\ to 28/12/2018\end{tabular}} & 
\cite{wang2021stock} &
15.80 & 23.25 & 0.60 \\
& & \textit{VertiFuseX} & 
\textbf{13.85} (\textit{12.34\%}$\downarrow$) & 
\textbf{21.64} (\textit{6.92\%}$\downarrow$) & 
\textbf{0.52} (\textit{13.33\%}$\downarrow$) \\
\cmidrule{1-6}

\multirow{2}{*}{S\&P 500} & 
\multirow{2}{*}{\begin{tabular}[c]{@{}l@{}}24/10/2002 \\ to 13/09/2018\end{tabular}} & 
\cite{lee2020stock} &
12.60 & --- & 0.55 \\
& & \textit{VertiFuseX} & 
\textbf{12.21} (\textit{3.1\%}$\downarrow$) & 
--- & 
\textbf{0.46} (\textit{16.4\%}$\downarrow$) \\
\cmidrule{1-6}

\multirow{2}{*}{FTSE 100} & 
\multirow{2}{*}{\begin{tabular}[c]{@{}l@{}}04/01/2010 \\ to 31/12/2018\end{tabular}} & 
\cite{wang2021stock} &
41.03 & 54.02 & 0.56 \\
& & \textit{VertiFuseX} & 
\textbf{39.80} (\textit{3\%}$\downarrow$) & 
\textbf{52.29} (\textit{3.2\%}$\downarrow$) & 
\textbf{0.54} (\textit{3.57\%}$\downarrow$) \\
\cmidrule{1-6}

\multirow{2}{*}{NYSE} & 
\multirow{2}{*}{\begin{tabular}[c]{@{}l@{}}31/12/2009 \\ to 28/12/2018\end{tabular}} & 
\cite{wang2021stock} &
67.96 & 96.13 & 0.55 \\
& & \textit{VertiFuseX} & 
\textbf{58.98} (\textit{13.21\%}$\downarrow$) & 
\textbf{88.63} (\textit{7.8\%}$\downarrow$) & 
\textbf{0.48} (\textit{12.92\%}$\downarrow$) \\
\cmidrule{1-6}

\multirow{2}{*}{DJI} & 
\multirow{2}{*}{\begin{tabular}[c]{@{}l@{}}31/12/2009 \\ to 28/12/2018\end{tabular}} & 
\cite{wang2021stock} &
155.11 & 225.01 & 0.64 \\
& & \textit{VertiFuseX} & 
\textbf{133.18} (\textit{14.14\%}$\downarrow$) & 
\textbf{206.83} (\textit{8.08\%}$\downarrow$) & 
\textbf{0.55} (\textit{14.06\%}$\downarrow$) \\
\cmidrule{1-6}

\multirow{2}{*}{DJI} & 
\multirow{2}{*}{\begin{tabular}[c]{@{}l@{}}01/01/1991 \\ to 31/12/2010\end{tabular}} & 
\cite{kumar2022three} &
--- & 158.66 & 0.81 \\
& & \textit{VertiFuseX} & 
--- & 
\textbf{100.84} (\textit{36.4\%}$\downarrow$) & 
\textbf{0.70} (\textit{13.6\%}$\downarrow$) \\
\cmidrule{1-6}

\multirow{2}{*}{NASDAQ} & 
\multirow{2}{*}{\begin{tabular}[c]{@{}l@{}}31/12/2009 \\ to 28/12/2018\end{tabular}} & 
\cite{wang2021stock} &
53.24 & 78.09 & 0.76 \\
& & \textit{VertiFuseX} & 
\textbf{49.45} (\textit{7.12\%}$\downarrow$) & 
\textbf{74.49} (\textit{4.61\%}$\downarrow$) & 
\textbf{0.71} (\textit{6.58\%}$\downarrow$) \\
\cmidrule{1-6}

\multirow{2}{*}{Nikkei 225} & 
\multirow{2}{*}{\begin{tabular}[c]{@{}l@{}}04/01/2010 \\ to 28/12/2018\end{tabular}} & 
\cite{wang2021stock} &
160.77 & 229.35 & 0.75 \\
& & \textit{VertiFuseX} & 
\textbf{157.82} (\textit{1.83\%}$\downarrow$) & 
\textbf{226.54} (\textit{1.23\%}$\downarrow$) & 
\textbf{0.72} (\textit{4\%}$\downarrow$) \\
\cmidrule{1-6}

\multirow{2}{*}{NSE} & 
\multirow{2}{*}{\begin{tabular}[c]{@{}l@{}}01/04/1996 \\ to 06/01/2020\end{tabular}} & 
\cite{gupta2022stocknet} &
69.93 & --- & 0.82 \\
& & \textit{VertiFuseX} & 
\textbf{62.99} (\textit{9.92\%}$\downarrow$) & 
--- & 
\textbf{0.65} (\textit{20.73\%}$\downarrow$) \\
\cmidrule{1-6}

\multirow{2}{*}{KOSPI} & 
\multirow{2}{*}{\begin{tabular}[c]{@{}l@{}}14/09/2014 \\ to 31/10/2022\end{tabular}} & 
\cite{baek2023cnn} &
31.47 & 40.05 & 1.10 \\
& & \textit{VertiFuseX} & 
\textbf{22.59} (\textit{28.2\%}$\downarrow$) & 
\textbf{28.29} (\textit{29.36\%}$\downarrow$) & 
\textbf{0.80} (\textit{27.2\%}$\downarrow$) \\
\bottomrule
\end{tabular}
\begin{tablenotes}
\small
\item \emph{Notes:} Error metrics for prior models are sourced from their original publications. An em-dash (---) indicates unreported metrics. \textit{VertiFuseX} achieves the lowest error in all comparable cases; numbers in \textit{italics} denote percentage improvements over the best-reported results from state-of-the-art models.
\end{tablenotes}
\end{threeparttable}
\end{adjustbox}
\end{table*}

For the Nikkei 225 index covering the period \mbox{(04/01/2010–28/12/2018)}. The VertiFuseX improvements were modest but consistent with MAE decreasing by 1.83\% (157.82 vs.\ 160.77), RMSE by 1.23\% (226.54 vs.\ 229.35), and MAPE by 4.00\% (0.72 vs.\ 0.75) compared to Reservoir Computing \citep{wang2021stock}. Even in the lower volatility, yen denominated market all performance metrics remained steady. On India’s NSE, the StockNet \citep{gupta2022stocknet} which multiplies its training set 252-fold through data augmentation over the period \mbox{(01/04/1996–06/01/2020)}. The \textit{VertiFuseX} still achieves 9.92\% lower MAE (62.99 vs.\ 69.93) and 20.73\% lower MAPE (0.65 vs.\ 0.82). This indicates adaptability to emerging market dynamics while confirming that architectural fusion is more effective than just relying on a huge amount of data. Finally, on the KOSPI during the period (14/09/2014–31/10/2022), \textit{VertiFuseX} achieved its largest gains. When compared with GA-CNN-LSTM \citep{baek2023cnn}, VertiFuseX lowered MAE by 28.2\,\% (22.59 vs.\ 31.47), RMSE by 29.36\,\% (28.29 vs.\ 40.05), and MAPE by 27.2\% (0.80 vs.\ 1.10).

\begin{figure}[ht]
    \centering
    \includegraphics[width=\linewidth]{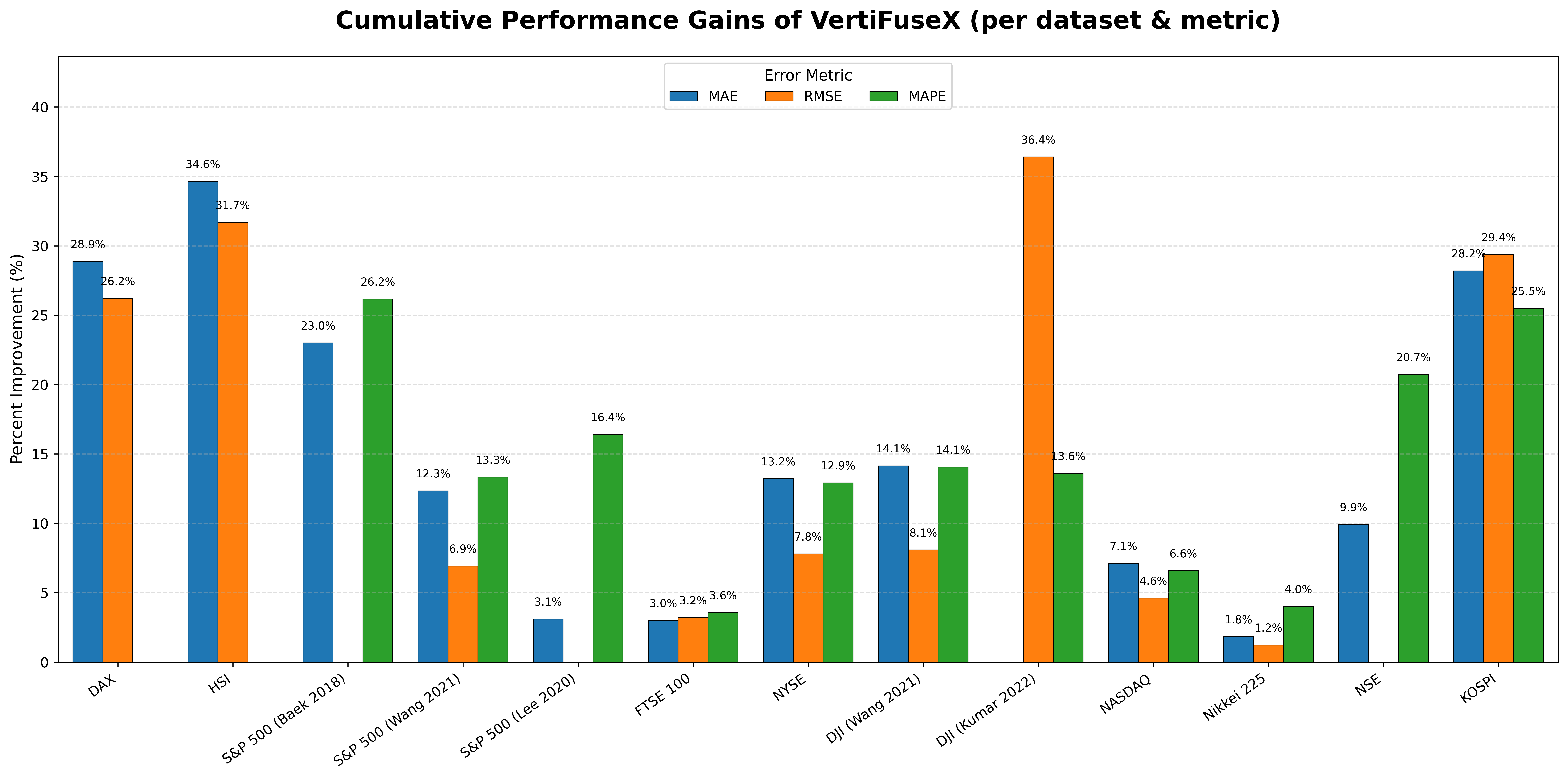}
    \caption{Cumulative percentage improvements of VertiFuseX over each state-of-the-art model across MAE, RMSE, and MAPE.}
    \Description{}
    \label{fig:cumulative_gain}
\end{figure}

To further consolidate these insights, Fig.~\ref{fig:cumulative_gain} builds on the heat map by summarizing all metric-specific improvements into a single cumulative score for each state-of-the-art method. The bar plot shows the cumulative percentage improvement achieved by \textit{VertiFuseX} over state-of-the-art methods across various financial datasets and error metrics. Each group of bars corresponds to a specific dataset, while the colors represent different error metrics, with blue for MAE, orange for RMSE, and green for MAPE. The height of each bar represents the percentage improvement ~($\Delta\%$) delivered by \textit{VertiFuseX} relative to the respective state-of-the-art method. The total height of each group of bars represents the aggregate performance gain across all three error metrics for that dataset. This visualization highlights the consistent and multi-metric superiority of VertiFuseX across diverse market indices.

In summary, VertiFuseX attains the lowest error across all 33 metric–dataset combinations in Table~\ref{tab:merged-comparison}. Improvements are consistent across \emph{both} scale-dependent (MAE/RMSE) and scale-free (MAPE) metrics, confirming that the innovative vertical multi-stream design generalizes across markets, timeframes, and error formulations. Especially large margins on volatile Asian indices (HSI, KOSPI) and on long-horizon DJI data. The uniform superiority of VertiFuseX across ten indices, three continents, and two decades of trading history signals strong practical relevance. 

\subsection{Computational Cost Analysis}
This section presents an analysis of the computational requirements of the proposed model, VertiFuseX, focusing on model complexity, training time, inference speed, and the efficiency–accuracy trade-off compared to baseline methods. These metrics are essential for evaluating the practical viability of deploying VertiFuseX in real-time financial forecasting scenarios. The VertiFuseX architecture comprises 675,716 parameters, of which 673,924 are trainable, and 1,792 are non-trainable. The model occupies approximately 2.58 MB of memory, a footprint well within the capabilities of consumer-grade CPUs and embedded edge devices. This compact design enables efficient execution within standard memory hierarchies and avoids reliance on server-grade GPUs or specialized accelerators. As a result, VertiFuseX is suitable for deployment in low-resource environments such as retail investor workstations where computational efficiency and predictable latency are critical.
All models were trained for 50 epochs with a batch size of 64. Across four major financial indices, S\&P 500, DJI, NYSE, and NASDAQ, the total training time ranged from 100.23 to 107.55 seconds, showing consistent behavior across datasets. To measure inference latency, we performed one warm-up pass (to absorb GPU kernel initialization), followed by 100 forward passes on a fixed batch of 64 samples, the same size used during training. The elapsed time is then normalized per sample to ensure meaningful comparisons across models or hardware configurations. As shown in Table~\ref{tab:comp_cost}, VertiFuseX delivers stable inference times across all datasets, ranging from 1.47 to 1.56 milliseconds per sample. This millisecond-level latency enables efficient portfolio-scale deployment via parallel execution, with computational cost scaling linearly with the number of assets, supporting real-time monitoring and risk management.

The results indicate that VertiFuseX operates near the efficiency--accuracy frontier. Unlike heavily parameterized hybrid or evolutionary architectures that rely on architectural redundancy or extensive hyperparameter optimization, VertiFuseX achieves strong predictive performance (up to 34.6\% lower MAE) through efficient penultimate-layer feature integration. Its lightweight memory footprint ($\approx$2.6 MB) contrasts with Transformer-based models, which incur higher computational and memory overhead due to self-attention operations. By avoiding attention mechanisms at inference, VertiFuseX maintains low latency and rapid training, prioritizing deployability over exhaustive global context modeling. With a deterministic inference latency of $\approx$1.5 ms per sample, the model avoids the runtime overhead typical of deep ensembles or dynamically adaptive architectures, making it well-suited for low-latency, real-time decision pipelines where both predictive accuracy and computational efficiency are essential. 

\begin{table}[ht]
\centering
\caption{Computational cost of VertiFuseX across four financial indices}
\label{tab:comp_cost}
\begin{tabular}{lcc}
\toprule
\textbf{Dataset} & \textbf{Training Time (s)} & \textbf{Inference Time (ms/sample)} \\
\midrule
S\&P\,500 & 100.23 & 1.47 \\
DJI       & 107.16 & 1.56 \\
NYSE      & 107.55 & 1.47 \\
NASDAQ    & 101.11 & 1.48 \\
\bottomrule
\end{tabular}
\end{table}

With an inference latency of $\approx 1.5$ ms, VertiFuseX can process intraday granularities (e.g., minute-level bars) without introducing latency bottlenecks in real-time trading pipelines. For real-time streaming deployment, the moving window in Fig. \ref{fig:window} operates as a fixed-size FIFO buffer that updates in $\mathcal{O}(1)$ time per new price tick. Given the measured inference latency of $\approx 1.5$~ms, the model operates well within typical low-latency trading intervals, leaving substantial margin for network and execution delays, making it suitable for live deployment.

\subsection{Algorithmic Trading Strategy and Economic Evaluation}\label{subsec:trading_Strategy}
While standard error metrics (MAE, RMSE, MAPE) quantify predictive accuracy, they provide limited insight into economic utility within stochastic financial markets. A model with low mean error may still generate negative returns if errors cluster during high-volatility regimes or misclassify directional changes during critical market inflections. We therefore evaluate VertiFuseX through a controlled trading simulation that reports risk-adjusted performance under the stated execution and risk-control assumptions. Trading signals are derived from VertiFuseX's one-step-ahead price forecasts. To filter noise and ensure capital efficiency, we employ an empirical confidence threshold based on the distribution of out-of-sample predictions. Specifically, a long position is initiated only when the predicted price change exceeds the 25th percentile of the test-period forecast distribution, selecting approximately the top 75\% of predicted movements:
\begin{equation}
    S_t = \mathbb{I}(\hat{y}_t > \tau_{25}) \cdot \delta_t
\end{equation}
where $\mathbb{I}(\cdot)$ is the indicator function and $\delta_t \in \{0,1\}$ is a risk-gating variable defined below. 
Although VertiFuseX is optimized using an MSE objective,the fusion at the penultimate layer helps prevent the mean-convergence issues common in scalar regression models. 
By preserving multi-scale variance in the combined representation $\mathbf{F}_{\mathrm{combined}}$ until the final projection, the resulting forecast $\hat{y}_t$ retains sufficient discriminatory power for directional signaling. 
This enables effective quantile-based thresholding ($S_t = \mathbb{I}(\hat{y}_t > \tau_{25})$) without requiring iterative multi-step simulation or auxiliary classification losses.

Execution adheres to institutional trading realities: orders entered at market close on day $t-1$ are filled at the next-day opening price $P_t^{open}$, incurring overnight gap risk. A proportional transaction cost of $\psi = 10$ basis points (0.10\%) is applied per trade to account for slippage and commissions. No leverage is employed, and the portfolio alternates between fully invested and fully cash positions. Position management follows an augmented Triple-Barrier Method (TBM) \citep{de2018advances}. For an entry at price $P_{entry}$, the trade exits if any condition is met:
\begin{enumerate}
    \item \textbf{Take-Profit (TP):} Unrealized return $r_t \geq +6\%$
    \item \textbf{Stop-Loss (SL):} $r_t \leq -3\%$ (downside truncation)
    \item \textbf{Horizon Limit:} Signal cessation (i.e., no positive forecast) after a minimum holding period of three trading days.
\end{enumerate}

Empirical analysis reveals that prediction errors exhibit temporal clustering: when the model misinterprets a regime shift, directional failures persist for multiple periods. To mitigate this "error autocorrelation", we introduce a cool-down mechanism that activates exclusively after stop-loss exits:
\begin{equation}
    \delta_{t} =
    \begin{cases}
        0 & \text{if a stop-loss exit occurred on any of the previous 3 trading days} \\
        1 & \text{otherwise}
    \end{cases}
\end{equation}
This imposes a mandatory 3-day trading pause following any risk-managed loss. Unlike volatility scaling that reacts to market magnitude, this mechanism responds to model failure, preventing ``revenge trading'' during error clusters. Formally, if drawdown $DD = \sum_{k=1}^{m} L_k$ where $L_k$ are consecutive losses, the cool-down reduces cluster size $m$ rather than loss magnitude $L_k$, directly targeting the dominant contributor to strategy drawdowns. The fixed three-day horizon was chosen conservatively to cover the observed short-term error clusters (typically 2-3 days) while remaining simple and interpretable. Adaptive or index-specific cool-down periods are left for future investigation. We evaluate the strategy across four U.S. indices (S\&P 500, DJIA, NASDAQ, NYSE) during the 365-day out-of-sample period ending December 31, 2024. Table~\ref{tab:trading_results} compares VertiFuseX against buy-and-hold and Naive momentum benchmarks.

\begin{table}[ht]
\centering
\caption{Economic Performance: VertiFuseX vs. Benchmarks (2024 Out-of-Sample)}
\label{tab:trading_results}
\resizebox{\textwidth}{!}{%
\begin{tabular}{@{}lccccccccc@{}}
\toprule
\textbf{Index} & \textbf{Strategy} & \textbf{Total Return (\%)} & \textbf{Sharpe} & \textbf{Sortino} & \textbf{Max DD (\%)} & \textbf{Win Rate (\%)} & \textbf{Exposure (\%)} & \textbf{Trades} \\
\midrule
\multirow{3}{*}{\textbf{S\&P 500}}
& VertiFuseX & 29.41 & \textbf{1.43} & 1.85 & \textbf{-9.75} & 67.7 & 74.8 & 31 \\
& Buy-and-Hold & \textbf{29.44} & 1.37 & \textbf{1.89} & -10.28 & 100.0 & 100.0 & 1 \\
& Naive Momentum & -5.08 & -0.44 & -0.52 & -11.28 & 46.7 & 56.7 & 92 \\
\midrule
\multirow{3}{*}{\textbf{DJIA}}
& VertiFuseX & \textbf{21.44} & \textbf{1.15} & \textbf{1.75} & \textbf{-8.94} & 57.7 & 74.8 & 26 \\
& Buy-and-Hold & 21.08 & 1.06 & 1.63 & -9.02 & 100.0 & 100.0 & 1 \\
& Naive Momentum & 0.18 & -0.19 & -0.26 & -9.92 & 52.9 & 56.7 & 85 \\
\midrule
\multirow{3}{*}{\textbf{NASDAQ}}
& VertiFuseX & 26.59 & 1.04 & 1.27 & \textbf{-11.97} & 64.9 & 74.8 & 37 \\
& Buy-and-Hold & \textbf{36.25} & \textbf{1.27} & \textbf{1.74} & -13.15 & 100.0 & 100.0 & 1 \\
& Naive Momentum & -0.39 & 0.00 & 0.00 & -14.89 & 50.6 & 58.4 & 85 \\
\midrule
\multirow{3}{*}{\textbf{NYSE}}
& VertiFuseX & 7.68 & 0.35 & 0.51 & -13.14 & 48.6 & 74.8 & 37 \\
& Buy-and-Hold & \textbf{17.31} & \textbf{0.88} & \textbf{1.32} & \textbf{-10.66} & 100.0 & 100.0 & 1 \\
& Naive Momentum & 4.92 & 0.18 & 0.26 & -9.04 & 56.5 & 53.7 & 85 \\
\bottomrule
\end{tabular}
}
\end{table}

\begin{figure}[htbp]
  \centering
  \subfigure[S\&P 500]{
    \includegraphics[width=0.47\linewidth]{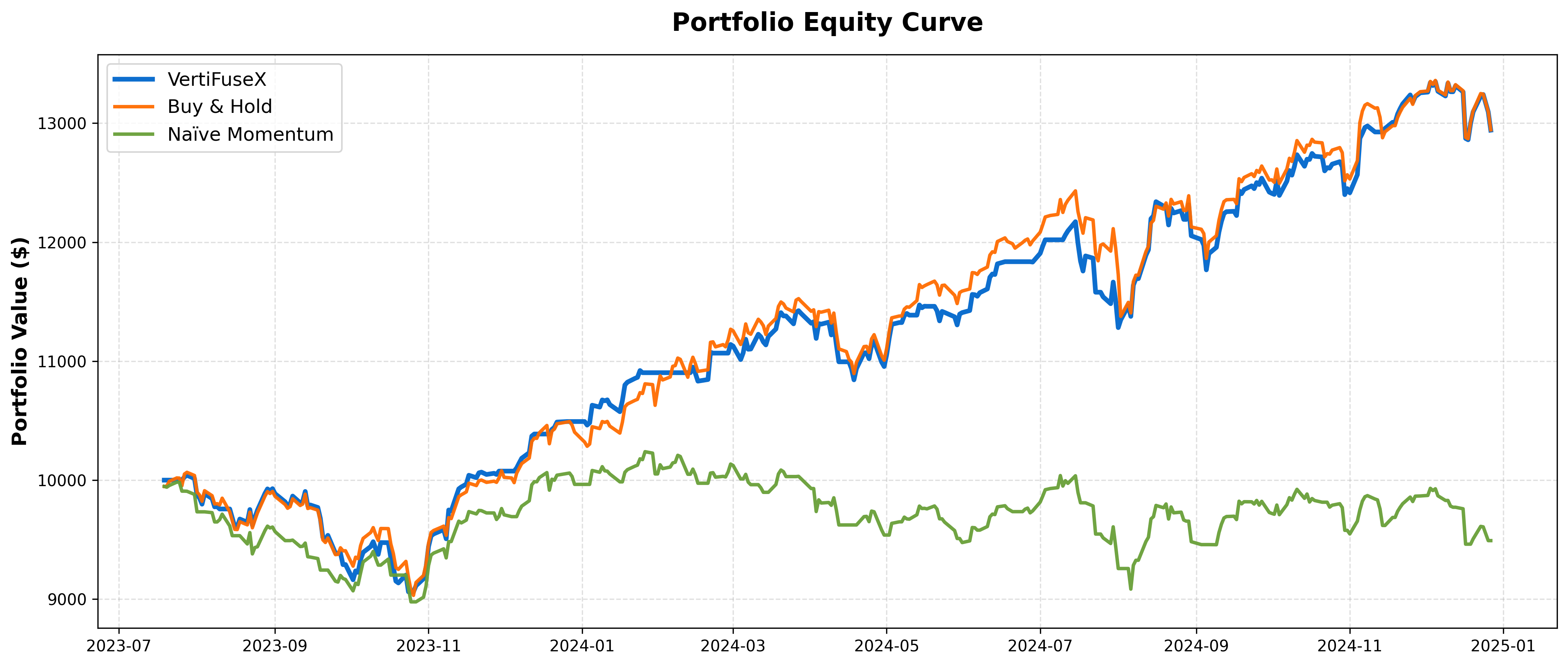}
    \label{fig:trading_sp500}
  }
  \hfill
  \subfigure[DJI]{
    \includegraphics[width=0.47\linewidth]{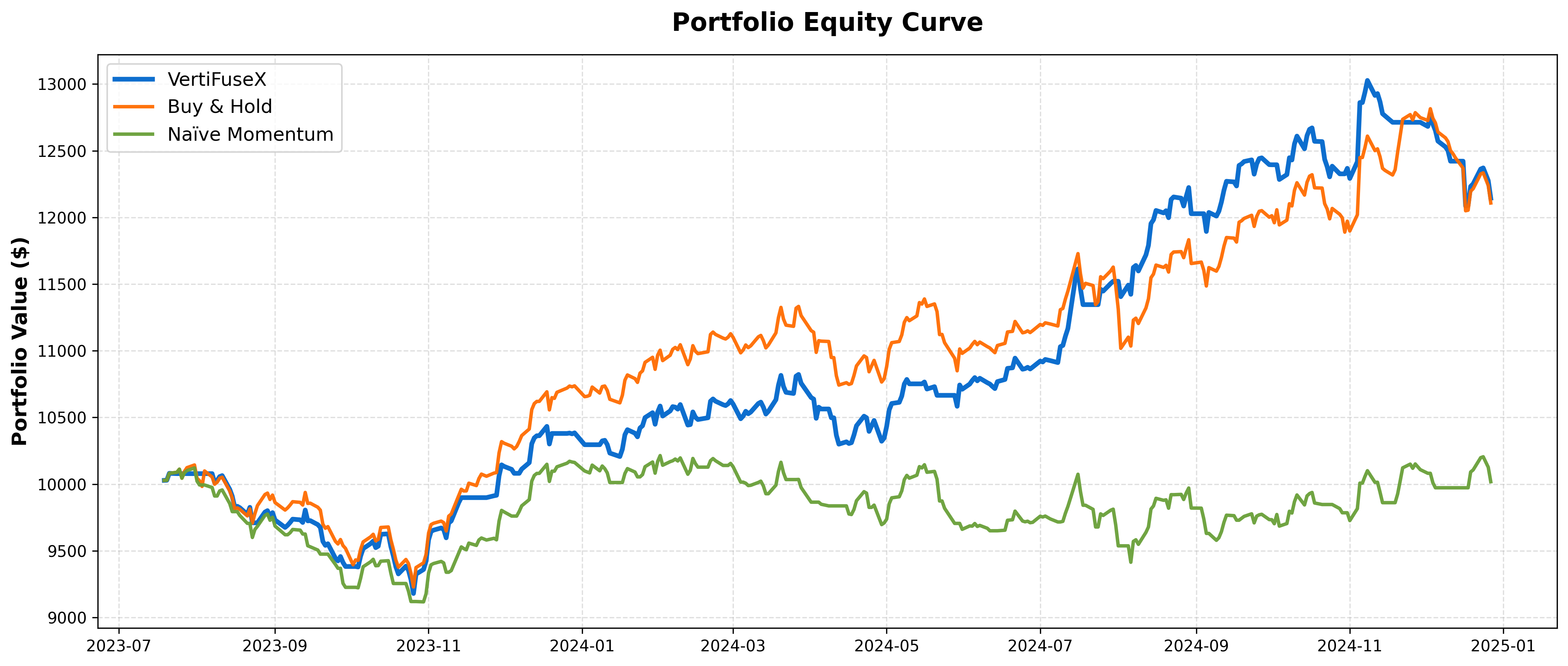}
    \label{fig:trading_dji}
  }
  \\[2em]
  \subfigure[NYSE]{
    \includegraphics[width=0.47\linewidth]{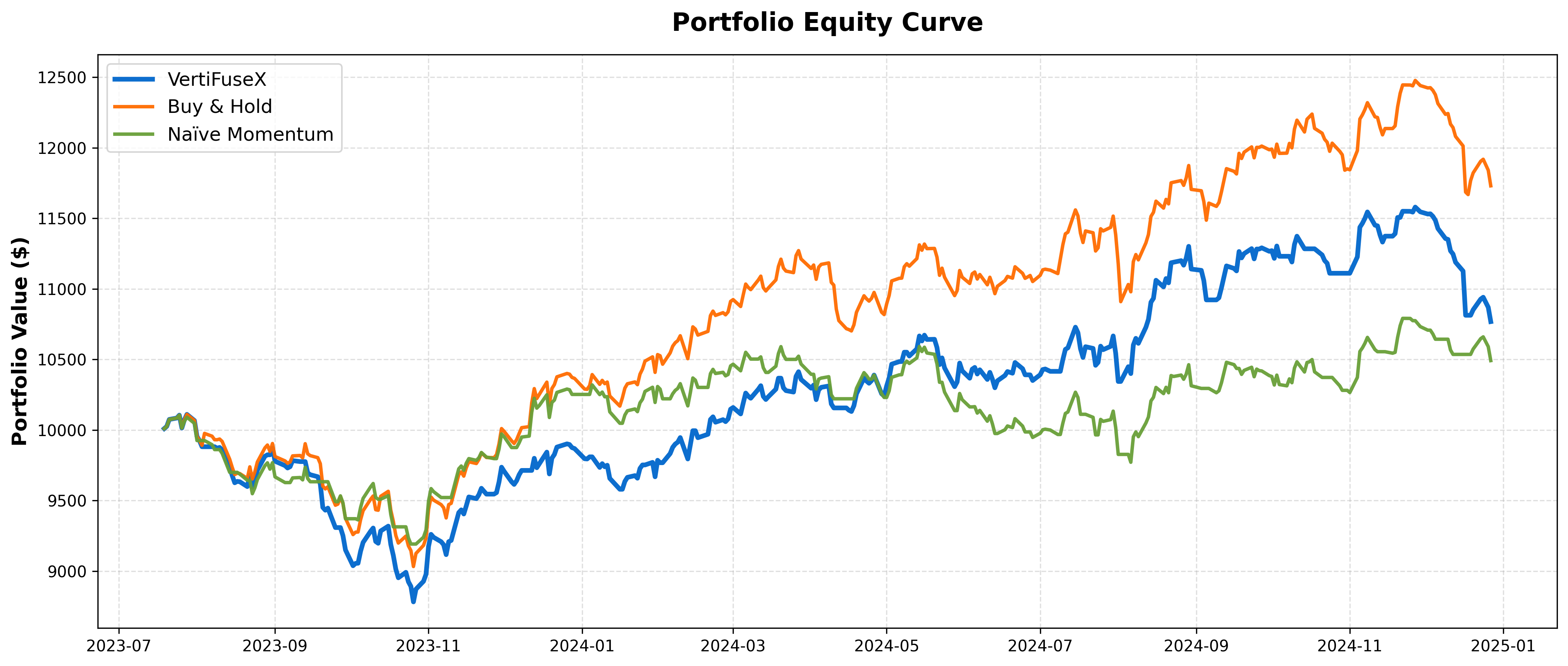}
    \label{fig:trading_nyse}
  }
  \hfill
  \subfigure[NASDAQ]{
    \includegraphics[width=0.47\linewidth]{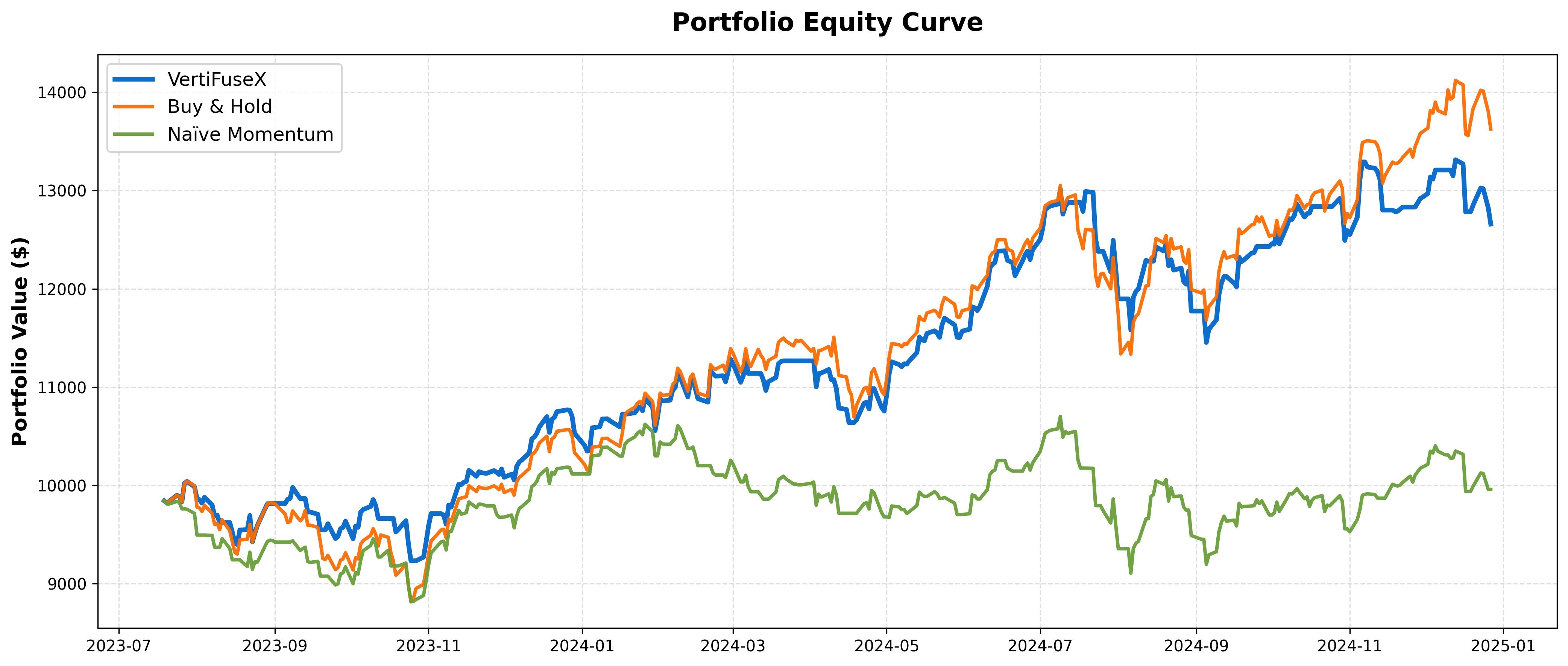}
    \label{fig:trading_nasdaq}
  }
  \caption{Equity curves of the proposed trading strategy on the four benchmark indices (S\&P 500, DJI, NYSE, and NASDAQ) in hand-drawn style.}
  \Description{}
  \label{fig:trading_equity_curves}
\end{figure}

The results demonstrate three key insights, as illustrated in the equity curves (Fig.~\ref{fig:trading_equity_curves}). On S\&P 500 and DJI, VertiFuseX achieves near-identical total returns to buy-and-hold (29.41\% vs. 29.44\% and 21.44\% vs. 21.08\%) while maintaining a cash reserve of approximately 25\% on average. This translates to superior risk-adjusted returns (Sharpe ratios of 1.43 vs. 1.37 and 1.15 vs. 1.06), confirming that the strategy filters non-productive volatility while maintaining participation in upward trends.

\begin{figure}[htbp]
  \centering
  \subfigure[S\&P 500]{
    \includegraphics[width=0.47\linewidth]{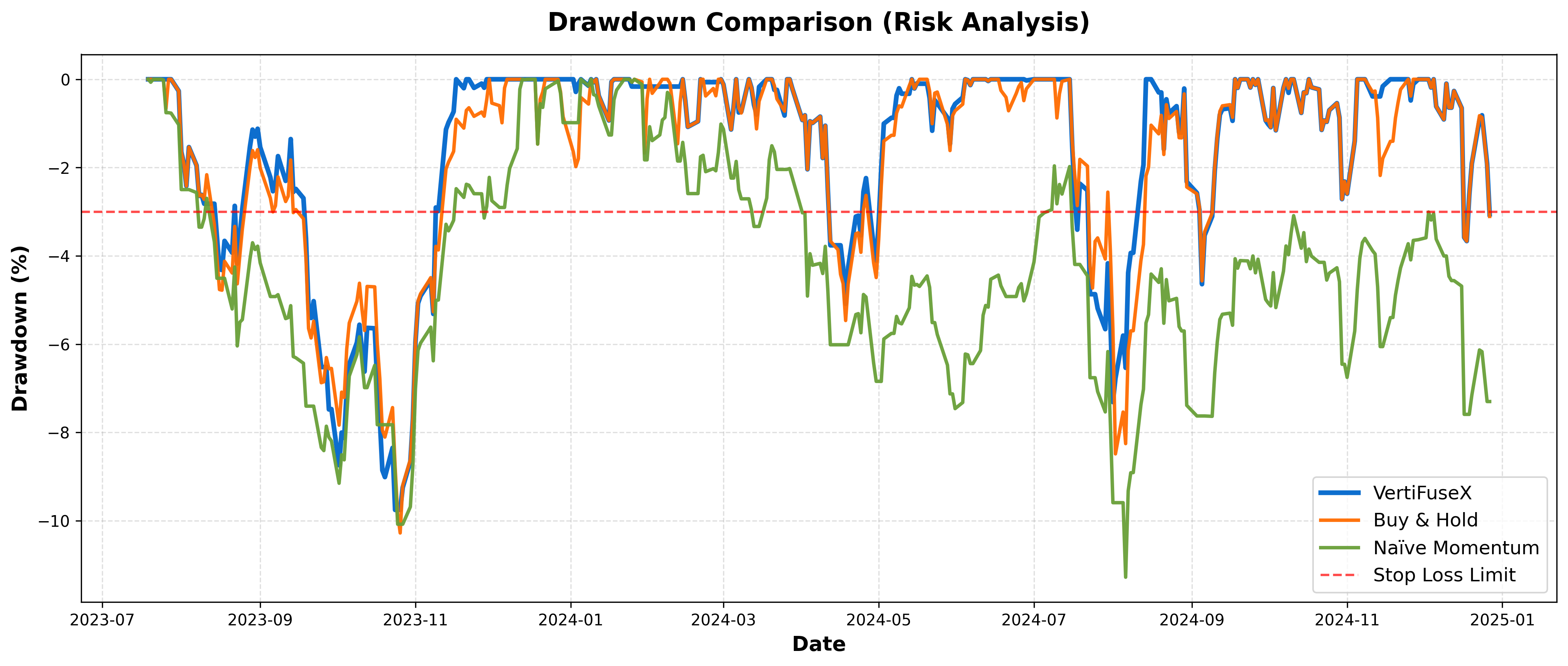}
    \label{fig:drawdown_sp500}
  }
  \hfill
  \subfigure[DJI]{
    \includegraphics[width=0.47\linewidth]{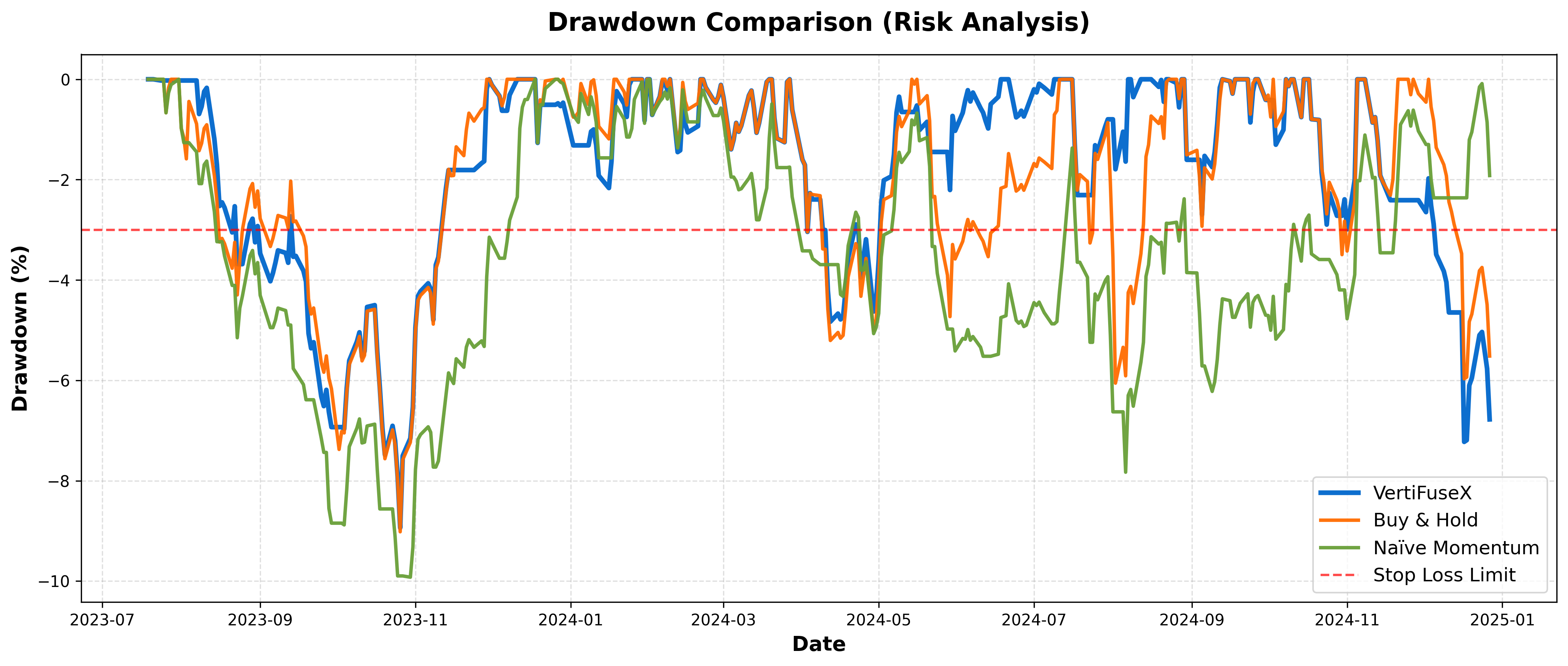}
    \label{fig:drawdown_dji}
  }
  \\[2em]
  \subfigure[NYSE]{
    \includegraphics[width=0.47\linewidth]{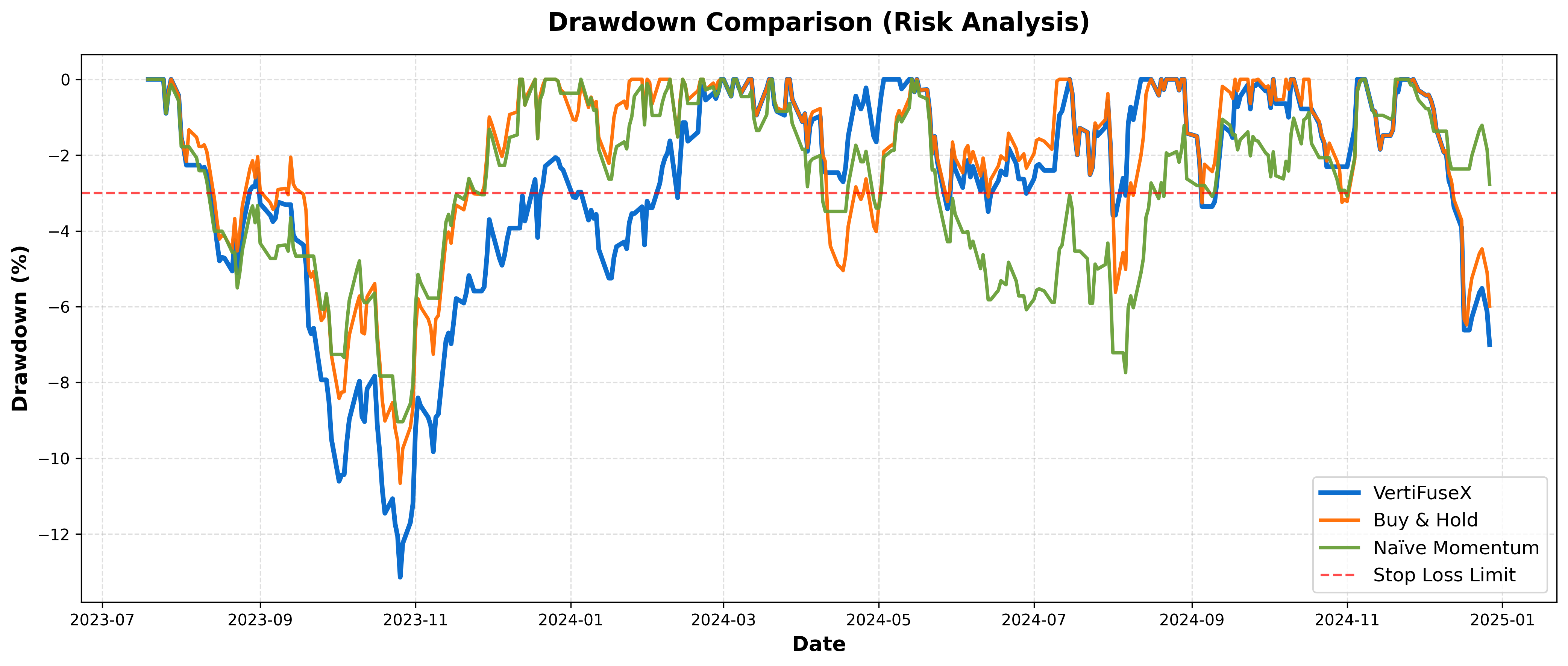}
    \label{fig:drawdown_nyse}
  }
  \hfill
  \subfigure[NASDAQ]{
    \includegraphics[width=0.47\linewidth]{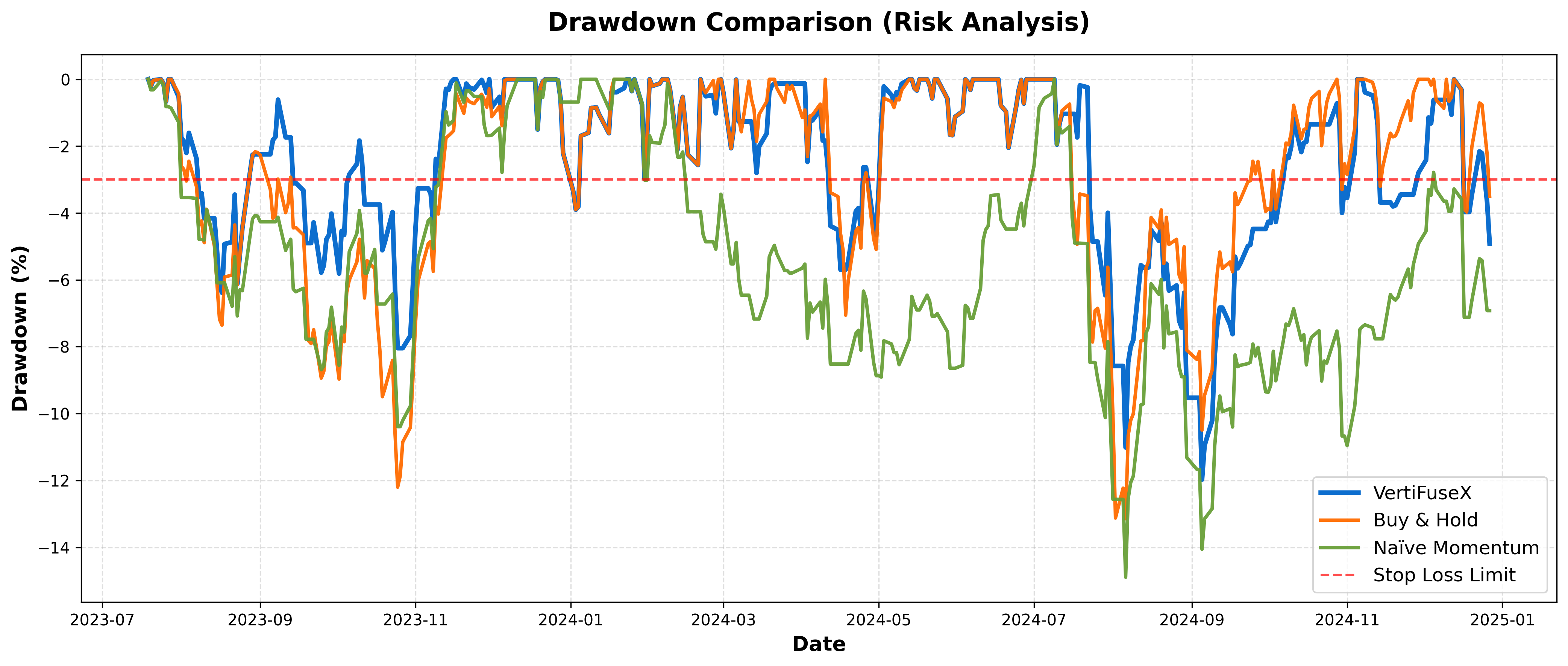}
    \label{fig:drawdown_nasdaq}
  }
  \caption{Drawdown comparison (risk analysis) for the proposed VertiFuseX strategy compared with Buy \& Hold, Naive Momentum, and a stop-loss limit on the four benchmark indices.}
  \Description{}
  \label{fig:drawdown_curves}
\end{figure}

The cool-down mechanism proves most effective in volatile regimes, as clearly visible in the drawdown comparisons (Fig.~\ref{fig:drawdown_curves}). On NASDAQ, while VertiFuseX trails in absolute return (26.59\% vs. 36.25\%), it reduces maximum drawdown by 1.18 percentage points (from -13.15\% to -11.97\%, an 8.97\% relative improvement). Examination of exit sequences reveals that stop-loss events frequently occurred in short clusters of 2–3 consecutive days. The cool-down mechanism systematically prevented immediate re-entry during these periods, thereby attenuating compound losses. Performance variation reflects differences in index composition and trend persistence. In particular, the broader NYSE Composite, with its higher sectoral heterogeneity and inclusion of smaller-cap stocks, exhibits lower trend persistence and stronger mean-reverting tendencies compared to the more concentrated, growth-oriented large-cap indices (S\&P 500, DJIA, NASDAQ). {NYSE is the weakest case in the study. VertiFuseX returns 7.68\% compared with 17.31\% for Buy-and-Hold, with a lower Sharpe ratio (0.35 compared with 0.88), a lower Sortino ratio (0.51 compared with 1.32), and a deeper maximum drawdown (-13.14\% compared with -10.66\%). It therefore underperforms Buy-and-Hold on NYSE despite retaining a positive total return.} Across the four indices, VertiFuseX maintains positive total returns and Sharpe ratios. Naive Momentum is weaker overall, although it records a small positive return on DJIA and a positive return on NYSE.

In summary, the economic performance of VertiFuseX is market dependent. {The strategy nearly matches Buy-and-Hold on S\&P~500 and slightly exceeds it on DJIA. On NASDAQ, it accepts a lower total return in exchange for a shallower maximum drawdown, while NYSE remains the weakest case. The two stress-period evaluations in Section~\ref{subsec:extrememarket} provide the clearest evidence of downside protection during the market conditions examined. These results support the signal-level economic relevance of VertiFuseX, but they do not establish uniform dominance over passive investment.}

\subsection{Ablation Studies}
\label{sec:ablation}
To isolate the contribution of the penultimate-layer vertical fusion mechanism, we conduct a controlled ablation study on the S\&P 500 and NASDAQ indices, representing stable diversified and high-volatility technology-heavy markets, respectively. All variants share identical preprocessing, input window ($w=20$), univariate closing-price features, training protocol, regularization, and fixed hyperparameters (Table~\ref{tab:initial_params}). Only the fusion strategy varies. In our evaluation, we examine four configurations. The first is the St-LSTM, a standalone stacked LSTM that serves as the strongest single-branch baseline discussed in section 4.1. Next, we consider the Average Ensemble, which utilizes decision-level averaging of scalar predictions from independently trained LSTM, Bi-LSTM, and St-LSTM branches. The third configuration, Final-Layer Fusion, concatenates the final-layer hidden states from the three branches (prior to regression) and employs a shared dense prediction layer. Finally, we introduce VertiFuseX, which utilizes penultimate-layer vertical fusion with learned affine transformations, as outlined in section~3.3. The results of these configurations are presented in Table~\ref{tab:ablation}.

\begin{table}[htbp]
\centering
\caption{Ablation of fusion strategies on S\&P 500 and NASDAQ (lower is better).}
\label{tab:ablation}
\footnotesize
\setlength{\tabcolsep}{4pt}
\begin{tabular}{l ccc ccc}
\toprule
& \multicolumn{3}{c}{\textbf{S\&P 500}} & \multicolumn{3}{c}{\textbf{NASDAQ}} \\
\cmidrule(lr){2-4} \cmidrule(lr){5-7}
\textbf{Model Variant} & \textbf{MAE $\downarrow$} & \textbf{RMSE $\downarrow$} & \textbf{MAPE $\downarrow$ (\%)} & \textbf{MAE $\downarrow$} & \textbf{RMSE $\downarrow$} & \textbf{MAPE $\downarrow$ (\%)} \\
\midrule
St-LSTM (Standalone) & 46.49 & 59.31 & 1.12 & 195.82 & 243.98 & 1.56 \\
Average Ensemble & 43.18 & 55.21 & 1.04 & 184.35 & 229.67 & 1.45 \\
Final-layer Fusion & 39.87 & 51.96 & 0.93 & 170.42 & 214.86 & 1.32 \\
\textbf{VertiFuseX} & \textbf{30.22} & \textbf{40.57} & \textbf{0.59} & \textbf{134.54} & \textbf{180.87} & \textbf{0.83} \\
\bottomrule
\end{tabular}
\end{table}

The results in Table~\ref{tab:ablation} reveal a consistent performance hierarchy across both indices and metrics. The Average Ensemble provides modest error reduction (7.1\% and 5.9\% lower MAE on S\&P 500 and NASDAQ, respectively, versus St-LSTM), aligning with classical variance reduction for correlated predictors. Final-Layer fusion delivers more substantial gains (14.3\% and 13.0\% lower MAE), demonstrating the advantage of learned weighting over static averaging of compressed final representations.

{VertiFuseX achieves a substantial performance improvement, reducing MAE by an additional 24.2\% on S\&P 500 and 21.1\% on NASDAQ relative to Final-Layer fusion. Since both fusion variants have comparable parameter counts at the combination stage, this margin cannot be attributed to increased capacity. This comparison at a comparable fusion-stage parameter count is consistent with the rationale in Section~\ref{sec:rationale_R2}, which suggests that final-layer fusion is limited not by fewer parameters but by operating on already-compressed outputs. The persistence of the gain at comparable capacity therefore points to the fusion location as the more likely explanation.}

The disproportionately large MAPE reduction (36.6\% and 37.1\% relative to Final-Layer fusion) further indicates enhanced stability during volatile periods, consistent with gradient-based saliency focusing on mid-range lags (Section \ref{subsec:baseline_com}). These results establish that penultimate-layer vertical fusion is the primary driver of performance, beyond simple ensembling or decision-level integration.

\section{Discussion}\label{sec:extrememarket}
This section discusses two key aspects of VertiFuseX: (i) robustness analysis under extreme market regimes and (ii) the limitations of the proposed model. We first analyze model behavior under stress periods and then outline the assumptions and constraints that bound practical deployment.

\subsection{Robustness Analysis Under Extreme Market Regimes}
\label{subsec:extrememarket}
To assess the out-of-sample stability of VertiFuseX during tail events, where standard metrics (MAE, RMSE, MAPE) are insufficient due to structural breaks and volatility clustering, we conduct a dedicated stress-testing analysis distinct from the primary evaluation. Performance is isolated across two regimes: the exogenous liquidity shock of the COVID-19 crash and the endogenous valuation compression of the 2022 bear market. The model is trained solely on data from 1 January 2010 to 27 December 2019, with all parameters frozen thereafter. No retraining or adaptation occurs. The test period spans 27 December 2019 to 31 December 2024 (1260 trading days), covering both regimes under strict out-of-sample conditions. The stress-period trading simulation retains the next-day-open execution, transaction-cost assumption, triple-barrier exits, and three-day cool-down described in Section~\ref{subsec:trading_Strategy}. For this stress-specific evaluation, signals enter long positions when forecasts exceed the 75th percentile of the training-period distribution. Benchmarks are Buy \& Hold and Naive Momentum (long when the previous day’s return is positive).

\subsubsection{COVID-19 Crash (19 February -- 30 April 2020)}

This high-velocity shock produced a drawdown greater than 33\% in the S\&P 500. VertiFuseX limited maximum drawdown to -26.45\% (7.22 percentage points below Buy \& Hold's -33.67\%) and achieved a total return of -9.63\% (outperforming Buy \& Hold by 6.45 percentage points). The Sharpe ratio was -0.56 (-0.98 for Buy \& Hold; -2.21 for Naive Momentum), with lower volatility (61.33\% vs. 68.13\% for Buy \& Hold).
Downside protection was achieved through internal signal reduction during volatility spikes, as increased forecast uncertainty limited threshold crossings. Additionally, timely stop-loss and cool-down activations effectively reduced exposure in mid-March.

\begin{table}[ht]
\centering
\caption{Performance Metrics During Extreme Market Regimes}
\label{tab:robustness}
\resizebox{0.5\textwidth}{!}{
\begin{tabular}{lccc} 
\toprule
\textbf{Metric} & \textbf{VertiFuseX} & \textbf{Buy \& Hold} & \textbf{Naive Momentum} \\
\midrule
\multicolumn{4}{l}{\textit{\textbf{Panel A: COVID-19 Crash (19 Feb -- 30 Apr 2020)}}} \\
\cmidrule(r){1-1} 
Total Return (\%)       & -9.63     & -16.08    & -19.78        \\
Sharpe Ratio            & -0.56     & -0.98     & -2.21         \\
Max Drawdown (\%)       & -26.45    & -33.67    & -23.61        \\
Volatility (\%)         & 61.33     & 68.13     & 46.18         \\
\addlinespace[10pt] 

\multicolumn{4}{l}{\textit{\textbf{Panel B: 2022 Bear Market (1 Jan -- 31 Oct 2022)}}} \\
\cmidrule(r){1-1}
Total Return (\%)       & -10.04    & -19.56    & -30.31        \\
Sharpe Ratio            & -0.51     & -1.04     & -2.41         \\
Max Drawdown (\%)       & -17.67    & -25.38    & -30.50        \\
Volatility (\%)         & 23.45     & 24.33     & 18.26         \\
\bottomrule
\end{tabular}}
\end{table}

\begin{figure}[ht]
    \centering
    \includegraphics[width=0.6\textwidth]{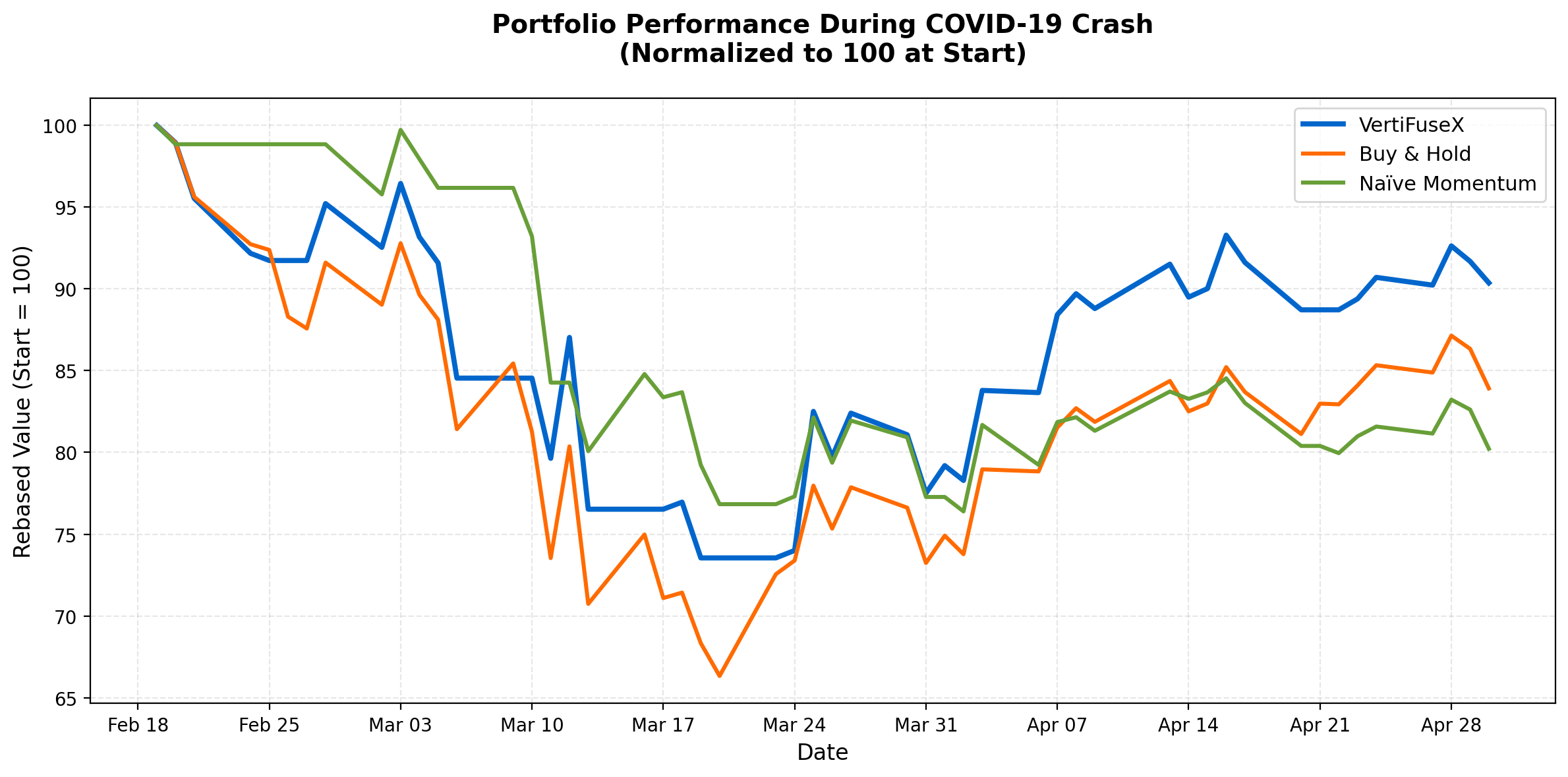}
    \caption{Normalized portfolio performance during the COVID-19 market crash (19 Feb–30 Apr 2020). 
    All strategies are rebased to 100 at the start of the period. VertiFuseX exhibits a shallower drawdown 
    and a faster post-trough recovery compared to Buy \& Hold and Naive Momentum, consistent with the 
    drawdown and Sharpe ratio improvements reported in Table~\ref{tab:robustness}.}
    \Description{}
    \label{fig:covid_equity}
\end{figure}

\subsubsection{2022 Bear Market (1 January -- 31 October 2022)}
This prolonged decline yielded a 25\% S\&P 500 loss. VertiFuseX recorded a total return of -10.04\% (outperforming Buy \& Hold by 9.52 percentage points) and maximum drawdown of -17.67\% (7.71 percentage points below Buy \& Hold's -25.38\%). The Sharpe ratio was -0.51 (-1.04 for Buy \& Hold; -2.41 for Naive Momentum), with marginally lower volatility (23.45\% vs. 24.33\% for Buy \& Hold). Naive Momentum returned -30.31\%.
Outperformance is achieved through selective exposure and minimum-hold constraints, while the cool-down mechanism helped prevent re-entry into false reversals.

\begin{figure}[ht]
    \centering
    \includegraphics[width=0.6\textwidth]{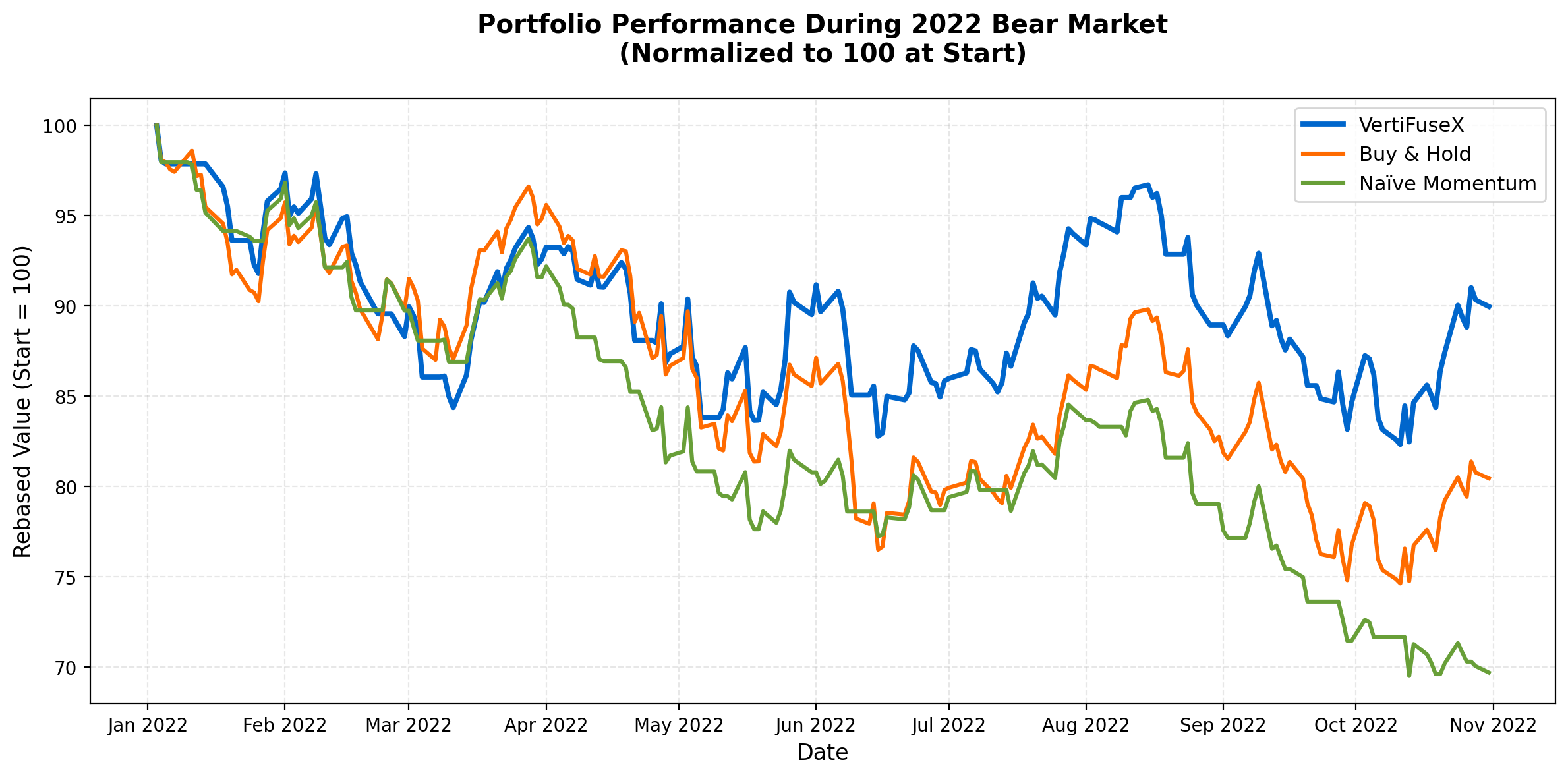}
    \caption{Normalized portfolio performance during the 2022 bear market (1 Jan–31 Oct 2022). 
    VertiFuseX maintains higher capital preservation and avoids prolonged drawdown accumulation 
    relative to Buy \& Hold and Naive Momentum, reflecting selective exposure and effective 
    risk controls during a sustained downturn.}
    \Description{}
    \label{fig:bear_equity}
\end{figure}

Fig.~\ref{fig:covid_equity} and Fig.~\ref{fig:bear_equity} provide a visual representation of the robustness results in Table~\ref{tab:robustness}, illustrating how reduced exposure during high-volatility phases and controlled re-entry contribute to improved drawdown and risk-adjusted performance across both crisis regimes.
The consistent reductions in drawdowns and risk-adjusted improvements across regimes validate the architecture's robustness. The vertical fusion of multi-scale temporal features enables more resilient forecasting under distributional shifts, while the empirical signal threshold acts as an implicit volatility filter. Explicit risk controls, particularly the cool-down after stop-loss, mitigate error clustering and provide layered protection without regime-specific tuning.

\subsection{Limitations}
While VertiFuseX demonstrates strong predictive performance and cross-market robustness, its design rests on several deliberate assumptions and is subject to theoretical and operational limitations that should be explicitly discussed and carefully considered for practical deployment.
\begin{enumerate}
    \item VertiFuseX is evaluated under a univariate, price-only assumption (closing prices), implying that historical price dynamics alone suffice for short-horizon forecasting. Although such dynamics may implicitly reflect market sentiment and macroeconomic conditions, performance may degrade during extreme exogenous events such as geopolitical shocks or sudden regulatory changes that are not reflected in prior market history. The current formulation does not explicitly incorporate broader macroeconomic indicators, cross-asset correlations, or market microstructure features.
  
    \item {VertiFuseX adopts a deliberately fixed hyperparameter configuration applied uniformly across all indices to prevent overfitting, ensure reproducibility, and promote consistent cross-market evaluation. Although this enhances architectural stability and avoids data leakage, it represents a conscious trade-off because peak accuracy for each market is sacrificed in favor of broader robustness. While internal feature reweighting allows adaptation to different volatility regimes, index-specific hyperparameter optimization could potentially yield further gains.}

    \item The primary evaluation relies on strict chronological out-of-sample testing with a single held-out test period per dataset, complemented by targeted stress testing in extreme regimes (COVID-19 crash and 2022 bear market). However, the study does not yet employ a full rolling-window walk-forward protocol with multiple overlapping folds. Consequently, aggregated performance statistics across repeated regime transitions and formal cross-regime statistical comparisons remain outside the current scope.
    
    \item The model produces deterministic point forecasts and gradient-based saliency analysis but does not provide probabilistic prediction intervals or calibrated uncertainty estimates. While suitable for low-latency directional forecasting, the outputs are not yet equipped for highly risk-sensitive applications requiring formal confidence bounds.

    \item The study includes a controlled algorithmic trading simulation (see Section 4.4) to evaluate its economic relevance. This simulation takes into account transaction costs, next-day execution issues (such as slippage and overnight gaps), confidence thresholding, and risk-adjusted metrics. However, it employs simplified trading logic and conservative assumptions, serving primarily as a proof of concept rather than a fully optimized trading system. Future work should focus on comprehensive strategy optimization, portfolio-level integration, and testing in real-world execution scenarios.

\end{enumerate}
\section{Conclusion and Future Work}

In this study, we present VertiFuseX, a hybrid LSTM-based architecture that enhances temporal modeling in financial time-series forecasting via penultimate-layer vertical fusion and multi-scale abstraction. Unlike traditional models using naive final-output concatenation, VertiFuseX fuses penultimate-layer features from LSTM, Bi-LSTM, and St-LSTM, preserving richer intermediate temporal representations before task-specific compression. In a comprehensive evaluation spanning 15 years of global market data across three continents, VertiFuseX achieves up to 54.3\% reductions in MAPE and over 40\% reductions in both MAE and RMSE relative to LSTM-family baselines, and outperforms seven state-of-the-art models across 33 comparative evaluations. {For practitioners, this suggests that monitoring mid-range price dynamics may support regime-aware forecasting and feature design, although we do not treat saliency patterns as trading rules.} {These results demonstrate that VertiFuseX advances financial forecasting by unifying simplicity, interpretability, and cross-market robustness under the evaluated chronological protocols, while remaining lightweight (approx. 675k parameters) with low inference latency ($1.5$ ms/sample) for real-time analytical systems.}

{Future work will strengthen validation and extend scope in four focused directions. First, although the present study uses a strict static chronological out-of-sample test, future work will evaluate VertiFuseX under rolling-origin walk-forward protocols to assess stability across multiple regime transitions and temporal folds. Second, while the current design deliberately uses univariate closing-price inputs to isolate the contribution of penultimate-layer vertical fusion, future studies will examine multivariate extensions using OHLCV variables, technical indicators, macro-financial covariates, and cross-asset inputs. Third, systematic sensitivity analyses over key design choices, including the lookback window, latent fusion dimension, dropout rate, and forecasting horizon, will clarify the stability boundaries of the architecture. Fourth, future work will extend the framework to leave-one-market-out transfer, multi-horizon forecasting, uncertainty-aware prediction intervals, alternative fusion operators including gated, attention-based, and cross-attention mechanisms in a multivariate setting, and complementary explainability methods such as Integrated Gradients and SHAP. These extensions would broaden empirical coverage and deployment relevance while preserving the central architectural principle of multi-stream penultimate-layer fusion.}

\bibliographystyle{ACM-Reference-Format}
\bibliography{mybibfilenew}

\end{document}